\documentclass[acmsmall]{acmart}
\usepackage{enumitem}
\usepackage{tabularx}
\usepackage{tabulary}
\usepackage{caption}
\usepackage{subcaption}
\usepackage{soul} 
\setcopyright{acmcopyright}
\copyrightyear{2025}
\acmYear{2025}
\acmDOI{10.1234/1234567.1234567}

\acmJournal{THRI}
\acmVolume{9999}
\acmNumber{1}
\acmArticle{1}
\acmMonth{8}

\usepackage[linesnumbered,ruled,vlined]{algorithm2e}

\usepackage{amsmath,amsthm,amssymb,amsfonts}
\usepackage{bm}

\newcommand{\famsec}{FaMSeC}
\newcommand{\cA}{\ensuremath{\mathcal{A}}} 
\newcommand{\cC}{\ensuremath{\mathcal{C}}} 
\newcommand{\cO}{\ensuremath{\mathcal{O}}} 
\newcommand{\cI}{\ensuremath{\mathcal{I}}} 
\newcommand{\cS}{\ensuremath{{S}}} 
\newcommand{\cSig}{\ensuremath{\Sigma}} 

\newcommand{\IS}{\ensuremath{\cI_S}} 
\newcommand{\IO}{\ensuremath{\cI_O}} 
\newcommand{\IH}{\ensuremath{\cI_H}} 
\newcommand{\IA}{\ensuremath{\cI_A}} 
\newcommand{\IM}{\ensuremath{\cI_M}} 

\newcommand{\solve}{\ensuremath{\mathcal{S}}}
\newcommand{\solveopt}{\ensuremath{\solve^*}}
\newcommand{\solvecand}{\ensuremath{\solve^C}}
\newcommand{\solvetrust}{\ensuremath{\solve^T}}
\newcommand{\taskclass}{\ensuremath{c_{\mathcal{T}}}}
\newcommand{\task}{\ensuremath{\mathcal{T}}}
\newcommand{\taski}{\ensuremath{\mathcal{T}_i}}

\newcommand{\rwdapprox}{\ensuremath{\widetilde{\mathcal{R}}}}

\newcommand{\rwdcandsim}{\ensuremath{\widetilde{\mathcal{R}}^C}}
\newcommand{\rwdtrust}{\ensuremath{\mathcal{R}^T}}

\newcommand{\rwdtrustpredict}{\ensuremath{\hat{\mathcal{R}}^T}}

\newcommand{\rwdtrustisim}{\ensuremath{\widetilde{\mathcal{R}}^T_i}}

\newcommand{\policyopt}{\ensuremath{\mathcal{\pi^*}}}
\newcommand{\policytrust}{\ensuremath{\mathcal{\pi}^T}}
\newcommand{\policycand}{\ensuremath{\mathcal{\pi}^C}}
\newcommand{\surrogate}{\ensuremath{\mathcal{M}(\mathcal{T})}}
\newcommand{\surrogatearg}[1]{\ensuremath{\mathcal{M}(#1)}}

\newcommand{\hell}{\ensuremath{H^2}}

\def\-{\raisebox{.75pt}{-}} 

\newcommand{\pigeneric}{\ensuremath{\pi}}

\newcommand{\piopt}{\ensuremath{\pi^*}}

\newcommand{\emcts}{\ensuremath{e_{mcts}}}

\newcommand{\itsmcts}{\ensuremath{its_{mcts}}}
\newcommand{\rwdexit}{\ensuremath{rwd_{exit}}}
\newcommand{\rwdcaught}{\ensuremath{rwd_{caught}}}
\newcommand{\rwdsense}{\ensuremath{rwd_{sense}}}
\newcommand{\metauo}{M_{\cO}}

\newcommand{\rev}[1]{\textcolor{black}{#1}}
\setstcolor{red}

\begin{document}

\title{``A Good Bot Always Knows Its Limitations'': Assessing Autonomous System Decision-making Competencies through Factorized Machine Self-confidence}
\author{Brett Israelsen}
\email{brett.israelsen@rtx.com}
\affiliation{%
  \institution{RTX Technology Research Center}
}
\author{Nisar R. Ahmed}
\email{Nisar.Ahmed@colorado.edu}
\affiliation{ 
  \institution{Smead Aerospace Engineering Sciences, University of Colorado Boulder}}
\author{Matthew Aitken}
\affiliation{%
  \institution{Allen Institute for Brain Science}
}
\author{Eric W. Frew}
\email{Eric.Frew@colorado.edu}
\author{Dale A. Lawrence}
\email{Dale.Lawrence@colorado.edu}
\author{Brian M. Argrow}
\email{Brian.Argrow@colorado.edu}
\affiliation{%
  \institution{Smead Aerospace Engineering Sciences, University of Colorado Boulder}
  \streetaddress{3775 Discovery Drive}
  \city{Boulder}
  \state{CO}
  \postcode{80301}
}
  
\renewcommand{\shortauthors}{Israelsen and Ahmed, et al.}

\begin{abstract}
How can intelligent machines assess their competency to complete a task? This question has come into focus for autonomous systems that algorithmically make decisions under uncertainty. We argue that machine self-confidence---a form of meta-reasoning based on self-assessments of system knowledge about the state of the world, itself, and ability to reason about and execute tasks---leads to many computable and useful competency indicators for such agents. This paper presents our body of work, so far, on this concept in the form of the Factorized Machine Self-confidence (FaMSeC) framework, which holistically considers several major factors driving competency in algorithmic decision-making: outcome assessment, solver quality, model quality, alignment quality, and past experience. In FaMSeC, self-confidence indicators are derived via ‘problem-solving statistics’ embedded in Markov decision process solvers and related approaches. These statistics come from evaluating probabilistic exceedance margins in relation to certain outcomes and associated competency standards specified by an evaluator. Once designed, and evaluated, the statistics can be easily incorporated into autonomous agents and serve as indicators of competency. We include detailed descriptions and examples for Markov decision process agents, and show how outcome assessment and solver quality factors can be found for a range of tasking contexts through novel use of meta-utility functions, behavior simulations, and surrogate prediction models. \rev{Numerical evaluations are performed to demonstrate that \famsec{} indicators perform as desired (references to human subject studies beyond the scope of this paper are provided).}
\end{abstract}



\maketitle

\section{Introduction}\label{sec:intro}
{\em ``It ain't what we don't know in life that gives us trouble. It's what we know that ain't so. ''} \\ -- Will Rogers
\bigskip

\noindent {\em ``We see a person do something, and we know what else they can do\ldots we can make a judgment quickly. But our models for generalizing from a performance to a competence don’t apply to AI systems. ''} \\ -- Rodney Brooks, \emph{IEEE Spectrum} interview, May 2023 \cite{zorpette2023just}
\bigskip

Despite their humble human origins, autonomous \rev{systems} like intelligent robots and self-driving vehicles often succeed and fail at tasks in distinctly \emph{non-human} ways. \rev{This is a problem because users unavoidably place \emph{trust} in such systems when delegating tasks of any size or complexity to them. Trust is a multi-dimensional construct, and some of those dimensions are related to being able to understand the trustee \cite{Israelsen2019-xm}.}
The combination of their sometimes superhuman capabilities---and equally `super' failures---drives many ongoing debates about the proper role of autonomous \rev{systems}, in which
particular success/failure cases are singled out as evidence for broad claims of (in)capability. In reality, by their very non-humanness, the feats and follies of 
\rev{such systems} should alert us to at least two fundamentally vexing issues signaled in the opening quotes above: 1) the innate ability of humans to infer the competencies of their fellow autonomous human beings \emph{does not} directly translate into an ability to infer the competencies of autonomous 
\rev{systems}; and 2) 
\rev{autonomous systems} generally lack an innate awareness of, or ability to reason/communicate about, their own competencies, which is \rev{a basic ability}
humans expect when interacting with others \rev{(related to the mental process of `mirroring' \cite{coeckelbergh2011artificial,carr2014mirroring})}. 
The systematic inability to provide knowledge about their competency boundaries\rev{, or the limits of competency,} opens the door to technological misuse, disuse, and abuse \cite{parasuraman1997humans,Israelsen2019-xm} -- issues which are bound to persist as this technology continuously evolves. Misconceptions about the capabilities of 
\rev{autonomous systems} in particular can quickly take root and become hard to dispel, unless well-defined design and evaluation practices are instilled to support meaningful assessments \cite{schaeffer2024emergent, tolmeijer2024trolleys}.

\rev{Autonomous systems come in many forms; some are `embodied' (i.e. air and land vehicles), while others are not (chatbots and decision-making assistants). Such systems also have varying levels of autonomy, requiring differing levels of human intervention/interaction across the spectrum. Our focus here is on the algorithms, data, and models that underlie such systems; we refer to these parts of an autonomous system as the `agent'. The agent can typically be deployed on many physical systems with minimal effort (as in the case of a fleet of identical vehicles); it could also be adapted, with varying degrees of effort, to fundamentally different but `similar' systems (e.g. a planning algorithm installed on a wheeled robot or an aerial drone). We give a more precise definition of what is meant by `agent'\footnote{\rev{`agent' here is \emph{singular}; we do not consider multi-agent systems, though some concepts may be transferrable.}} in this paper in Section~\ref{sec:background}}.
 
Short of putting the genie back in the lamp or achieving the mythical holy grail of pure, human-free autonomy anytime soon \cite{bradshaw2013seven}, the major question 
is not whether the science says \rev{autonomous agents} with generalizable superhuman abilities could/should co-exist with humans. Rather, it is whether the \emph{engineering} behind such technology is sound and capable enough for people who are \emph{not} roboticists, computer scientists, et cetera, to responsibly evaluate and use it for solving the problems they actually need to solve \cite{Jordan2019Artificial}, \rev{and to do so without having to rely on blind trust.} To examine this issue more deeply, this paper considers how the competencies of autonomous 
\rev{can be quantified} in principled ways to aid designers, users, and ultimately the agents themselves, in deciding how and when they should be entrusted to undertake specific decision-making tasks under uncertainty.

Perceived competency is one of 
\rev{several} factors that influence human trust in technology \cite{Israelsen2019-xm}. \rev{Competency is a useful construct in evaluating the ability of an agent to perform as desired/expected; it is commonly used to evaluate how well a person is able to perform a certain task.} 
Echoing psychology and business parlance, we define competencies \rev{of autonomous agents} broadly as the set of skills (generalizable, non-task-specific capabilities) and behaviors (observed performance and approaches to particular tasks) that an evaluator expects from an algorithmic agent if they are to execute tasks up to some desired standards \cite{stevens2013critical}. 
\rev{\emph{Assessment of competency} is the process of evaluating the degree to which an agent possesses particular competencies.} 
Artificially intelligent agents are designed with capabilities 
\rev{such as} perception, reasoning, learning, control, and communication. Each of these are realized through a combination of 
engineered components. \rev{These agents are expected to have a certain level of competence with respect to each of their associated capabilities. However, whether an agent can be competent in a specific situation is highly dependent on several factors, including assumptions made before and during the design and implementation of the agent. In this paper, we present definitions for the competency and competency assessment of artificial agents, as well as a framework for gauging, in particular, the \emph{decision-making competencies} of agents. }
\rev{As algorithmic decision-making agents come in many varieties, we focus here on the broad category of model-based rational (goal-seeking) decision-making agents that can be defined via Markov Decision Processes (MDPs), which are the basis of many modern stochastic control, optimal planning, and reinforcement learning techniques used in autonomous systems. }

\rev{Evaluating competency is not an easy endeavor} given the huge variety of ways in which algorithmic decision-making can be implemented (from \rev{more} general purpose stacks to specialized task-specific programs). How can component competencies \rev{spanning this variety of implementations} be sensibly evaluated and fairly compared? 
\rev{A straight-forward answer is elusive} 
\rev{because} decision-making algorithms have become increasingly sophisticated and opaque, with their many variables, assumptions, and data-driven components interacting in non-intuitive, non-physically grounded, and non-deterministic ways. 
Moreover, agents often operate in settings with 
myriad 
uncertainties that cannot be easily quantified due to inherent task complexity (including lack of complete information in dynamic settings), practical design limitations (including use of approximations to ensure computational tractability), and 
\rev{the complexity of accurately interpreting and encoding true user intent}. These factors will invariably push agents beyond their limits \rev{at some point}, necessitating human intervention. 
\rev{Assuming competency could be determined, communicating competency} is also challenging 
\rev{since} users, system designers, 
\rev{and others} 
don't \rev{have ready access to} an agent's decision-making process. 
This is especially true if an autonomous agent, 
\rev{such as an autonomous delivery vehicle,}
is tasked in a variety of contexts requiring significantly different skill and behavior standards. \rev{Moreover, it is generally difficult to tell whether a given competency assessment is `correct', i.e. the perceived competency of an agent may or may not (to varying degrees and in different circumstances) describe its actual competency.} 

These issues have motivated wide-ranging research on techniques aimed at improving the explainability, transparency, and interpretability of decision-making algorithms for users \cite{ali2023explainable,kaur2022trustworthy,dwivedi2023explainable,minh2022explainable,elizalde2009generating,huang2018establishing}. 
Yet, these methods usually assume agents are already capable of performing assigned tasks in the first place, i.e. they don't allow agents to \rev{independently} assess 
whether certain tasks \emph{should} be assigned to them at all, or if competency limits will be (or have been) exceeded. 
Others have addressed this gap via formal methods or introspective knowledge-based reasoning frameworks \cite{basich2023competence,basich2020learning,Svegliato-IROS-2019,frasca2020can,frasca2022framework,infantino2013humanoid, Aguado-Sensors-2021,Gautam-ICRA-2022,Cao-TRO-2023}, but require strong assumptions about a priori task knowledge and agent design that lose purchase in the presence of uncertainties. \rev{Furthermore, }some of these methods 
only focus on closing the self-assessment loop to enable efficient agent adaptation and task-specific performance improvement, sacrificing the ability to provide holistic human-interpretable competency statements. \rev{This is motivated and discussed in further detail in Section~\ref{sec:related_work}.}

To address these gaps, we aim to develop a principled \emph{engineering-based} framework which allows autonomous agents to (truthfully and objectively) self-examine and communicate their decision-making competency limits.
\rev{Here the term \emph{engineering-based} is meant to highlight the fact that, in almost every other engineering discipline, the design of a system, product, or component includes calculating `design confidence' and using `factors of safety' in uncertainty-dominated settings.} 
Two core \rev{engineering} principles stipulate that designers should: 1) employ \emph{well-calibrated uncertainty models} for key assumptions and design variables; and 2) strive for \emph{sound design margins} to minimize sensitivity to uncertainties \cite{Eckert-2014, Eckert-2019}. Translated 
into practical terms, design confidence increases with larger design margins, and decreases with more pronounced uncertainties. 
These ideas are well-established for designing \rev{and analyzing} automated feedback control systems and statistical estimators like Kalman filters, which have enjoyed continued use for the past several decades as direct forerunners to (and still vital components of) modern-day autonomous systems and robots. 
Combined with rigorous mathematical formalisms, these principles provide a basis for \rev{(or at least an informal feasibility proof of) the idea that} automated systems 
\rev{should be able to} 
assess their \rev{own} operational limits \cite{djuric2019self,safonov1997special}. 
We argue that there is ample opportunity to formally  
\rev{apply} these principles to the design of interpretable, competency-aware decision-making 
\rev{agents}.

This paper presents the culmination of our recent and ongoing work to this end~\cite{aitken2016thesis, aitken2016workshop, Israelsen2019-xm, Israelsen-thesis, israelsen2020machine, acharya2022competency, acharya2023learning, conlon2022generalizing, conlon2022iros, conlon2024event}. \rev{Our approach has been }via the concept of \emph{machine self-confidence}---a 
\rev{competency} meta-reasoning \rev{framework} based on self-assessments of an agent's knowledge about the state of the world and itself, as well as its ability to reason about and execute tasks~\cite{Hutchins-HFES-2015,Sweet-SciTech-2016}. 
We elaborate more deeply and rigorously on the conceptual foundations of machine self-confidence (touched on only briefly in previous work), and show that these lead to several useful kinds \rev{of} inter-related introspective competency indicators for typical algorithmic decision-making processes that can be conveyed in human-understandable ways. 
The main contributions of this work \rev{are as follows:} 
\rev{\begin{enumerate}
    \item We provide general definitions of \emph{competency} and \emph{competency assessment} for algorithmic agents, and link these to rational decision-making agents defined via Markov Decision Processes (MDPs), as well as to the key concept of bounded rationality. 
    We also establish definitions for \emph{competency indicators} for gauging agent capabilities via  \emph{competency standards}, specified by external (human) evaluators to define the baseline for acceptable (or required) performance. 
    The relationships of these concepts to agent and context variables for specified tasks are explained. These ideas are illustrated by a detailed running example involving an autonomous delivery robot operating in an uncertain adversarial environment. 
    \item A detailed description of our computational framework for algorithmic decision-making competency self-assessment, called \emph{Factorized Machine Self-Confidence} (\famsec{}). \famsec{} includes five competency indicators: \emph{expected outcome assessment, solver quality, model and data validity, alignment with user intent, and historical performance and experience}. These indicators leverage `problem-solving statistics' that are readily available to MDP agents and collectively enable self-assessments for broad classes of decision-making agents under various types of uncertainties and tasking contexts. We define each factor and provide insights for their complementary nature, discussing their expected behaviors on different hypothetical instances of the autonomous delivery problem. 
    \item Detailed theoretical considerations and computational implementations for two of the five \famsec{} factors: expected outcome assessment and solver quality. 
    While implementations for these two factors were briefly introduced in \cite{Israelsen2019-xm}, we significantly expand on that work by also providing more detailed and complete descriptions of our implementations, along with other possibilities for implementations. We also provide more thorough justifications for competency standards and problem-solving statistics for our proposed (non-unique) computational implementations involving the novel use of meta-utility functions, behavior simulations, and surrogate prediction models for competency assessment. Properties of the proposed indicators are illustrated via simulation results of the autonomous robot delivery problem in different task settings.  
    \item Discussion of the remaining three \famsec{} indicators, which were not given detailed consideration in our prior works introducing the associated factors \cite{aitken2016workshop,Israelsen2019-xm}. We specifically describe how these factors pose challenging fundamental issues for computational realizations, highlight aspects of the interrelated and asymmetric nature of all five \famsec{} factors, and outline promising ideas to examine for future development of computational factor-based assessments. 
    \item Consideration of other major practical issues involved with algorithmic competency self-assessment, specifically: distinguishing between a priori, in situ, and post hoc assessments; applying \famsec{} to agents based on other (non-MDP-based) decision-making algorithms; validating \famsec{} indicators and distinguishing between actual vs. perceived agent competency; and validating competency reporting with human users (human-subject studies are not included in this paper, but relevant work to that end is discussed and cited). 
\end{enumerate}}

\rev{We hope this paper can guide and inspire further competency assessment-related work. It is challenging to work on a problem that hasn't been defined clearly; the definitions herein should be helpful in focusing how the community thinks about this problem. Furthermore, our efforts to move beyond philosophy and theory with the \famsec{} framework, and associated competency indicators, should also be helpful at least as a prototype for future efforts; we recognize there is much work still to be completed.}
\rev{From a theoretical standpoint, \famsec{} addresses many of the limitations found in other approaches described above. This work also demonstrates feasibility of using a meta-cognitive approach for assessing competency in rational decision-making agents. The meta-cognitive competency indicators we describe}
formally depend on task context, uncertainty, and observed vs. expected agent performance \emph{and} behaviors (all defined more precisely later). 
\famsec{} indicators can be readily adapted to evaluate how an agent executes tasks in multiple assessment contexts, without requiring evaluators to have knowledge of algorithmic design details.  
This permits flexible yet holistic evaluation of decision-making competencies, which is especially useful when design foresight is constrained for uncertain tasks. In such cases, agent models, objectives, data, and reasoning mechanisms could fall short of/exceed competency standards in different, unexpected ways for variable and new task contexts. 
\rev{Of course, our current implementation of \famsec{} has known limitations, which we summarize in Section~\ref{sec:concs}. 
For example, important work remains to understand how competency assessment frameworks like \famsec{} relate to formal methods for certifying autonomous agent behaviors (see Section~\ref{sec:related_work}). 
Yet, there is evidence that \famsec{}, as presently formulated, has clear benefits despite these limitations \cite{acharya2022competency,conlon2022generalizing,israelsen2020machine,conlon2022iros,conlon2024event}.  
And though \famsec{} will not be an `out of the box' solution to every possible competency assessment problem, the framework will hopefully provide a structured way for autonomous system designers to think through the issues confronting them (i.e. what assumptions/variables \emph{can} an agent assess within the confines of bounded rationality? what \emph{are} the right competency indicators and standards?) and shed light on new ways forward. 
}

The rest of the paper is outlined as follows.
In Section~\ref{sec:background}, a general computational description of algorithmic competency self-assessment is presented first to highlight and discuss core concepts. In Section~\ref{sec:famsec}, these are then translated into the \famsec{} framework for competency self-assessment and reporting for general decision-making agents. 
Sections~\ref{sec:outcomeassess} and \ref{sec:solverquality} provide detailed descriptions and running application examples for Markov decision process agents, showing how two of the more easily accessible \famsec{} factors (outcome assessment and solver quality, respectively) 
can be practically computed and reported in human-interpretable ways for a wide range of possible tasking contexts through the novel use of meta-utilities, behavior simulation, and surrogate performance models. 
Section~\ref{sec:otherfactpract} provides a synopsis of challenges for implementing the remaining factors, as well as generalizing and validating competency assessment in real-world systems. 
The paper reviews the literature on algorithmic assurances and competency self-assessment in Section~\ref{sec:related_work}, highlighting conceptually 
related work in robotics and explainable AI\footnote{As there are many approaches to algorithmic competency self-assessment, this review is deferred to the end so that the main ideas and our contributions can be presented as clearly and succinctly as possible.}. 
Finally, conclusions and limitations of our approach are presented in Section~\ref{sec:concs}.

\section{Background and Definitions}\label{sec:background}

This section introduces background for the major technical ideas and issues underlying competency self-assessment for algorithmic decision-makers. 
It opens with a brief review of Markov decision process (MDP) models, which describe a wide range of rational decision-making algorithms and are therefore useful archetypal representations for key concepts. 
A notional application involving an autonomous delivery vehicle operating in an adversarial environment is then presented to motivate and illustrate the various aspects of algorithmic decision making under uncertainty that underpin our view of the competency self-assessment problem. 
Formal definitions and a general framework for computational competency evaluation are then described, followed by considerations for practical implementation that motivate the formulation of our \famsec{} framework in Section~\ref{sec:famsec}. 

\subsection{Rational Agents and Competency Evaluation}\label{sec:rationalagentscomp}
Although there are many different notions of competency evaluation, each is ultimately based on comparing 
the expected outcomes of an agent's decision-making process to actual outcomes observed in reality. By examining different models of decision-making agents, the details of any such 
process can be refined to develop suitable approaches for competency self-assessment. 
To this end, we focus here on a broad class of rational (utility-maximizing) algorithmic agents that perform decision-making under uncertainty.    


 \subsubsection{Markov Decision Processes}
Markov decision processes (MDPs) provide a rigorous, well-understood algorithmic framework for rational decision making that also furnishes agents with the ingredients for competency evaluation (see Figure \ref{fig:CompAssessGenBlockDiag} later). In addition to their widespread use in modern control theory, robotics, and AI, MDP constructs lie at the core of a much broader landscape of algorithms for decision-making under uncertainty, embodied for example by hierarchical decision processes, stochastic optimal control, partially observable Markov decision processes (POMDPs), bandit techniques, and reinforcement learning (RL) \cite{kochenderfer2015decision}. Hence, given their importance, prevalence, and familiarity to many researchers, MDPs will be referred to extensively in this work to examine ideas behind  competency self-assessment for decision making algorithms from a model-based perspective. As discussed later in the paper, this does not exclude consideration of alternative rational decision making frameworks, as long as these furnish the same essential components to inform competency self-assessment as are used for MDPs\footnote{This is not a high bar in our opinion, since the `essential components' are fairly fundamental elements to most decision-making algorithms algorithms.}.
A brief review of MDPs is provided next to establish concepts that are useful to develop algorithmic competency self-assessment techniques; see references such as \cite{kochenderfer2015decision, thrun2005probabilistic} for detailed reviews. 

 An MDP for a sequential decision-making problem is parameterized by a tuple $(S,A,T,R)$, where: $S$ is the state space; $A$ is the set of actions that an agent can take from each state; $T$ is the state transition model, representing the conditional probability $P(s_{t+1}|s_{t},a_{t})$ of reaching state $s_{t+1} \in S$ from state $s_{t} \in S$ when taking action $a_t \in A$ at instance $t$; and $R$ represents the reward model, typically modeled as a function $r_t = r(s_t,a_t)$. By  construction of $T$, the dynamics of the decision-making problem are Markovian, i.e. the probability of reaching $s_{t+1}$ is conditionally independent of all other past states given only the immediate past state $s_{t}$ and action $a_{t}$. For optimal sequential decision-making, a policy $\pi(s)$ must be found to maximize a utility or value function $V(s)$ for all valid $s \in S$.  In most MDPs, the value function typically corresponds to the expected discounted cumulative reward $V(s)=\mathbb{E}[U(s)]$, where 
 \begin{align}
     U(s) = \sum_{t=0}^{\infty}{\gamma^t r_t},
 \end{align}
 and $\gamma \in [0,1)$ is a discount factor that trades off between near-term and long-term reward payoffs. The optimal policy thus corresponds to  
\begin{align} \label{eq:optmdppol}
\displaystyle \pi^{*}(s) = \arg \max_{\pi} V^{*}(s),  
\end{align}
 where $V^{*}(s)$ is the optimum expected value function. 
 
There are many approaches to obtaining $\pi^{*}$ either exactly or approximately; this is also referred to as `solving' the MDP. Hence, an MDP solver is an algorithm that operates on MDP specification $(S,A,T,R, \gamma)$ and returns a policy. 
The well-known value iteration algorithm \cite{thrun2005probabilistic, sutton2018reinforcement} provides a theoretically guaranteed way of finding $V^*$ and $\pi^*$ through iterated solutions to the underlying Bellman equations corresponding to (\ref{eq:optmdppol}). However, due to computational limitations, value iteration is not always feasible or practical, since it requires exhaustively searching the entire state-action space $S \times A$ to update estimates of $V(s)$ until convergence. This is seriously challenging in MDPs with large state-action spaces, as well as in settings where the time to obtain a solution policy $\pi$ is a key consideration (the two issues are typically correlated). 

 As such, it is common for tasks like autonomous mobile robot navigation to use a class of MDP solvers which approximates $V^*(s)$ and/or $\pi^{*}$ within realistic operational constraints.
 For instance, many  approximate MDP solvers find approximate policies $\hat{\pi}$ only over a restricted subset of the state-action space corresponding to states in $S$ that are reachable from the current state $s$; other approximate solvers learn regression models of $\pi^*(s)$ with respect to $S$ to simplify the optimization process \cite{kochenderfer2015decision}. Of the approximate solvers in the former category, Monte Carlo Tree Search (MCTS) has emerged as one of the most popular `online' MDP solvers, which yield good quality approximate policies $\hat{\pi}$ on demand for very large state-action spaces and provide improved approximations to optimal $V^*(s)$ and $\pi*$  with longer run times. Since the parameters required by the MCTS algorithm will be used later to illustrate the effect of different variables in competency evaluation for MDPs, some additional basic background is provided in Appendix \ref{sec:app_MCTS}; more complete details can be found in \cite{kochenderfer2015decision, kocsis2006bandit}. 

\subsubsection{Bounded Rationality: Ideal vs. Realistic Agents} 
The need for approximations like MCTS hints at two fundamental premises that apply in real life to all rational decision-making agents, whether or not they are based on MDPs. Firstly, such agents are boundedly rational \cite{russell1991principles, simon2000bounded}\footnote{This is also an explicit premise of some -- though not all -- other works on competency assessment, e.g. \cite{basich2020learning,basich2023competence, Svegliato-IROS-2019}.}. That is, agents programmed to operate rationally in the real world do not have infinite computing or representational power, and thus are inevitably imperfect reasoners with limited access to information and limited execution abilities. 
Secondly, such computational agents are designed by human designers who themselves are also boundedly rational. From these two premises, it follows that any agent's ability to execute tasks is influenced by a multitude of factors that fall into one of two broad categories: (1) those which were explicitly considered in the design of the decision making algorithm prior to deployment; and (2) those which were not. 
In other words, an algorithmic decision-making agent is built according to a specific set of design assumptions for its intended tasks. 
Assumptions are essential for allowing designers to focus limited computational resources and effort on particular aspects of decision-making problems that can be reasonably handled within the confines of bounded rationality. Assumptions are inherent, for instance, in the choice of state variables and policy solver parameters for MDP agents, as well as in the selection of training data to identify policies or world models for reinforcement learning agents. 
To the extent that design assumptions align with reality during actual task execution, there should be a high likelihood the agent will perform the task as competently as it was designed to. But, if \emph{any} assumption should fail to hold, this likelihood diminishes, often drastically, and competency is typically compromised. 

It is practically impossible to guarantee the validity of all design assumptions in all task contexts. This means the key question for competency assessment arguably centers on determining the likelihood and extent to which competencies for a particular agent operating could be impacted in a given operating context. 
These points will be examined in more detail through the illustrative example described next. 

\subsection{Illustrative Example: Autonomous Delivery in an Adversarial Setting} 
\label{ssec:deliveryMDP}

\begin{figure}[t]
    \centering
    \includegraphics[width=0.8\textwidth]{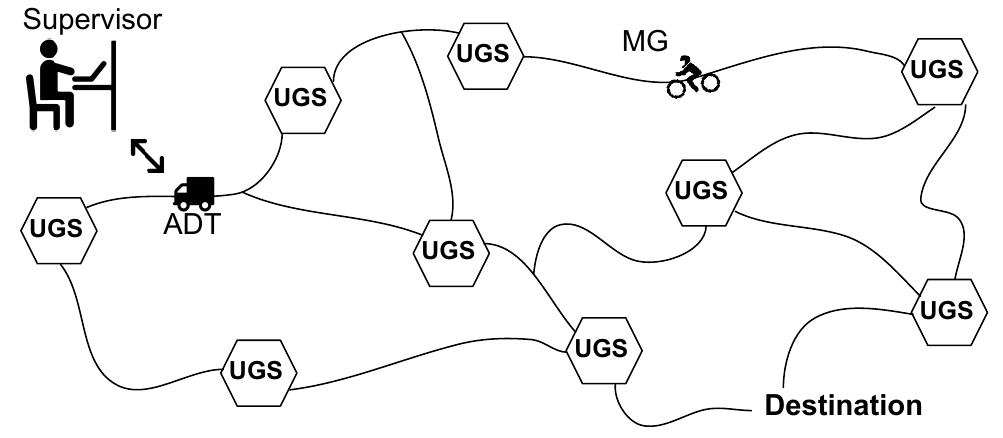}
    \caption{Road network for Autonomous Delivery Truck (ADT) problem.} 
    \label{fig:RoadNet}
\end{figure}

Figure~\ref{fig:RoadNet} depicts an autonomous doughnut delivery truck (ADT) navigating a road network in order to reach a delivery destination, while avoiding a motorcycle gang (MG) that will steal the doughnuts if it intercepts the ADT. The MG's location is unknown but can be estimated using measurements from unattended ground sensors (UGS). \rev{In this example the agent consists of the algorithms installed on the vehicle, and the agent's} decision space involves selecting a sequence of discrete actions (i.e. go straight, turn left, turn right, go back, stay in place) in order to safely navigate along segments of the road network and minimize the time to reach a specified destination on the map for each delivery hop. \rev{ Note that this application example was developed originally in \cite{Israelsen2019-xm} as a simplistic representation of much more complex problems, especially compared to other kinds of problems that we and other researchers have looked at in other prior work. We use this example again here to develop ideas in an accessible and fundamental way, which (as we shall discuss later) also flexibly accommodates the fact that some aspects of competency assessment are easier to examine from an algorithmic standpoint than others. 
Also note that since the focus of this paper is on agents, we may occasionally use the term ADT out of convenience, but really be referring to the agent.}

The ADT's motion, UGS readings, and MG's behavior are treated as random variables, and the problems of decision making and sensing are in general strongly coupled. Some trajectories through the network might allow the ADT to localize the MG via the UGS before heading to the delivery destination but incur a high time penalty. Other trajectories may afford rapid delivery with high MG location uncertainty but increase the risk of getting caught by the MG, which can take multiple paths. A human dispatch supervisor monitors the ADT during operation. The supervisor does not have detailed knowledge of or control over the ADT, but can (based on whatever limited information is available) decide to proceed with or abort the delivery run before it starts. The extent to which the 
\rev{agent} can complete the assigned task, i.e. its competency, will clearly be a significant factor in the supervisor's evaluation. 

To reach its destination, the 
\rev{agent} in the adversarial delivery problem must make a sequence of decisions within a discrete state and action space. An MDP-based \rev{agent} is well-suited for this problem. Specifically, the physical states $s$ describing the combined motion of the ADT and MG can be discretized in time and space to produce a Markov process model defined by some initial joint state probability distribution and joint state transition matrix $T$, which depends on the steering actions $a$ taken by the ADT. A reward function $R(s_k,a_k) = R_k$ can be specified for discrete decision instances (time steps) $k$ to encode user preferences over the combined state of the ADT and MG, for instance: $R_k =  -100$ for each time step the ADT is not co-located with the MG but not yet at the goal; $R_k = -1000$ if the ADT and MG are co-located; and $R_k = +1000$ if the ADT reaches the goal without getting caught. The reward values here are selected arbitrarily for illustrative purposes, but primarily reflect preferences for the ADT's actions and state occupancy. As detailed above, they can be tuned to induce more/less risk averse ADT behavior with respect to an overall utility or value function. If the MG’s state is observable at each step $k$ (due to very reliable and accurate UGS that cover all areas of the road network), then the ADT’s planning problem can be framed as an MDP. Otherwise, probabilistic beliefs about the MG's state can be inferred from the UGS observations, leading to a mixed observability MDP (MOMDP). If the ADT's states are also uncertain, this leads to a POMDP. For simplicity and without loss of generality, the simpler MDP framing is adopted here. 

\subsubsection{Competency Assessment for MDP-based 
\rev{agent}} \label{ssec:probsolvstatsmdps}
Recall the two categories of design factors in  boundedly rational agent design: (1) those which are explicitly considered; and (2) those which are not. 
For instance, factors in category (1) include the typical dynamics of the ADT navigating along the road network, as well as the existence and typical behaviors of the MG. Factors in category (2) could include how unusual weather and road conditions impact the UGS. 
Other factors beyond the supervisor's or 
\rev{agent's} control which could also impact the 
\rev{agent's} capabilities include whether enough computing power has been allocated onboard the ADT to \rev{allow the agent to} identify safe and efficient navigation solutions in all situations. The nature of the task also matters. Consider that an \rev{agent for an} ADT which can deliver donuts is just as likely capable of delivering bread loaves or medical supplies under similar circumstances, since such cargo does not fundamentally alter the task's difficulty\footnote{The MG's tastes notwithstanding.}. On the other hand, the imposition of post hoc constraints like minimizing fuel consumption, avoiding potholes, or strictly obeying speed limits at all times will (all else being equal) clearly be impactful, if designers did not originally account for these. 

As discussed in \cite{Cao-TRO-2023,Gautam-ICRA-2022}, explicitly delineating and reasoning over the built-in set of design assumptions provides a powerful starting point for determining whether a task is within an agent's expected competency range. 
We consider an alternative 
strategy for competency self-assessment that relies on the explicit construction and evaluation of various indicators to determine whether certain necessary conditions for decision-making competent behavior are met. 
A key distinction from previous work is that we separate decision-making competent behavior into two related but distinct levels. For one level of competency, the only concern is with \emph{what} outcomes the agent achieves. On another level, we are also interested in examining \emph{how} the agent achieves those outcomes. Whereas previous treatments have largely focused one or the other of these aspects of competency, we argue that \emph{both} aspects must be treated simultaneously within a holistic self-evaluation framework for autonomous systems. The next subsection provides a formal framework to rigorously ground these ideas for algorithmic competency assessment, before considering how such indicators can be practically evaluated for agents in operating in different contexts. 

\subsection{The Autonomous Tasking and Competency Evaluation Process}
Autonomous competency self-assessment is best viewed from the perspective of an agent attempting to complete a delegated task. 
In the adversarial delivery problem described above for example, the agent is the software which makes decisions autonomously for the ADT, so that it can meet its two assigned objectives of avoiding the MG and reaching the delivery destination in a timely manner. 
Since the agent will render decisions that affect the ADT's physical state and evolution in the road network environment, the agent also has an effect on how parts of the environment (in this case, the MG) will `respond' and thus render outcomes. However, the agent cannot affect all parts of the environment that impact the agent's ability to perform the task and achieve desired outcomes (e.g. the layout/structure of the road network). This example shows that it is not generally possible to analyze or understand competency by looking at a software agent in a vacuum. 
Rather, the assigned task \emph{and} the context in which it must operate in must be considered together\footnote{This echoes the `scissors of cognition' metaphor of
Herbert Simon (who also coined the term `bounded rationality'\cite{simon2000bounded}): `[Human] rational behaviour is shaped by a scissors whose blades are the structure of task
environments and the computational capabilities of the actor.' \cite{simon1990invariants}.}. 

\begin{figure}[tbp]
\centering
\includegraphics[width=1.0\columnwidth]{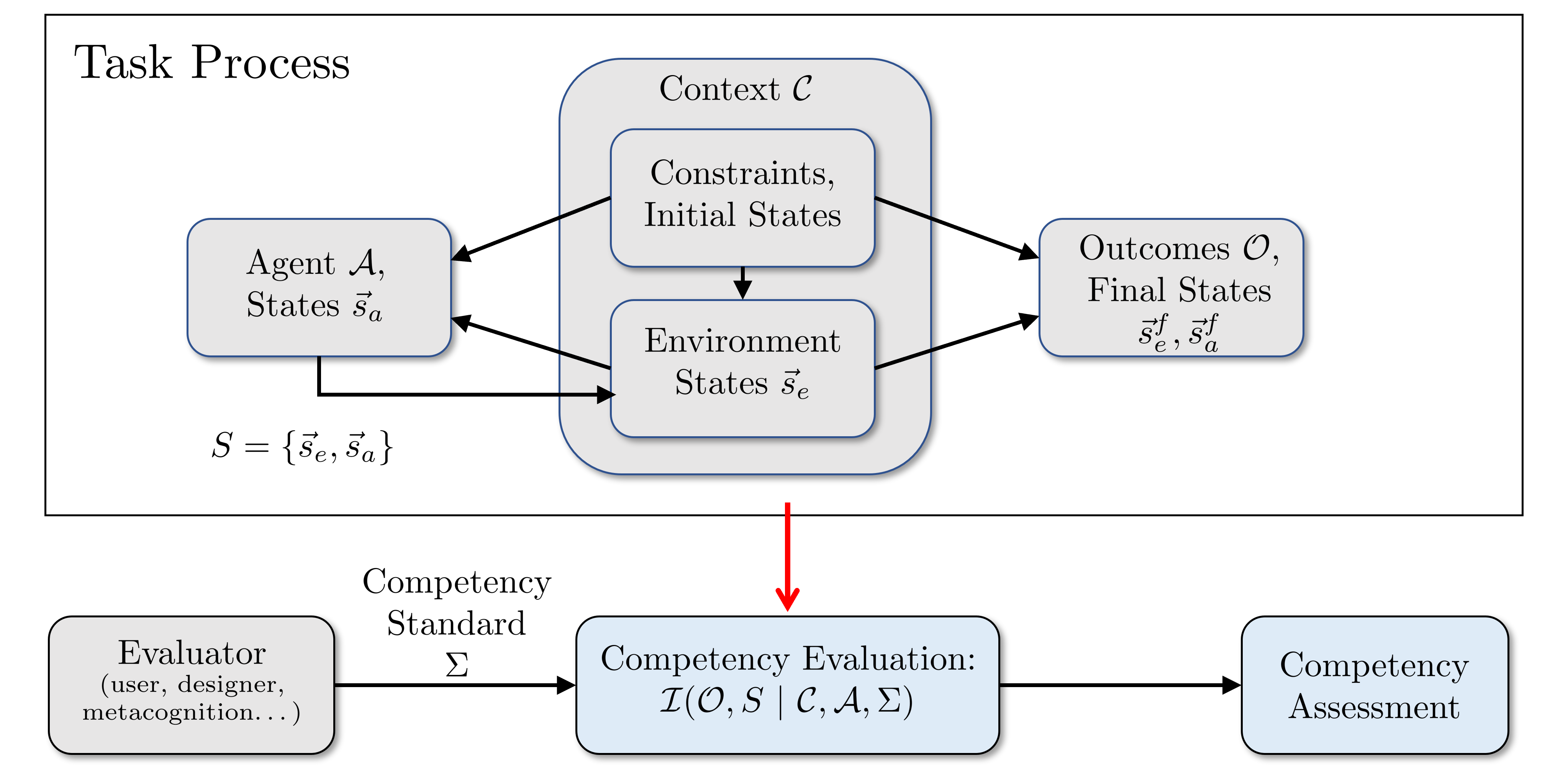}
\caption{Representation of key high-level factors in agent task completion and competency evaluation \rev{in terms of an agent \cA{} operating in context \cC{} being evaluated on outcomes \cO{} with respect to some standard of competency \cSig.} \rev{Variables are defined in significantly more detail in the text.} }
\label{fig:CompAssessGenBlockDiag}
\end{figure}

Figure~\ref{fig:CompAssessGenBlockDiag} illustrates the high-level factors involved in an agent completing a task. The construction in this figure generalizes to any algorithmic decision-making agent, whether or not it is based on an MDP or the like. The degree to which the task can be considered `complete' is measured by outcomes of the agent's sequence of actions, or proposed actions. To be precise, the components of the process are defined as follows:

\begin{description}
\item[Agent:] In this paper the agent, \cA, is taken to include the algorithms, models, and data necessary for a system to complete a given task or set of tasks as defined by a designer. Crucially, this also includes approximations that play an intrinsic role in optimizing/solving problems intended for said algorithms, models, and data. We view \cA{} separately from the environment and hardware in which it is implemented; these are instead considered in the `context' block of Figure~\ref{fig:CompAssessGenBlockDiag}. \rev{The state $\vec{s}_a$ represents variables that \cA{} uses to operate and solve its task. In the delivery example, \cA{} is an instantiation of an MDP solver algorithm installed on the ADT's computing hardware, and thus incorporates design assumptions about the task through an MDP specification for states, rewards, transition function, et cetera; the state $\vec{s}_a$ corresponds to the implemented MDP state, i.e. the location of the ADT and MG.} 

\item[Context:] \cA{} operates within a given context \cC. Figure~\ref{fig:CompAssessGenBlockDiag} illustrates that \cC{} is composed of the constraints and initial states of the agent \rev{as well as the environment state $\vec{s}_e$}. The constraints include the physical \rev{setting} and limitations of the \rev{selected} hardware, \rev{whereas $\vec{s}_e$ is a `catch all' set of \rev{dynamic} variables not in $\vec{s}_a$,} but are impactful to \cA{}'s operation and task performance. \cA{} operates within a physical environment; this environment may change via \cA's actions, other means, or a combination of both. The context in which an agent operates may have a large impact on the outcomes of \cA's actions. Disturbances, test distribution shift, equipment variation, and changing dynamics are all common phenomena that alter \cC. \rev{Some of these factors and variables may have been considered during the design and implementation of \cA, while others may not have been. For example, it might have been assumed that the ADT can fit on any segment of a given road network, can clear typical obstacles like potholes, rocks, etc., and can accurately localize itself with its sensors and that the UGS can accurately localize the MG. It may also have been assumed that the hardware for implementing \cA{} remains functional, e.g. the onboard processor will remain thermally and mechanically stable, and has sufficient computing capacity to find accurate solutions and execute timely actions. Such considerations may become significant in edge cases, e.g. operating in hot humid outdoor conditions, bumpy roads, or very large and complex road networks. As such, \cC{} includes the specific road network instantiation and choice of sensors, while $\vec{s}_e$ could include variables like processor status, temperature, humidity, traffic conditions, and so on.} 

\item[Outcome:] An agent acting in a given context might produce many outcomes of interest \cO{}. 
Traditionally such outcomes have been monitored to ensure desired performance criteria are met. Such criteria often include: training/test error, response time, steady-state error, cumulative expected reward, objective function value, classification accuracy, alpha/beta error, and numerous others \cite{Norton-ACM-2022}. More recently outcomes of interest have expanded to include fairness, explainability, ethical behavior, and various other more nuanced and precise measures of holistic agent performance \cite{israelsen2024tiered}. As mentioned earlier, these outcome criteria can also be respectively mapped to \emph{what} and \emph{how} competencies. \rev{Examples of outcomes in the autonomous delivery problem could include: time to delivery; number of times the MG intercepted the ADT; number of donuts damaged during delivery; and degree to which the ADT exhibited courtesy to other denizens of the road. } 
\end{description}

With the definitions above, we now precisely define competency (self-)evaluation as illustrated in the bottom portion of Figure~\ref{fig:CompAssessGenBlockDiag}, and \rev{link this process to the delivery example for clarity}:
\begin{description}
\item[Evaluator:] An entity which is responsible for gauging \cA's competency, using whatever information is available about \cA{} and the delegated task at hand. The Evaluator could be a user, system designer, or computational entity (e.g. a separate metacognition module that monitors \cA), or even \cA{} itself. In this view, the Evaluator relies upon on a set of \emph{competency standards}, which specify desirable outcomes and behaviors on the delegated task. \rev{For the delivery problem, the Evaluator could be the delivery dispatch operator, a baked goods business owner, a systems engineer, a city official or municipal software agent in charge of maintaining the road network, et cetera.}

\item[Competency Standard \cSig:] This is a benchmark or baseline, or \emph{set} of these, to which outcomes of interest \cO{} and/or states $S$ will be compared to assess the degree of competency. Competency standards may originate from an individual user, a corporate policy, a design standard/specification, professional/legal organization, etc. A standard is often reliant on the context \cC{} and agent \cA{} itself. \rev{Simple examples of standards that might be included in \cSig{} for the autonomous delivery problem include: an expected time to complete a task (as a scalar mean, or a distribution); a minimum expected reward to be attained during performance of a task; expected rate of successful deliveries; acceptable fraction of damaged delivered donuts; or number of police tickets and complaints received in response to the ADT's behavior.} 

\item[Competency Evaluation:] The generation of a competency statement regarding \cA{} by applying a competency indicator function \cI{} to compare a set of outcomes of interest \cO{} and states against a given set of competence standards \cSig{} under the context \cC{}. Competency evaluation may occur before, during, or after a task is undertaken by an agent \cA{}, \rev{meaning the relevant subset of $S$ could include initial, current, or final states as appropriate}. This can be generalized to multiple indicator functions with corresponding standards. 
\rev{For the delivery problem, a dispatch operator may want to assess competency before the ADT makes a delivery run to decide whether to proceed or delay delivery. Or the operator can assess \cA{} during a delivery run if new information is received about a closed portion of the road network, or assess \cA{} after a delivery to determine whether software or hardware upgrades are required.}

\item[Competency Assessment:] The degree to which an agent, \cA, is capable of achieving a desired set of outcomes and/or behaviors defined by competency standards on a single delegated task, or set of such tasks. Competency may be described in broad terms with regards to a generic set of abilities or qualities, in essence `marginalizing' over some variables \rev{(e.g. `\cA{} allows the ADT to safely navigate the city streets and carry heavy loads in hot humid weather')}, or contextualized to indicate specific capabilities or operational boundaries \rev{(`Using \cA{} in this configuration for this road network, the ADT can safely avoid the MG over a 80 km range with a 200 kg donut payload while driving at 50 kph max, with UGS operating at nominal temperatures between 0-30 deg C')}. 
\end{description}

\rev{Note that in practical settings it may not always be obvious how each of these elements should be interpreted and applied. In this work, we use the ADT as a simplified example of a much more complicated problem, and show that even in such a scenario there are many important and complex questions and interactions that arise.}

\subsection{Strategies for Implementing Algorithmic Competency Self-Assessments} \label{ssec:strategies}

The assumption alignment tracking approach of \cite{Cao-TRO-2023, Gautam-ICRA-2022} is an example instance of the framework in Fig. \ref{fig:CompAssessGenBlockDiag}, whereby multiple binary indicator functions are used to assess whether assumptions essential to \cA{}'s ability to perform the task hold during operation in different contexts \cC{}. 
However, it is generally quite difficult to capture and validate \emph{all} of the assumptions implicit in a given algorithm design for a particular task. The number of assumptions underlying a sophisticated decision making agent performing complex real-world tasks can be staggeringly large. 
Indeed, it is not hard to imagine how a combined set of programs required to check these assumptions could quickly become more computationally complex and expensive to run than the original decision-making algorithm itself. 
Moreover, a designer cannot know with certainty in advance which assumptions are later likely to have a significant impact or enumerate all the possible ways in which they may be violated. 
The violation of some engineering assumptions also may well change the probabilities of success or failure at a given task, without necessarily completely ruling either possibility out. 
For example, the ADT could `get lucky', e.g. if the MG suffers mechanical failures. Decision-making algorithms can also discover systematic loopholes such as `reward hacking' behaviors that deviate from designer intent without necessarily breaking design logic or violating assumptions. 

It is also important to consider how competency self-assessments are communicated to human informees such as supervisors. 
Human supervisors of other (human/non-human) agents typically do not scrutinize intrinsic task assumptions before deciding what tasks to delegate to whom. 
Rather, it is more natural for them to form a theory of mind and reason about the possible evolution of other agents' behaviors, on the basis of attributed percepts, beliefs, goals, intentions, and capabilities \cite{baker2017rational,jara2020naive}. 
The problem of automation surprise in airline pilots, for example, can be explained in terms of mismatches between 
actual state changes observed following commands versus
pilot expectations for how commands \emph{should} change aircraft states (based on simple mental model projections of aircraft and automation behavior) \cite{gelman2014example}. 

With these issues in mind, we develop a complementary approach
to explicit assumption alignment tracking that is driven by a decision-making process model. We suppose that such a process model describes the extent to which the (boundedly rational) supervisor's preferences and expectations for the task map to decision-making outcomes and behaviors -- specified by competency standards -- are likely achievable by (a boundedly rational) \cA{}.  
Focusing on decision-making processes defined via MDPs specifically, one may ask well-posed competency questions such as: how does the use of either an exact or approximate MDP solver impact an otherwise fixed autonomous agent $\cA{}$'s ability to produce desired task outcomes $\cO{}$? How sensitive is that ability to the fidelity of state transition model $T$ or choice of reward function $R$? How is the desired range of $\cO{}$ related to the utility function $V(s)$ that $\cA{}$ tries to optimize? 
These questions can be grounded more concretely for the ADT problem - how often the ADT will get caught or can reach the goal can be equivalently framed in terms of evaluation against corresponding competency standards related to the MDP agent's performance and behavior. For instance, the 
\emph{what} question of whether the ADT reaches the goal in less than $X$ seconds without getting caught can be equivalently framed in terms of overall value or cumulative reward obtained on any given run \rev{using \cA{}}. Likewise, aspects of the policy and policy solver used in \cA{} can be analyzed to address the \emph{how}
question of whether \rev{\cA{} allows} the ADT behaves as expeditiously as possible, e.g. does it attempt maneuvers to evade the MG, given \rev{\cA{}'s} assumed/available knowledge of the world and task? 
Thus, rather than enumerating and validating all a priori design assumptions, the consequences of maintaining/violating a combined set of assumptions (baked into deployed agents \cA{}) are instead made manifest and analyzed. 

A key insight here is that outcomes and competency indicators are generally \emph{random variables} in the MDP framing. These random variables will have a corresponding set of probability distributions determined by the problem and policy structure, as well as other back-end details which determine how $\cA{}$ formulates and solves the problem. 
As such, in a manner akin to goodness of fit evaluation, $\cA{}$ could be programmed to analyze its own MDP-based decision-making process model with respect to competency standard $\Sigma$, and thereby self-assess \emph{what} and \emph{how} competency indicators. 
The margins between $\Sigma$ and \cA{}'s predicted/actual outcomes and behaviors can then be interepreted as `problem-solving statistics' to gauge competency. 
The more \cA{} exceeds (deceeds) competency standards, the stronger (weaker) its corresponding competency indicators should become. 
This concept captures a major feature of the competency evaluation process in the minds of system architects and engineers as they attempt to design, validate, and deploy autonomous agents under uncertainty. Though it takes years of training and experience for human experts to develop mental models that `know what to look for' when addressing these concerns in any particular application, the evaluation pattern of comparing predicted/observed behavior to expected standards fundamentally remains much the same for assessing \emph{designer} confidence in a system  across any problem instance. We seek to imbue this pattern 
in agents via the meta-analytic algorithmic assurance of \emph{machine self-confidence}. 




\section{{Factorized Machine Self-Confidence}} \label{sec:famsec}
The core idea of this work is that 
competency self-assessment for rational autonomous agents can be realized through a hierarchical process of machine self-confidence computation. In particular, for an agent \cA{} acting via probabilistic decision-making algorithms: self-confidence evaluation corresponds to automated reasoning over a set of embedded problem-solving statistics. 
In their raw form, these statistics describe different (but interrelated) aspects of the \cA's expected and observed abilities to achieve particular outcomes \cO{} in given contexts \cC. When assessed collectively in a metacognitive light, these statistics also convey information about the quality of such predictions and thus also about the fitness of \cA's decision-making process for the task and context \cC~ at hand. 

Where do these statistics come from and how are they defined? 
Like all statistics, they derive from quantitative measures that knowledgeable human expert evaluators (e.g. system designers, or policy creators) would use to assess the validity of \emph{operational assumptions} in different tasks and context settings. Formal translation of these measures provides the basis for design of competency indicators \cI{} and standards \cSig{} for self-evaluation and subsequent self-assessment. 
If properly communicated, algorithmic assurances based on competency self-evaluations and self-assessments can provide valuable insights about agent competency to human informees. It is assumed here that informees in general lack detailed technical knowledge about the the inner workings of \cA's decision-making algorithms, but have sufficient knowledge to carry expectations about desired outcomes \cO{} and acceptable agent behaviors, in order to inform the design of \cI{} and choices for \cSig{}. 

This line of reasoning follows the original notion of machine self-confidence described by Hutchins, et al. \cite{Hutchins-HFES-2015}. The major novelty considered in the present work is an automated algorithmic implementation of the machine self-confidence evaluation and reporting process, which accounts for multiple \textit{competency factors} (i.e. indicators \cI) driven by the aforementioned problem-solving statistics. 
Figure~\ref{fig:famsec} provides a block diagram view of a typical MDP-type autonomous algorithmic rational decision-maker $\cA{}$, which: represents its task, environment, and own state through a model; uses the model within a solver to devise a plan for enacting decisions that align with parsed user intentions (e.g. represented by an objective function for optimization); has the ability to predict, observe, and record possible outcomes, as well as its own and the environment's states; and may acquire user feedback on its performance. This diagram shows where the problem-solving statistics involved with self-confidence assessments are generated and combined into five different competency indicator functions, defined as follows in terms of the questions they address about different components of the mental process model for $\cA{}$: 

\begin{figure}[htbp]
\centering
\includegraphics[width=0.95\columnwidth]{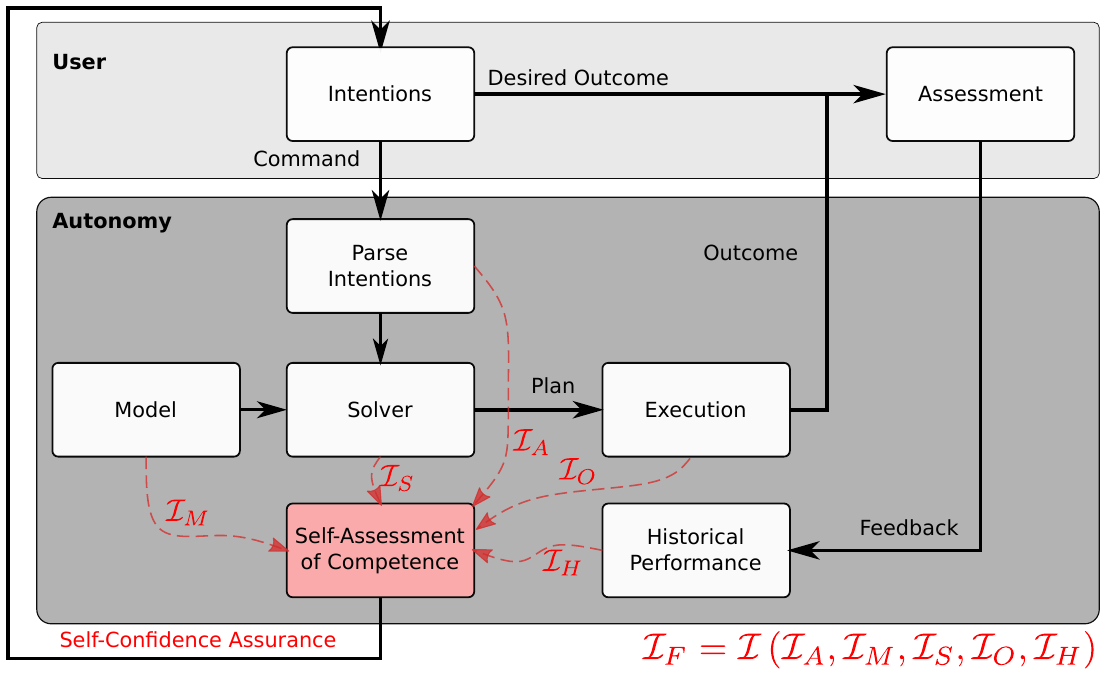}
\caption{Block diagram showing how \famsec{} indicator functions (red) relate to components (white boxes) of a typical rational decision-making algorithmic agent $\cA{}$ (dark grey box) interacting with user (light grey box). }
\label{fig:famsec}
\end{figure}

%

\begin{enumerate}
\item \IA---\textit{\textbf{alignment with user intent}}: To what extent were the user's intentions properly understood and translated by $\cA{}$ into context-appropriate mission specifications and tasks? This factor derives from features and parameters of the `Parse Intentions' block. For instance, if a natural language interface is used for mission planning, this factor could assess how well \cA{} believes user inputs are correctly mapped to reward functions for different mission profiles. Competency standards $\Sigma$ here would thus reflect sufficient benchmarks for demonstrable understanding of user intent and objective function alignment, e.g. in terms of ranking specific state trajectories and actions in accordance with user preferences \cite{sanneman2023transparent}. 
\item \IM---\textit{\textbf{model and data validity}}: Are $\cA{}$'s learned and/or assumed models used for decision-making good enough proxies for the real world? This factor assesses how well the set of measurements and events predicted by the autonomous system line up with what it actually observes in reality. In this case $\Sigma$ would therefore generally reflect thresholds for goodness of fit between models and \rev{data observed at test time, or in other words during a task.}
\item \IS---\textit{\textbf{solver quality}}: Are the approximations and learning-based adaptations used by $\cA{}$ for solving decision-making problems appropriate for the given task and model? For MDPs, this translates to evaluating the quality of a particular policy. Since approximations are almost always needed to solve otherwise intractable decision making problems, this factor examines the extent to which such approximations are appropriate. This factor also accounts for the impact of any learning mechanisms required to make complex decisions under uncertainty, e.g. based on suitability of training data or the learning process to solving the problem at hand \footnote{\rev{This is distinct from the role of \IM{} -- \IM{} assesses the assumption that \cA’s model is sufficient, whereas \IS{} assesses the sufficiency of available data for the solver while assuming \cA{}'s model is sufficient.}}. 
As will be elaborated in Sec. \ref{sec:comparing_solvers}, $\Sigma$ here represents a minimum acceptable level of quality with respect to some notion of `solution goodness', akin to a required convergence tolerance for numerical optimization or estimate confidence for a point estimation algorithm. 
\item \IO---\textit{\textbf{expected outcome assessment}}: Do the sets of possible events, rewards, costs, utilities, etc. for a particular decision lead to desirable outcomes? Even if $\cA{}$ perfectly understands and analyzes a task, and can arrive at globally optimal solutions, it may still not be able to always avoid running into undesirable states along the way. This factor evaluates the particular decision-making strategy implemented by the system to assess the inherent favorability of the full landscape of possible task outcomes. 
Here $\Sigma$ represents the acceptable set of ranges for outcomes of interest, which -- as discussed in Sec. \ref{ssec:goingmeta} -- may go above and beyond merely looking at utility or reward functions. 
%
\item \IH---\textit{\textbf{historical performance and experience}}: What can be gleaned from the system's own actual experiences and other available historical information for performing similar tasks? This factor notionally allows the autonomous system to predict, transfer, and update assessments of self-confidence based on prior experiences, and thus embodies meta-memory and meta-learning for enabling and improving self-assessments, as well as knowledge transfer between agents performing the same/similar tasks. As such, this indicator can be relevant whether or not the algorithms used by $\cA{}$ explicitly use their own built-in learning mechanisms. Generally $\Sigma$ here represents thresholds defining the extent to which various kinds of past experiences ought to inform current/future task capabilities.
\end{enumerate}

Since each indicator \cI can be thought of as one of many `factors' comprising an overall competency assessment indicator $\cI_{F}$, this framework is dubbed \emph{Factorized Machine Self-Confidence}, or \famsec.  
Here, $\cI_{F}$ could be either a concatenated set of the indicator outputs or the output of a function which composes these into a fused summary assessment. 
Unlike the approach originally devised by Hutchins, et al., competency evaluations and assessments produced by \famsec{} do not have to be manually prescribed each time under different task contexts \cC{}. Rather, 
following Fig. \ref{fig:CompAssessGenBlockDiag},
the various indicators \cI{} are treated explicitly and automatically by \famsec{} as functions of compiled statistics describing \cO{} and of the states \cS{} of \cA{} and the environment, with respect to \emph{user-defined} competency standards \cSig{}. The automatability of \famsec thus rests on the  commonality of \IA, \IM, \IS, \IO, and \IH{} to a broad class of (probabilistic) decision-making algorithms like MDPs that are capable of furnishing relevant statistics behind each factor.  Note in Fig. 2 that the general expression for an indicator is $\mathcal{I}(\cdot|\cC{},\cA{},\Sigma)$; in the remainder of the paper, we assume each of the indicators is evaluated for a particular \cC{}, \cA{}, and $\Sigma$ configuration, and therefore suppress the explicit dependency on these variables when referring to the individual indicators.  

By formally dissecting a goal-oriented algorithmic decision making process, \famsec{} provides a grounded starting point for automatically assembling meta-reasoning indices into human-understandable competency self-assessments. 
\rev{Taken together, the competency indicators provide insight into both \emph{what} outcomes the agent can achieve and \emph{how} the agent achieves those outcomes. Note that in our framing, the \emph{what} aspects of competency here are covered by the \IO{} factor and can include behavioral indicators that relate to the `causal manner' in which other outcomes are achieved (how the agent actually performs its task, e.g. the agent avoids putting the ADT into specific areas of the road network, or maintains a distance of at least 4 nodes between the ADT and the MG, etc.). 
The remaining \emph{how} aspects of competency are covered by the remaining four factors to assess the algorithmic components that generate artifacts used to examine the \emph{what} aspects} \footnote{\rev{The distinction is non-exclusive, as factors like \IS{} also open the door to higher-level \emph{what} aspects of competency, e.g. in addition to analyzing how well a policy solver in configuration $X$ remains near some benchmark utility value for task $Y$ when used under conditions $Z$, the related question of what achievable agent utility limits might exist could also be explored in the same context.}}.

However, \famsec{} is not the only way to do this, and the five indicators mentioned here are not necessarily exhaustive; others could reasonably be postulated. They are also not all applicable in all cases, e.g. \IA{} and \IM{} become irrelevant if a task is perfectly modeled by an MDP with a perfectly user-aligned reward function. And as discussed in Sec. \ref{ssec:otherfactors}, some indicators such as \IA{} and \IH{} are more challenging to concretely formulate than others. Nevertheless, \famsec{} gives a systematic way to identify competency variables and their (asymmetric) functional dependencies for goal-oriented decision making. As detailed later in Sec. \ref{ssec:otherdecalgs}, extensions to \famsec{} could be formulated for rational/MDP-type agents and other kinds of algorithmic decision-makers using suitable interpretations of Fig. \ref{fig:CompAssessGenBlockDiag}.

With these caveats in mind,  three questions must be addressed for \famsec{} to be useful,: 
\begin{enumerate}
\item how should the indicators generally behave under different conditions? 
\item assuming they are computable, how should the indicators be calculated and communicated?
\item how can a competency self-assessment result based on the indicators be validated?
\end{enumerate}

The next subsection examines question (1) through different hypothetical  variations of $\cA{}$ and contexts $\cC{}$ in the ADT problem. \rev{The main takeaway of these vignettes (and future examples to be shown later in the paper) is to illustrate and emphasize the varying degrees to which each indicator can potentially capture sensitivities to different aspects of task and environment contexts.} 
Addressing question (2), Secs. \ref{sec:outcomeassess} and \ref{sec:solverquality} respectively describe in detail how two particular indicators -- $\IO$ and $\IS$ -- can be tractably determined for MDP (and related) agents \rev{via accessible problem-solving statistics}. Section \ref{sec:otherfactpract} addresses ongoing work and prospects for answering question (2) for the remaining indicators and also takes up question (3) generally.

\subsection{Expected Properties of \famsec{} Indicators} 

The left column of Table~\ref{tab:boundaryConditions} describes example hypothetical variations of the nominal ADT delivery problem in the left column, selected to examine expected behaviors for the five indicators. In relation to Fig. \ref{fig:CompAssessGenBlockDiag}, each scenario represents a different type of shift in the task context $\cC{}$ for the ADT agent $\cA{}$; in relation to Fig. \ref{fig:famsec}, each scenario alters one or more of the components of the block diagram defining the inputs to each \famsec{} indicator. 
Here, the outcome of interest $\cO{}$ expresses a conjunction of two conditions: (i) whether the ADT reaches the goal within a specified time, and (ii) whether it does so without getting captured by the MG. This could be expressed as a binary `yes/no' outcome or a number, e.g. accumulated reward or time to reach the goal, which may be assigned an arbitrarily large value if captured. $\cA{}$ does not update its MDP model with each task experience, but can record previous experiences. 
The same nominal discrete-time/discrete-state transition model $T$,  approximate MCTS policy solver, and $R$ are assumed in all cases, per Sec. \ref{ssec:deliveryMDP}. 
The right column of Table~\ref{tab:boundaryConditions} shows the expected behaviors for the most relevant
\famsec{} indicators in each case, using notional values for their fixed ranges and standards $\Sigma$. 
Looking across these scenarios, the basic expected properties of each indicator can be understood. 

In scenario 1, $\IA{}$ takes on a `high' confidence value because it is given (or inferred) that the MDP used by the ADT reflects an exact (or nearly so) understanding of how the supervisor wants the ADT to conduct the task. More generally, a notionally `perfect' $\IA{}$ score implies that the utility function is not only perfectly aligned with user preferences, but that the state transition model $T$, reward model $R$, etc. also perfectly \emph{describe the task the supervisor desires to achieve}, i.e. the MDP realization could be treated as an exact task specification. 
On the other hand, $\IH$ expresses lower confidence in this scenario, since the ADT has not attempted the task previously and thus has no evidence to suggest its actual ability is more/less limited on the basis of its MDP model. 

\begin{table}[t]
\newcolumntype{Y}{>{\centering\arraybackslash}X}
\centering
\caption{Examples of Expected \famsec{} Indicator Function Behaviors }
\label{tab:boundaryConditions}
\footnotesize
\begin{tabularx}{\textwidth}{c Y Y}
\toprule
 \# & Scenario (Context Shift from Nominal)   & Expected Behavior of SC Indicator (Trends/Limiting Behavior)  \\ \midrule
1 & ADT doing task for first time, perfectly aligns with supervisor's behavioral preferences & $\IA \rightarrow higher, \quad \IH \rightarrow lower$ \\[20pt]
2 & $R$ uncertain from supervisor input, system fared poorly on same road network in past & $\IA \rightarrow lower, \quad \IH \rightarrow lower$ \\ [20pt]
3 & Time-space resolution of state transition model $T$ increases (decreases) & $\IM\rightarrow higher \quad (\IM \rightarrow lower$) \\ [20pt]
4 & Number of samples used by sampling-based MCTS solver increases (decreases) & $\IS\rightarrow higher \quad (\IS \rightarrow lower$) \\ [20pt]
5 & Extra constraint to pass through location $X$ before reaching goal, as $X$ gets further from goal and/or acceptable time for task decreases ($\Sigma$ more stringent) & $\IO{} \rightarrow lower$  \\ [20pt]
\end{tabularx}
\end{table}

In scenario 2, the ADT now has some negative prior experiences using the MDP model for a given road network, leading to a lower $\IH{}$ value. Suppose the supervisor revises their state-action preferences to compensate for this on the next delivery attempt. Due to the potential for errors in translating the supervisor's demands into a `correct' MDP (e.g. via use of an imperfect natural language interpreter to derive $R$), the ADT is also now independently less sure that it has an exact task specification, and so $\IA$ is lower to reflect decreased certainty due to a shifting context $\cC{}$. 

Note that in these two scenarios the values of $\IO$, $\IM$, and $\IS$ depend on the nature of the actual task instance itself (driving `what' competencies) and how $\cA{}$ is configured to solve the specified MDP (driving `how' competencies, given that the MDP correctly describes the task). In particular, $\IO$ depends the difficulty of traversing the road network, the distance between the initial ADT location and goal, etc. $\IM$ depends on how well the probabilistic MDP model describes the actual dynamics and uncertainties associated with the delivery task, whereas $\IS$ depends on how well MCTS identifies locally optimal policies with allotted computational resources. Scenarios 3, 4, and 5 provide examples of context shifts that directly impact $\IM$, $\IS$, and $\IO$, respectively, neglecting influence regarding $\IA$ and $\IH$. In scenarios 3 and 4, $\cA{}$'s bounded rationality is the main determinant of competency, whereas the task's increased inherent difficulty and performance demands (i.e. more stringent competency standard $\Sigma$) are the major factors in scenario 5. 

The interplay between variables in these last three context shifts also leads to other interesting side effects. In scenario 3, for instance, increasing the resolution of $T$ means that an MCTS solver may need additional samples and/or time to arrive at a locally optimal approximate policy. This impacts $\IS{}$ depending on whether or not additional computational resources are allocated for maintaining a good policy approximation. Similarly, in scenario 4, the quality of the policy used to assess whether desired task outcomes $\cO{}$ are achievable directly impacts $\IO{}$. 




These examples show how the \famsec{} indicators can reflect typical expert considerations of problem-solving statistics for competency evaluation, and how some indicators can be easier to quantify than others. In the following, focus will be given to what are arguably the two most quantitatively accessible indicators: outcome assessment ($\IO$) and solver quality ($\IS$).  
\rev{We devote the following technical treatment largely to these two factors since they have many possible realizations for MDP agents and provide insights for understanding the remaining factors, which introduce other related challenges. }
Considerations for evaluating the remaining three factors and for assessing the validity of self-confidence evaluations are given in Sec. \ref{sec:otherfactpract}. For simplicity, it is assumed throughout that confidence self-assessments are to be made in the context of \textit{a priori} competency evaluations, i.e. prior to starting the execution of an assigned task instance. In situ and post hoc competency evaluations (during/after task execution) are also considered in Sec. \ref{sec:otherfactpract}, \rev{where we also discuss other aspects of dependencies between the different factors.} 

%

\section{Outcome Assessment}\label{sec:outcomeassess}
Simply stated, evaluating $\IO$ consists in comparing the outcomes $\cO$ produced by agent $\cA$ (where $\cO$ is described by random variables in the MDP setting) to the desired outcomes $\cO^{des}$ for a given task (where $\cO^{des}$ is a portion of the full standard, $\Sigma$, related to outcomes for a given context $\cC$). Given a task, $\cO^{des}$ represents how outcomes of interest \emph{should} look according to the evaluator's utility (or utilities), while $\cO$ are the outcomes expected/actually generated by the agent following its prescribed utility. $\cO^{des}$ and $\cO$ could be formulated in terms of rewards, but generally can be expressed in other ways. 
The main idea of evaluating $\IO$ this way is to assess the value of \emph{assigning a task} to $\cA${} (with its given state-action model, solver, and reward function in the MDP setting). The difference between $\cO^{des}$ and $\cO$ is quantified as another random variable, and the distribution of this random variable is interpreted via $\IO$. 
More probability mass for high/low differences indicates that the $\cA{}$ believes $\cO^{des}$ will be harder/easier to typically achieve per evaluator expectations, resulting in lower/higher confidence in $\IO$. 
A key implication is that built-in utilities like the MDP value function $V(s)$ are insufficient on their own to define $\IO$, although they still reveal useful 
competency information. The concept of a \emph{meta-utility} is thus formally introduced to define $\IO$ indicator functions which address this issue. 
This argument is detailed next, followed by a look at possible meta-utility functions for $\IO$ and application examples. %

\subsection{Going Meta: from Utilities to Meta-Utilities}
\label{ssec:goingmeta}
\subsubsection{Limitations of Analyzing Built-in Utilities}
It is natural to consider $\IO{}$ in terms of $\cA$'s built-in utility, which by construction provides coherent preferences over states, actions, and outcomes to drive rational decision making. 
If for instance the desired outcome, $\cO^{des}$, corresponds to a lower bound value function $V^{min}(s)$ to be achieved at each state $s$ by the MDP value function $V(s)$, 
an evaluator could declare $\cA$ competent at the task (to first order) if $V(s) \geq V^{min}(s) \ \forall s$; the idea is extendable to state-action value functions $Q(s,a)$. 
While conceptually simple, such an approach runs into several problems. 

Firstly, it is quite onerous and impractical to specify/check standards for $V(s)$ or $Q(s,a)$, especially in complex tasks and large problem domains. For $Q(s,a)$ in particular, all states $s$ and actions $a$ would generally have to be accounted for, even if they are unlikely to ever be visited. Secondly, though MDP agents typically define $V(s)$ in terms of average (discounted) cumulative rewards, this definition carries very limited information about the full range of possible task outcomes. 
That is, such a $V(s)$ compresses the full probability distribution of cumulative rewards under a given policy to only the \emph{average outcome}. Examining $V(s)$ by itself thus may say nothing about other important indicators of risk-reward tradeoffs that inform human decision making under uncertainty, such as the spread of outcomes, worst/best possible outcomes, skewness of outcomes, tail risks, etc. under a given model and policy (see next subsection). 
Moreover, by mapping all outcomes into additive rewards/costs, the effects of favorable and unfavorable outcomes are inseparably mixed together in $V(s)$ and $Q(s,a)$, which makes specifying $\cO^{des}$ non-trivial. 
 
Finally, rewards and utilities for rational decision making are well-known to be non-unique \cite{kochenderfer2015decision}; they could be adjusted in an infinite number of ways (e.g. via affine transformations for additive rewards) without affecting agent behaviors and thereby without affecting competency evaluations. 
Indeed, Markov-type rewards are only truly representative of goals/objectives for rational agents under certain special conditions \cite{bowling2023settling}, which do not hold in all cases. 
This again makes it hard to define $\cO^{des}$. In practical terms, even if $\cA$ is well-designed and capable of executing a given task, its built-in utility $V(s)$ is only an inexact best guess approximation of true underlying user/designer utility, and this approximation may yet still need to yield to other approximations (like MCTS) to ultimately admit a computable policy $\pi$. 
The end result is that built-in rewards and utility functions become algorithmic tuning knobs that are iteratively adjusted until $\cA$'s behaviors via $\pi$ align satisfactorily with the designer's (and hopefully a user's) mental model of desired task outcomes. 

However, the evaluator/informee in Figs. \ref{fig:CompAssessGenBlockDiag} and \ref{fig:famsec} is not bound by considerations of computability for $\pi$, since they may take for granted that $\pi$ is already available during competency assessment. 
This suggests they need not rely solely on assessing differences between $V(s)$ and a desired value for it, but can more generally assess gaps between desired outcomes $\cO^{des}$ (per their own mental model) and actual task outcomes $\cO$ produced by $\cA$ acting under its (designed/approximate) utility representation. 

\subsubsection{Meta-utilities}
$\IO$ can be defined in terms of a \emph{meta-utility function} $\metauo = M(\cO,\cO^{des})$ (i.e. separate from $\cA$'s built-in utility $V$), which models evaluator preferences on the \emph{expected differences} between $\cO^{des}$ and $\cO$ produced by $\cA$ acting under a given policy.  
The distinction between $\metauo$ and $V$ is that $\metauo$ is deliberately decoupled from the process of finding $\cA$'s policy, and thus preferences for $\metauo$ can be specified independently of $\cA$'s decision-making process (before or after the fact). In contrast, $V$ is constructively defined and minimized as part of an optimization process which simultaneously produces $\cA$'s policy. The main purpose of constructing $V$ for an MDP is to provide a useful means to an end: a coherent basis for comparing actions and tractably identifying an optimal policy within the confines of bounded rational computation. Otherwise, as discussed earlier the specific values of rewards and of $V$ itself (as an expectation statistic) are not by themselves necessarily meaningful to users, designers, or evaluators in terms of gauging whether $\cA$ \emph{will} achieve $\cO^{des}$. 

Thus, $\metauo$ can be thought of as \emph{the utility of the algorithmic decision-making process (based on the built-in utility $V$) in achieving desired outcomes.} This corresponds very closely to hierarchical utilities used to optimally select among different possible policy/model update strategies in meta-reinforcement learning \cite{grant2018recasting, krueger2017enhancing} and competency-aware planning \cite{Svegliato-IROS-2019,Basich_2020,basich2020learning}. 
A key distinction here in relation to \famsec{} is that $\cA{}$ only uses $\metauo$ to convey self-assessments via $\IO$, rather than close the loop to improve performance. 
By defining $\metauo$ separately from the policy optimization process, an evaluator has total freedom to assign preferences on margins between $\cO^{des}$ and $\cO$ as needed (subject to the usual utility function provisos), to provide an independent figure of merit for $\cA$'s policy as optimized with respect to $V$.  
With this in mind, two practical requirements on defining $\metauo$ 
are imposed. 

Firstly, for $\metauo$ to be an acceptable meta-utility that describes a more/less competent MDP agent $\cA$, improvements to $V$ should also lead to improvements in $\metauo$. 
Stated more formally: ideally $V$ and $\metauo$ ought to be congruent such that, for $V^i$ corresponding to an indexed policy $\pi^i$ and outcome set $\cO^i$ with index $i$, it follows $V^i \geq V^j  \leftrightarrow M(\cO^i,\cO^{des}) \geq M(\cO^j,\cO^{des})$ for policy $\pi^j$, with $i \neq j$. 
This leads to a necessary (but not sufficient) condition for competency assessment via \famsec, whereby (for fixed $\Sigma$ and $\cC$) $\cA$ under $\pi^i$ is `more competent' than $\cA$ under $\pi^j$ only if $M(\cO^i,\cO^{des}) \geq M(\cO^j,\cO^{des})$. 
This requirement permits evaluation of a candidate function $\metauo$ for describing $\IO$ when $\cA$ is already known to be competent for some given $\Sigma$ and $\cC$ (since, for an optimal MDP-based $\cA$, $\metauo$ can be benchmarked against suboptimal variations in $\pi$ and corresponding $V$).  

Secondly, in terms of expressing competency self-assessments via machine self-confidence, the exact choice of $\metauo$ for generating $\IO$ can have significant consequences for human informee interpretability, and thus must also be carefully considered. 
In light of the congruency requirement above, the most trivial and convenient choice would be to set $M=V$, despite what has already been said about the issues of using $V$ by itself. 
However, whereas $\cA$ rationally maximizes its built-in expected utility function, behavioral economic theory and cognitive science have extensively demonstrated that humans often evaluate and make decisions quite differently than purely rational computational agents would with respect to probabilistic event outcomes. $\metauo$ should therefore also be selected so that machine-based reporting and human perceived comparisons of $\cO$ and $\cO^{des}$ are well-aligned. 

Considering both requirements, any one of several frameworks for modeling human decision-making under uncertainty could be used to define $\metauo$. Two possibilities are considered next. In Sec. \ref{sec:solverquality}, meta-utilities are generalized to the formulation of the solver quality \IS{} indicator, where $\cO$ and $\cO^{des}$ are replaced with different variables and standards reflecting \emph{how} \cA{} performs a task. 

\subsection{Candidate Meta-Utilities for Perceived Risk-Reward Tradeoffs}
\subsubsection{Cumulative Prospect Theory}
\begin{figure}[t]
\begin{minipage}{.45\textwidth}
	\begin{center}
	\includegraphics[width=50mm]{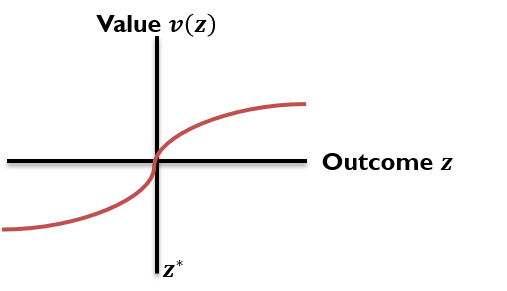}
    \end{center}
	\subcaption[Typical Value Function]{Typical Value Function}
	\label{fig:valuefunc}
\end{minipage}
\begin{minipage}{.45\textwidth}
	\begin{center}
	\includegraphics[width=75mm]{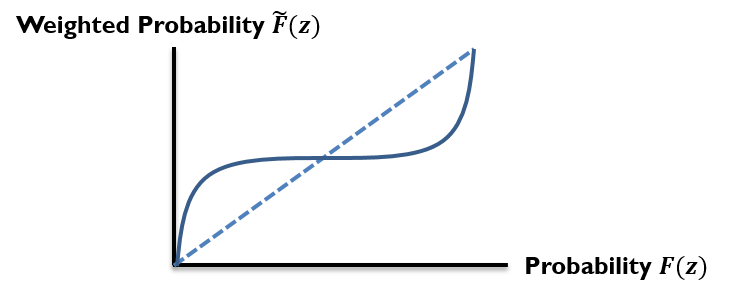}
    \end{center}
  	\subcaption[Typical Weighting Function]{A Typical Weighting Function}
	\label{fig:weightfunc}
\end{minipage}
\caption{CPT models of value and probability functions for perceived risk-reward tradeoffs.}
\end{figure}

Tversky and Kahneman \cite{tversky1992advances} famously proposed Cumulative Prospect Theory (CPT) to describe how humans assess uncertain outcomes. There are three main differences between CPT and the classical Expected Utility Theory (EUT) used by most algorithmic agents $\cA$. 
The first difference is that humans evaluate outcomes in terms of marginal gains/losses relative to a reference outcome point, rather than solely with respect to final outcome magnitudes. The second difference is that humans weight losses differently than they do gains. 
A simple example of this is loss aversion, or the tendency for humans to weight losses higher than they weight gains. 
The final difference is that human's tend to overweight the probability of extreme outcomes, and underweight the probability of average outcomes. CPT combines these differences into a modified version of the utility function for classical EUT,
\begin{align}
\label{eqn:cpt}
M_{\cO}=M(f(z),\left\{l^{-},g^{+} \right\}) := \int_{-\infty}^{l^-} v(z)\frac{d}{dz}(w(F(z)))dz + \int_{g^+}^{+\infty}v(z)\frac{d}{dz}(-w(1-F(z)))dz,
\end{align}
\noindent where $f(z)$ is the ordered distribution (probability density) of all outcomes, $z$ is an outcome realization for random variable $\mathcal{Z} \in \cO$, and $F(z)$ is the cumulative probability distribution of all outcomes up to $z$. 
For instance, in the ADT problem, $\mathcal{Z}$ could be the non-discounted cumulative reward outcome
\begin{align}
\rwdapprox(T)=\sum_{k=0}^{T}R_k, \label{eq:rtilde}
\end{align}
obtained by an MDP agent that executes $T$ steps under a given policy. 
The function $v$ maps outcomes $z$ to values of perceived loss or gain, and the weighting function $w$ maps cumulative probabilities over outcomes $z$ to subjective cumulative probabilities $\tilde{F}(z)$. Example $v$ and $w$ are shown in Figures \ref{fig:valuefunc} and \ref{fig:weightfunc}. 
In this case, the competency standard $\Sigma$ for $\cO^{des}$ is defined by the outcome bounds $l^{-},g^{+} \in \mathcal{Z}$. 
Note that the first integral term on the right-hand side of Eq. (\ref{eqn:cpt}) accounts for total perceived losses below the nominal target outcome loss upper bound of $z=l^-$, whereas the second integral term accounts for nominal gains above the nominal target lower bound gain of $z=g^+$. Although $l^- = g^+ = 0$ is typically stipulated in descriptions of CPT, the bounds need not be the same or zero in general, depending on the construction of $z$ and $v(z)$.  
If $v(z) = z$ for all $z$, and $w(F(z)) = F(z)$ for all $F(z)$, then (\ref{eqn:cpt}) simplifies to $\mathrm{E}[\mathcal{Z}]$ as in EUT. 
The key takeaway here, however, is that (\ref{eqn:cpt}) generally contains more information about the overall distribution shape for outcomes and outcome values, rather than just the expected value of outcome values. 

\subsubsection{Upper/Lower Partial Moments}
Eq. (\ref{eqn:cpt}) offers an attractive family of candidate meta-utilities to define $\IO$; it has also attracted much attention in the human-machine interaction community as a basis for explainable and interpretable decision-making \cite{kwon2020humans, geng2020prospect}. Yet, it is not straightforward to identify $v(z)$ and $w$ for individual users, or to precisely obtain $F(z)$ in general settings. To mitigate these issues, one may turn instead to simpler alternative non-parametric `CPT-compatible' families of utility functions developed in fields like econometrics and mathematical finance, which also deal extensively with human decision-making under uncertainty. These utilities are non-parametric in the sense that they do not make restrictive distributional assumptions for $F(z)$ (e.g. they are applicable to empirical distribution data), and CPT-compatible because they include CPT utilities as a special case. 
One particularly simple candidate family of such utilities is given by the $\alpha$-upper/lower partial moment ($\alpha$-UPM/LPM) ratio \cite{cumova2014portfolio, viole2016predicting}, 
\begin{align}
M_{\cO} = M(f(z),\left\{l^{-},g^{+} \right\}; \alpha):=\alpha\mbox{-UPM/LPM} = \frac{\int_{g^+}^{+\infty}(z-g^+)^{\alpha}{f}(z)dz}{\int_{-\infty}^{l^-} (l^{-}-z)^{\alpha}{f}({z})d{z}},
\label{eqn:upm/lpm}
\end{align}
for integer parameter $\alpha \geq 0$ and continuous outcomes $z$ with probability density function $f(z) = \frac{d}{dz}F(z)$. 
The $\alpha$-UPM/LPM ratio expresses the ratio between favorable and unfavorable outcome event moments, which are defined by expectations of $\alpha$ moments relative to reference outcome values. 
For $\alpha=0$, (\ref{eqn:upm/lpm}) simply becomes the likelihood ratio under $f(z)$ for the probability of $z$ exceeding $g^+$ to the probability of $z$ deceeding $l^{-}$. For $\alpha \geq 1$, the ratio describes how the shape of $f(z)$ contributes to the distribution of $\mathcal{Z}=z$ above $g^+$ and below $l^-$.  
If $\alpha=1$, the numerator and denominator respectively give the upper and lower semi-mean of $z$ with respect to $g^+$ and $l^-$; whereas
for $\alpha=2$, these respectively give the upper and lower semi-variance of $z$ with respect to $g^+$ and $l^-$, etc. For any $\alpha$, the ratio approaches $0$ as the lower moment dominates; goes to $\infty$ as the upper moment dominates; and is $1$ when the partial moments balance. 

The $\alpha-$UPM/LPM can be generalized to a form which reproduces typical CPT utilities as special cases \footnote{This requires a different moment degree $\alpha$ in the numerator and $\beta$ denominator, yielding a result with potentially mixed `units'; consideration is limited here to $\alpha=\beta$ for `apples-to-apples' simplicity.} and which provides favorable guarantees for decision-making under uncertainty via stochastic dominance properties \cite{cumova2014portfolio}. 
As such, the authors of \cite{cumova2014portfolio,viole2016predicting} advocate for UPM/LPM-based metrics to evaluate the goodness of asset return distributions in stochastic financial portfolio analysis, in terms of each asset's expected ability to maintain desired minimum gains and losses. 
The key property of UPM/LPM metrics in this regard is the ability to account for the full distributional shapes of separated gains and losses in relation to desired performance targets, particularly when non-ideal outcome distribution features (long-tails, skewness, etc.) are present. Typical measures like the total distribution mean and variance (provided they exist) overlook such information and do not separate gains from losses.
As applied to competency evaluation, $\alpha$-UPM/LPM can capture information about the distribution of $\mathcal{Z} \in \cO$ achieved by $\cA$ under its built-in utility, relative to reference values determined by $\cO^{des}$. 
Though UPM/LPM metrics lead to non-trivial computation issues as objective functions for optimal decision making \cite{cumova2014portfolio}, they are easy to implement for plug-in analysis and monitoring of decision outcome data when a policy already exists. 
As such, UPM/LPM metrics are well-suited for gauging margins between $\cO$ and $\cO^{des}$, while accounting for uncertainty under context $\cC$. 
 
\begin{figure}[t]
    \begin{center}
	\includegraphics[width=0.5\textwidth]{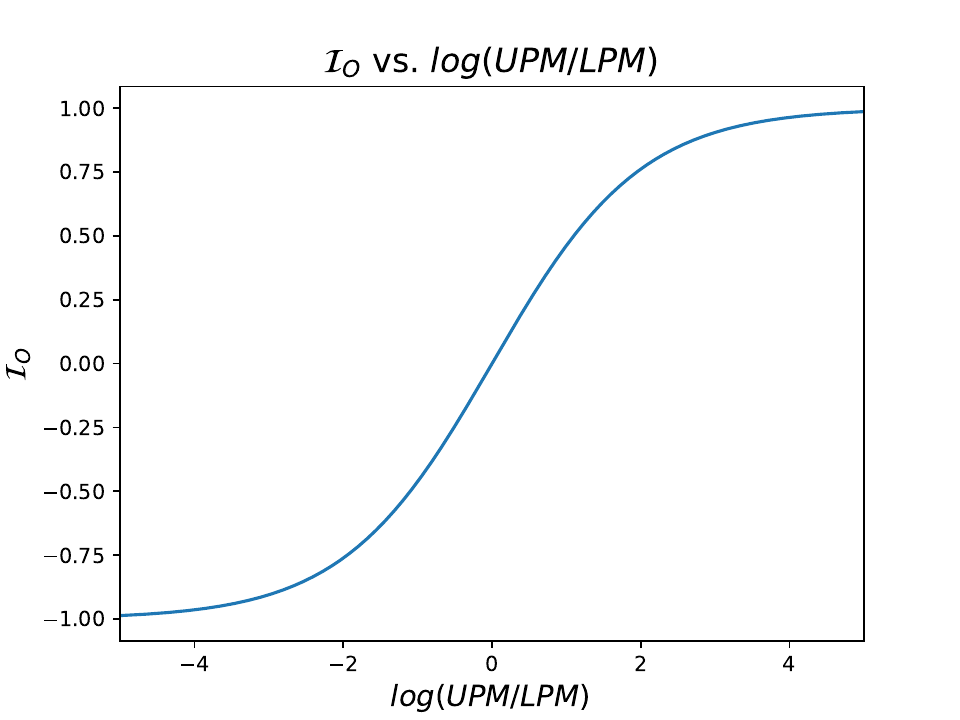}
    \end{center}
  	\caption{The $\IO$ Outcome Assessment factor curve corresponding to Eq. (\ref{eqn:oa}) .}
\label{fig:oa_curve}
\end{figure}

A general recipe for constructing $M(\cO,\cO^{des})$ from eq. (\ref{eqn:upm/lpm}) to generate $\IO$ is described next, followed by a specific algorithmic implementation for competency self-assessment in MDP-based agents under different choices for $\cO^{des}$. 
First, assume that random variable $\mathcal{Z}$ describes $\cO$ with realizations $z$ and pdf $p(z)$, with ordering on $z$ such that $z_1 > z_2$ implies that the outcomes $\cO_1$ described by $z_1$ are strictly preferable to outcomes $\cO_2$ described by $z_2$. $\mathcal{Z}$ could be any other figure of merit for task performance and agent behavior in general, e.g. the non-discounted cumulative reward obtained by an MDP-based $\cA$ in cases where the instantaneous reward function $R$ for built-in value function utilities $V$ meaningfully reflect user preferences for actions and task states
\footnote{A scalar $z$ is assumed here without loss of generality; if $\cO$ consists of a number of different outcome types, then each outcome type $i$ is ascribed its own distinct $z^i$; this case will be discussed later in generalized $\IO$ implementations.}. Next, let $g^+$, and $l^-$ define the competency standard $\Sigma$ for $\cO^{des}$, where $g^+$ determines the minimum acceptable gain $z$ and $l^-$ the maximum acceptable loss. For simplicity, it is assumed hereafter that $g^+=l^-= z^*$,  i.e. (un)favorable outcomes correspond to strictly (negative) positive values of $z-z^*$. 
As mentioned earlier, the outcome pdf $f(z) = \frac{d}{dz}F(z)$ may not be straightforward to obtain exactly in general. Assume that $f(z)$ can be estimated to an arbitrarily accurate degree by $\cA$ or the evaluator via simulated policy execution, e.g. via exact Monte Carlo sampling, resulting in an estimated pdf $\tilde{f}(z)$. 
Combining these assumptions into (\ref{eqn:upm/lpm}) for $\alpha=1$ yields the empirical UPM/LPM, 
\begin{align} 
    M_{\cO} = M(\tilde{f}(z),z^{*}; \alpha=1):=\mbox{UPM/LPM} = \frac{\int_{z^*}^{+\infty}(z-z^{*}){\tilde{f}}(z)dz}{\int_{-\infty}^{z^*} (z^{*}-z){\tilde{f}}({z})d{z}}. \label{eq:empUPMLPM}
\end{align}
Though any $\alpha \geq 0$ can be used to characterize different features of $\tilde{f}(z)$, $\alpha=1$ is particularly desirable and used hereafter, since it produces summary statistics that are intuitively understood as average excess outcome gains and losses relative to a minimally acceptable outcome standard $z^*$. 
The interpretation of (\ref{eq:empUPMLPM}) for $\alpha=0$ as the odds ratio for favorable to unfavorable outcomes (also called the Omega ratio \cite{kapsos2014optimizing}) is also arguably intuitive and indeed useful in special cases involving binary outcomes (discussed later). However, using $\alpha=0$ is coarse-grained in general as it does not indicate the expected \emph{degree} to which $\cA$ is (in)capable of achieving $z^*$. 
Statistics produced for $\alpha \geq 2$ as higher order semi-moments may be technically useful to system designers, but are more abstract and would likely require greater effort by non-technical users/stakeholders to interpret. 
To aid with interpretability, eq. (\ref{eq:empUPMLPM}) can be further transformed into a bounded scale, e.g. using a simple monotonic logistic-log function such as 
\begin{align}
\IO = \frac{2}{1+e^{-k \log{\mbox{\tiny (UPM/LPM)}}}} - 1 
= \frac{1-(\mbox{UPM/LPM})^{-k}}{1+(\mbox{UPM/LPM})^{-k}} = \frac{\mbox{UPM}^k - \mbox{LPM}^k}{\mbox{UPM}^k + \mbox{LPM}^k}
\label{eqn:oa}
\end{align}
which is shown in Fig. \ref{fig:oa_curve} for $k=1$, where the parameter $k>0$ tunes the steepness of the curve around UPM/LPM$=1$.
This $\IO$ assessment maps outcome distributions to a number between -1 and 1, representing the `goodness' of the distribution for $\cO$ (represented by $\hat{f}(z)$) relative to $\cO^{des}$ (represented by $z^*$).
In the context of competency self-evaluation, when $\IO=-1$, $\cA$ is certain it will grossly underperform relative to $z^*$, and thus projects total lack of self-confidence in achieving $\cO^{des}$ on any given task instance. When $\IO=1$, $\cA$ is certain it will vastly overperform relative to $z^*$ and thus projects total self-confidence in achieving $\cO^{des}$. For any $\IO$ between these two extremes, $\cA$ projects graded self-confidence. Mid-range positive/negative values of $\IO$ imply net gains/losses relative to $z^*$ are more likely to result for an average task execution, whereas $\IO=0$ implies no net loss/gain on average. Therefore, for a fixed $z^*$, the more positive (negative) $\IO$ becomes, the more (less) competent $\cA$ is predicted to be at the task. If $\cA$ is instead held fixed while $z^*$ is varied, then $\IO$ naturally tends to increase (decrease) as $z^*$ decreases (increases), corresponding to easier (harder) task outcome standards for the same outcome distribution produced by $\cA$. 

\subsection{Application Examples}
\subsubsection{Synthetic $\IO$ Evaluations}
\begin{figure}[t]
\centering
     \begin{subfigure}[t]{0.3\textwidth}
         \centering
         \includegraphics[width=\textwidth]{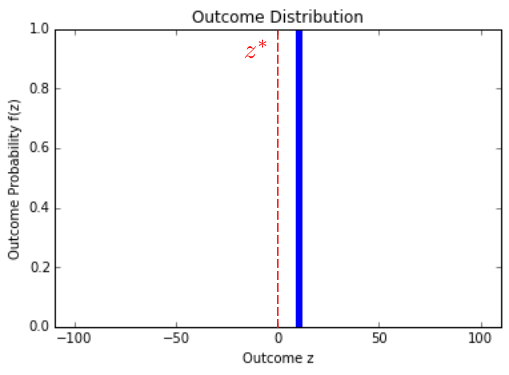}
         \caption{ $\IO=1$ }
         \label{fig:exout1}
     \end{subfigure}
     \hfill
     \begin{subfigure}[t]{0.3\textwidth}
         \centering
         \includegraphics[width=\textwidth]{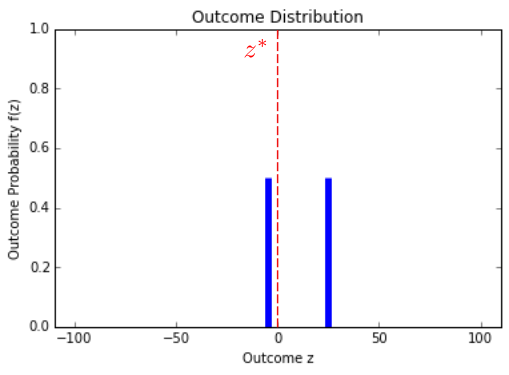}
         \caption{$\IO=\frac{2}{3}$}
         \label{fig:exout2}
     \end{subfigure}
     \hfill
     \begin{subfigure}[t]{0.3\textwidth}
         \centering
         \includegraphics[width=\textwidth]{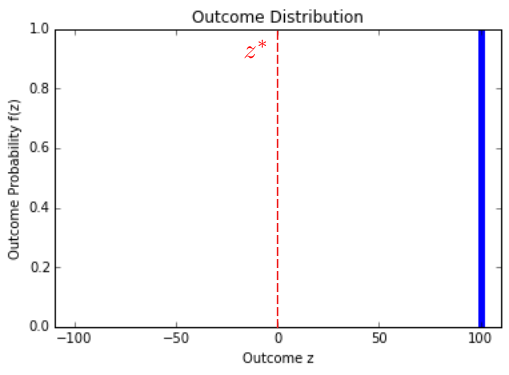}
         \caption{$\IO=1$}
         \label{fig:exout3}
     \end{subfigure}
\newline
\centering
     \begin{subfigure}[t]{0.3\textwidth}
         \centering
         \includegraphics[width=\textwidth]{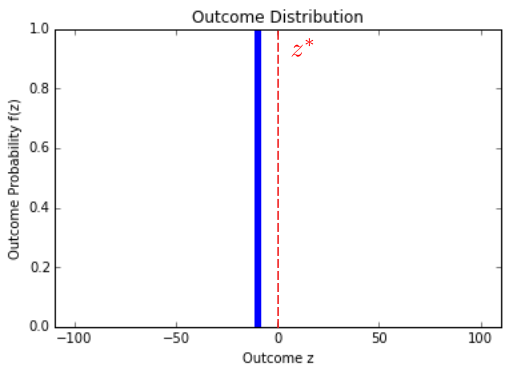}
         \caption{$\IO=-1$}
         \label{fig:exout4}
     \end{subfigure}
     \hfill
     \begin{subfigure}[t]{0.3\textwidth}
         \centering
         \includegraphics[width=\textwidth]{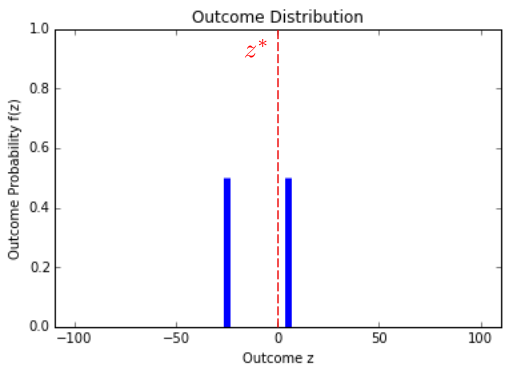}
         \caption{$\IO=-\frac{2}{3}$}
         \label{fig:exout5}
     \end{subfigure}
     \hfill
     \begin{subfigure}[t]{0.3\textwidth}
         \centering
         \includegraphics[width=\textwidth]{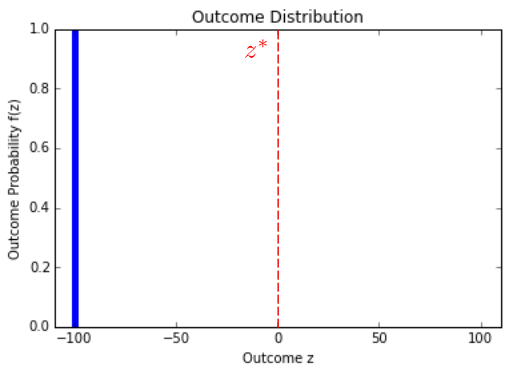}
         \caption{$\IO=-1$}
         \label{fig:exout6}
     \end{subfigure}
\newline
     \begin{subfigure}[t]{0.3\textwidth}
         \centering
         \includegraphics[width=\textwidth]{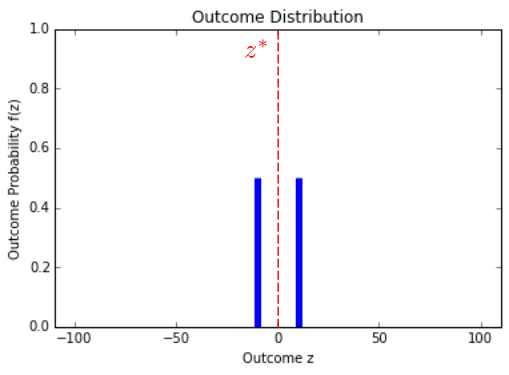}
         \caption{$\IO=0$}
         \label{fig:exout7}
     \end{subfigure}
     \hfill
     \begin{subfigure}[t]{0.3\textwidth}
         \centering
         \includegraphics[width=\textwidth]{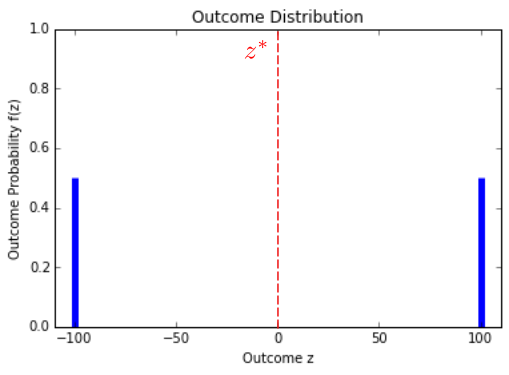}
         \caption{$\IO=0$}
         \label{fig:exout8}
     \end{subfigure}
     \hfill
     \begin{subfigure}[t]{0.3\textwidth}
         \centering
         \includegraphics[width=\textwidth]{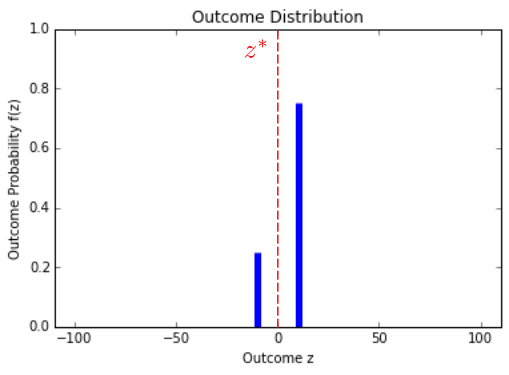}
         \caption{$\IO=\frac{1}{2}$}
         \label{fig:exout9}
     \end{subfigure}
\newline
     \centering
     \begin{subfigure}[t]{0.3\textwidth}
         \centering
         \includegraphics[width=\textwidth]{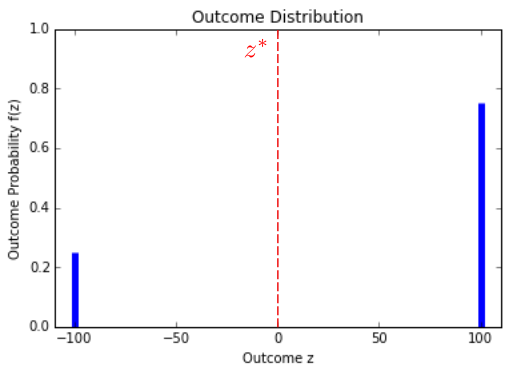}
         \caption{$\IO=\frac{1}{2}$}
         \label{fig:exout10}
     \end{subfigure}
     \hfill
     \begin{subfigure}[t]{0.3\textwidth}
         \centering
         \includegraphics[width=\textwidth]{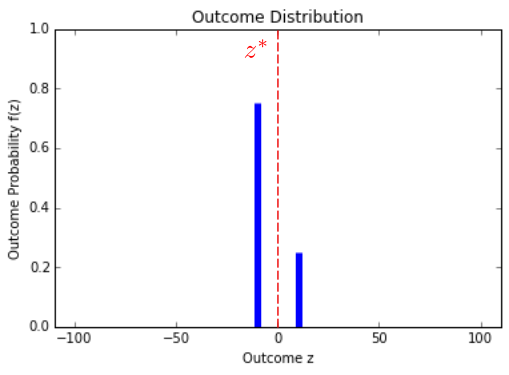}
         \caption{$\IO=-\frac{1}{2}$}
         \label{fig:exout11}
     \end{subfigure}
     \hfill
     \begin{subfigure}[t]{0.3\textwidth}
         \centering
         \includegraphics[width=\textwidth]{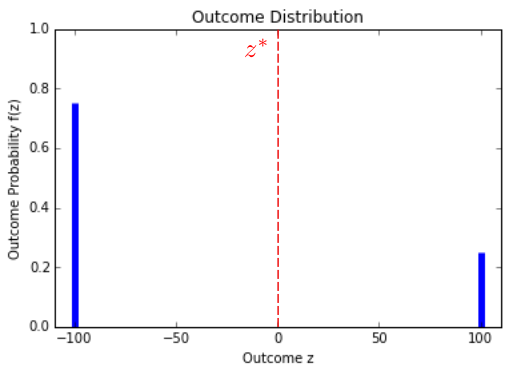}
         \caption{$\IO=-\frac{1}{2}$}
         \label{fig:exout12}
     \end{subfigure}
        \caption{ $\IO$ behavior for various synthetic outcome distributions, with $z^*=0$ for all cases.}
        \label{fig:exout1to12}
\end{figure}

Fig. \ref{fig:exout1to12} (a)-(l) demonstrate the basic properties of the UPM/LPM $\IO$ function, in particular how the UPM/LPM ratio with $\alpha=1$ can balance optimism and pessimism relative to $z^*$ in a `first order' sense. Each pdf $\hat{f}(z)$ in this example consists of mixtures of Dirac delta functions on an arbitrary scalar $\mathcal{Z}$, so that all probability mass is placed on one or two specific outcomes $z$ to simplify the partial moment calculations in (\ref{eq:empUPMLPM}). The resulting $\IO$ result from eq. (\ref{eqn:oa}) is shown below each graph for $z^*=0$. 

Cases (a)-(c) result in similarly high positive $\IO$ values, despite significant differences between the underlying $\hat{f}(z)$ pdfs. In (a) and (c), all probability mass for $z$ lies to the right of $z^*=0$, which drives the UPM/LPM to infinity and makes $\IO=1$, indifferent to the magnitude of the positive outcomes $z$. The interpretation for very high self-confidence in these cases is that $\cA$ is certain it can exceed $z^*$. In (b), although the probability for $z=-5$ is the same as for $z=25$, the greater moment arm in the UPM drives the UPM/LPM ratio to 5, so $\IO \approx \frac{2}{3}$. Here $\cA$ predicts that it is fairly likely - but not entirely certain - it will exceed $z^*$. The Omega ratio is 1 in this case, indicating equal likelihood of either exceeding or falling below $z^*$; this contrasts with $\IO$, which reflects competency in terms of both the degree and likelihood of exceeding $z^*$. 
Cases (d)-(f) are mirror opposites to (a)-(c), showing certainty that $z$ will \emph{not} exceed $z^*$ in (d) and (f) (very low self-confidence), and relatively low self-confidence that $z$ will exceed $z^*$ in (e). 

Cases (g) and (h) show equally likelihood to achieve outcomes of equal but opposite magnitude, so $\IO=0$ reflects fair/neutral self-confidence. Cases (i)-(j) and (k)-(l) show instances where $\cA$ achieves the same positive and negative $\IO$, respectively, for different pairs of $\hat{f}(z)$. These examples show a limitation of the $\alpha=1$ UPM/LPM ratio: ratios for closely spaced positive/negative outcomes cannot be distinguished from those that are symmetrically more widely spaced when outcome probabilities are held fixed. Similar issues arise when the outcome magnitudes and their probabilities are altered in opposite proportions. 
For example, a moderately positive value of $\IO$ could imply: a likely small set of gains with highly unlikely large losses; a very large unlikely gain with a likely set of low-magnitude losses; or something between these extremes. 
Strategies for addressing this limitation via generalizations of $\IO$ will be considered in Sec. \ref{ssec:iogenextensions}.

\subsubsection{Illustration for ADT Problem}
Consider an MDP agent $\cA$ for the ADT problem that uses an exact policy $\pi$ produced via value iteration for an infinite horizon planning scenario with fixed discount factor $\gamma \in [0,1)$, using the nominal reward function $R$ defined earlier. Suppose the only outcome of interest is $\mathcal{Z}=$\rwdapprox$(T)$  (total non-discounted task reward collected by $\cA$  surviving $T$ steps on a given task instance). 
Recall from the definition of $R$ that $\cA$ obtains: a large penalty when the ADT is caught by the MG; a large reward when the ADT reaches the goal $R_{\mbox{\tiny goal}}$; and a small loitering penalty $R_{\mbox{{\tiny loiter}}}$ for each time step the ADT is not at the goal and not caught. The choice of $z^*=0$ results in a competency standard corresponding to a `break even' total non-discounted reward, i.e. $\cA$ is competent at the task if -- at a minimum -- it can reach the goal in no more than $\frac{R_{\mbox{\tiny goal}}}{|R_{\mbox{{\tiny loiter}}}|}$ time steps without getting caught.  

\begin{figure}[tb]
    \begin{center}
	\includegraphics[width=0.75\textwidth]{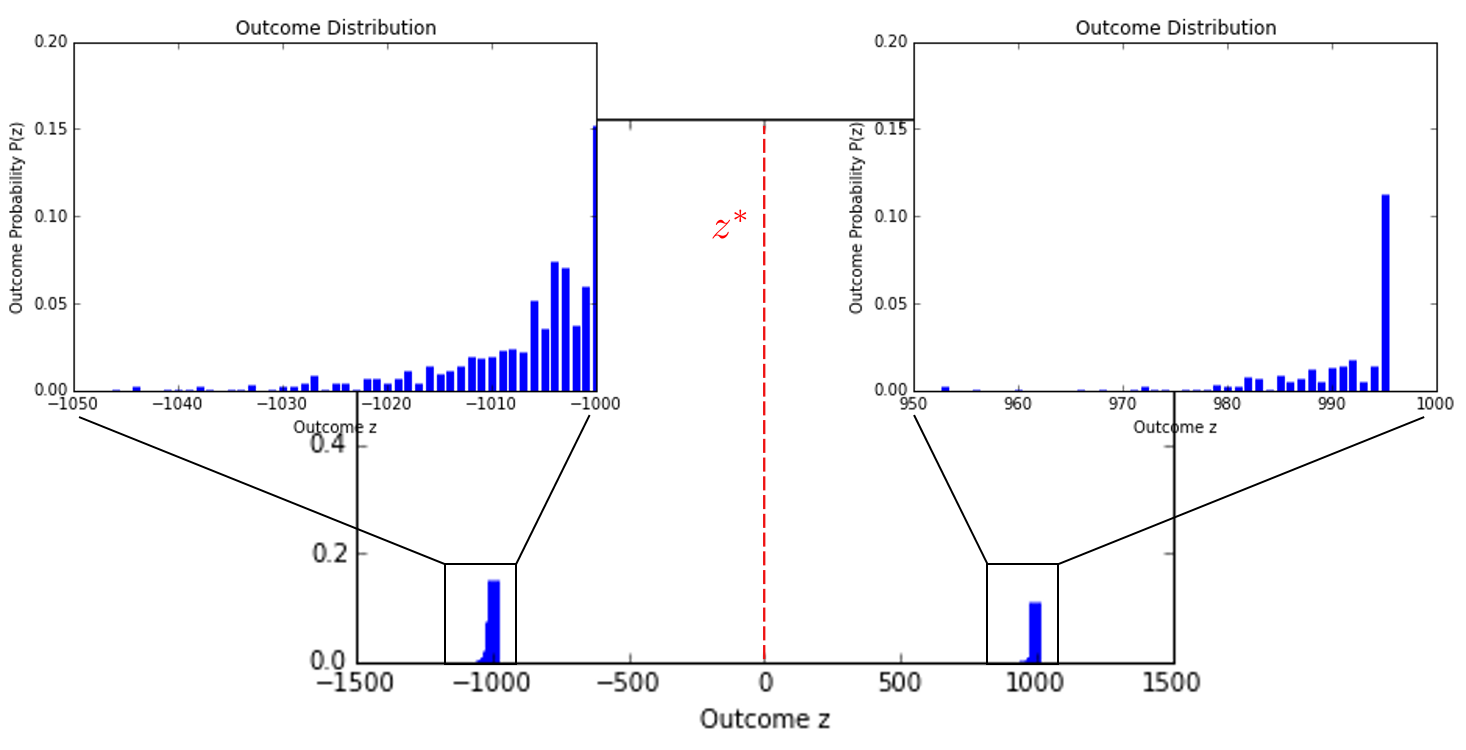}
    \end{center}
  	\caption{Typical $\hat{f}(z)$ for $\mathcal{Z}$ defined as total non-discounted reward collected by MDP agent in ADT problem.}
\label{fig:exprdist}
\end{figure}

Figure \ref{fig:exprdist} shows an example histogram of $\hat{f}(z)$ for an optimal policy on a typical problem instance for 10,000 Monte Carlo runs. The bimodality of the outcomes in this sample plot indicates that (for the same environment and starting conditions) $\cA$ generally succeeds at guiding the ADT safely to the goal slightly less often than it gets caught by the MG. This results in $\IO<0$ for $z^*=0$ reflecting underconfidence in $\cA$. Since $\pi$ is an optimal MDP policy for this environment and task, this result implies that the environment and starting conditions make it intrinsically difficult for $\cA$ to perform up to the competency standard. 

\begin{figure}[tb]
    \begin{center}
\includegraphics[width=0.85\textwidth]{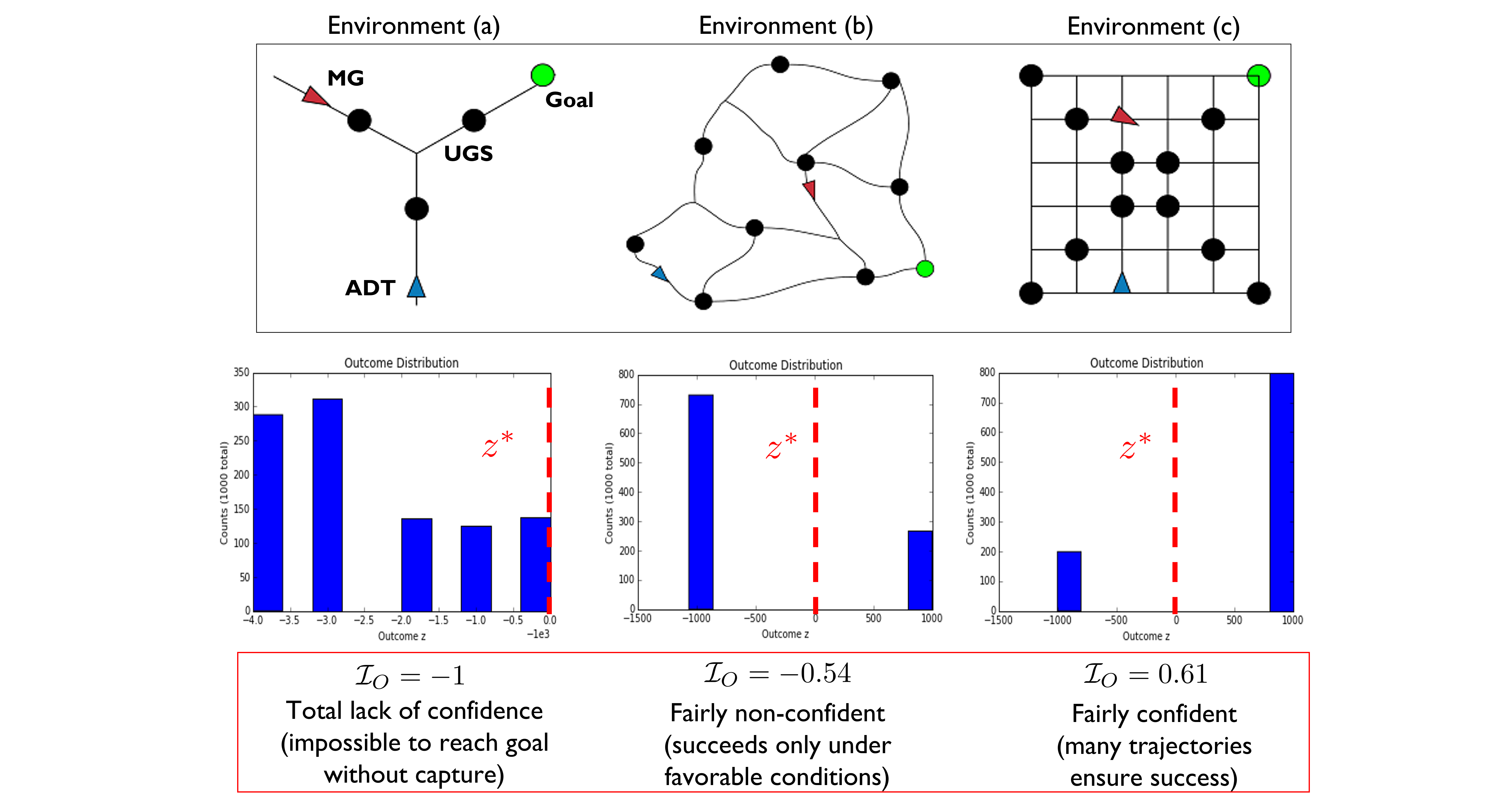}
    \end{center}
  	\caption{$\IO$ ($z^*=0$) in progressively easier task environments for ADT problem using optimal MDP policy (non-discounted reward distributions simulated over 1000 Monte Carlo runs per scenario).}
\label{fig:exprdist_env}
\end{figure}

Figure \ref{fig:exprdist_env} provides more insight into the relationship between $\IO$ and the complexity of the task environment for an optimal MDP agent, using the same reward function specification as before. These examples shows how changes to the task context $\cC$ can affect competency without changing other aspects of the task or $\cA$'s implementation. In (a), it is clearly impossible for the ADT to reach the goal without capture, and so $\IO$ indicates that the agent has no chance of succeeding. In (b), the situation improves since the chances for the ADT to escape the MG are higher, but there is still a significant lack of confidence due to the limited number of paths to the goal and so success hinges on favorable MG movements. The environment in (c) provides multiple escape routes and paths to the goal, thus significantly boosting confidence.

These last three examples also highlight the difficulty of trying to encapsulate and analyze complex multi-objective behaviors into a single outcome measure. As long as the ADT does its best to competently avoid the MG and survive for as long as possible in the road network, it automatically incurs a loitering penalty for not reaching the goal. This incentivizes the optimal policy $\pi$ to engage in riskier behaviors, and as a result $\IO$ as assessed via \rwdapprox$(T)$ really only captures the ADT's competency at reaching the goal quickly (as opposed to also accounting for ability to evade the MG). If the MDP reward function $R$ were tweaked to not penalize loitering and/or not incentivize reaching the goal as quickly as possible, then the interpretation of \rwdapprox$(T)$ changes, making $z^*$ less straightforward to specify in a consistent manner for the same $\pi$. As discussed next, extensions of eqs. (\ref{eq:empUPMLPM}) and (\ref{eqn:oa}) which permit analysis with multiple (possibly competing) outcome measures can be used to address such issues. 

\subsubsection{General remarks and extensions}
\label{ssec:iogenextensions}
As mentioned earlier, the Omega ratio is a special case of the $\alpha-$UPM/LPM ratio for which $\alpha=0$. The Omega ratio is well-suited to binary-valued task outcomes, i.e. success and failure, or more generally to situations where the margin of achievement above/below the $z^*$  threshold does not matter as much as relative the probability mass of outcomes above/below it. 
Refs. \cite{acharya2022competency,acharya2023learning} use the Omega ratio version of \IO{} for competency evaluation with model-based deep reinforcement learning agents performing a variety of tasks in noisy stochastic environments. The Omega ratio is particularly useful in these works to show that their deep learning architectures allow machine self-confidence assessments (expressed through \IO{} only) to correctly encode and calibrate the long-term probabilistic forecasts of task successes and failures. 
Ambiguities due to margin-outcome magnitude symmetries  
can also be resolved by considering still other UPM/LPM ratio generalizations, for instance by using $\alpha \geq 2$ or separate lower and upper bound thresholds for margin evaluation. Additional insight can  also be gleaned by varying $z^*$ to generate $\cA$'s `confidence profile' against a range of competency standards, e.g. see \cite{acharya2022competency}. 

The UPM/LPM ratio for \IO{} can also be generalized to deal with many different types of outcomes, including non-reward based measures and outcomes described by non-continuous/discrete variables $\mathcal{Z}$. 
Refs. \cite{conlon2022generalizing, conlon2022iros, ConlonACMS2024} describe one such extension for generalized outcome assessments (GOA), which adopts a discrete UPM/discrete LPM (dUPM/dLPM) ratio for $\alpha=1$ over an integer-valued outcome variable $z$ with (empirical) probability $P(z)$, 
\begin{align}
      M_{\cO} = M(\tilde{P}(z),z^{*}; \alpha=1):=\mbox{dUPM/dLPM} = \frac{\sum_{z \geq z^*}(z-z^{*}+1){\tilde{P}}(z)}{\sum_{z<z^*} (z^{*}-z){\tilde{P}}({z})}. \label{eq:empdUPMdLPM}
\end{align}
Here, the values for $z$ represent an indexed ordering over mutually exclusive and exhaustive equivalence classes for sets of outcomes in $\mathcal{Z}$, such that $z'<z$ implies that all outcome events associated with $z'$ are less favorable than $z$. In essence, the discrete $z-$space representation of outcomes enables the construction of a post hoc utility over combined sets of multiple outcomes. As in the continuous UPM/LPM ratio, $z^*$ represents a minimal acceptable outcome threshold, from which expected probabilistic moments above/below can be computed. From the GOA standpoint, it is not strictly necessary for different outcome measures to be combined into a single scalar outcome variable for analysis.
Hence, (\ref{eqn:oa}) and (\ref{eq:empdUPMdLPM}) could also each be straightforwardly computed in parallel for any set of separated measures to obtain vector-valued \IO{} indicators. 

\section{Solver Quality}\label{sec:solverquality}
Whereas $\IO$ interprets differences between actual agent outcomes $\cO$ (following policy $\pi$) and desired outcomes $\cO^{des}$ (in competency standard $\Sigma$), this comparison does not say much about \emph{how} $\cO^{des}$ \emph{should} be obtained and whether $\cA$ comes close to this behavior. 
For MDP agents, an evaluator would like to assess how closely $\cA$'s process for solving the policy $\pi$ meets a given standard for obtaining a satisfactory policy solution in task context $\cC$ (assuming the MDP model for state dynamics, rewards, and utility can be taken for granted). 
Questions with implied standards along these lines include, for instance: does it take too long for $\cA$ to obtain $\pi$?; does $\pi$ cover enough of the state-action space?; is $\pi$ close enough to the true optimum policy?
We argue that $\IS{}$ should interpret a new set of problem-solving statistics obtained under context $\cC$, which are derived from margins between the portion of $\Sigma$'s representing the solver goodness and the actual goodness of $\pi$ produced by $\cA$.  The policy $\pi$ produced by $\cA$ can generally be considered a random variable, which induces a corresponding sample distribution over measures of $\pi$'s goodness given $\cC{}$ (where the measure of goodness could coincide with variables in $\cO$ or $\cO^{des}$, but does not have to). 
However, the problem of specifying computable indicators for $\IS{}$ and corresponding standards in $\Sigma$ introduces new challenges that (to our knowledge) have not been fully addressed in either the algorithmic meta-cognition or human-machine interaction literature. These issues are formally described next through the introduction of another meta-utility describing the value of using $\cA$'s solver to produce $\pi$ according to a solution quality competency standard. Then, a novel definition and approach to computing $\IS{}$ is introduced based on the idea of using learned surrogate models to probabilistically evaluate expected policy performance in different task contexts. Detailed results applying this new approach to the ADT problem are then presented and discussed. 

\subsection{\IS{} Definition}
Consider a task \task{} of class \taskclass{}, which represents the set of all models of $\cA{}$'s task process in Figure~\ref{fig:CompAssessGenBlockDiag} and which $\cA{}$ is therefore theoretically capable of solving. In our running example \taskclass{} is the set of all autonomous delivery problems that can be posed as MDPs with different starting locations, destinations, road-network layouts, etc., and \task{} is a realization of an MDP for a single delivery problem. Let \solve{} denote the solver (algorithmic process) that $\cA{}$ uses to determine actions for task \task. For an MDP agent, \solve{} takes in an MDP specification for \task and produces a policy \pigeneric{}. 

A formal definition for \IS{} must make precise the notion of \cA{}'s `competence' with regard to \solve{} to enable some form of computable evaluation. For MDPs, any particular task instance \task{} has a corresponding optimal policy \piopt{} which by definition leads to corresponding best achievable value function. This suggests that a natural performance metric for assessing the competence of \cA{} using generic solver \solve{} would be some quantitative comparison of the (not necessarily optimal) policy \pigeneric{} to \piopt. Such a `strong' comparison can be done in many ways depending on the application context, e.g. by directly comparing actions taken by the policies in different states of interest, or comparing the resulting expected total reward distributions or other policy artifacts. 

However, \piopt{} is usually unavailable. Thus, \IS{} should also allow for `weak' comparison of \pigeneric{} to other `baseline' reference solvers that yield policies which are comparable in some sense to the true optimal \piopt. This may require comparing two completely different types of solvers, e.g. a deterministic approximation vs. a Monte Carlo approximation. So \IS{} should also ideally enable comparison across solver classes including online/anytime solvers, for which complete state coverage and policy convergence may not be possible in large state/action spaces (e.g. recall that MCTS only considers a small portion of the total state-action space, this is reviewed in Appendix \ref{sec:app_MCTS}).
 
At the same time, \IS{} should account for the expected self-confidence metric trends and boundary conditions mentioned earlier in Table \ref{tab:boundaryConditions}. In particular, \IS{} should naturally account for characteristics of \solve{} and features of \task{} simultaneously. For instance, if \task{} is not too complex or is characterized by a very small amount of uncertainty (nearly deterministic), then it is possible that even a simplistic \solve{} making many approximations could produce a \pigeneric{} that closely approximates \piopt, in which case \IS{} should indicate very high confidence. Moreover, while it is impossible to assess performance on \emph{all} of \taskclass{}, \IS{} should reflect \solve{}'s performance on \emph{any} task instance \task{} in \taskclass{}, including previously unseen tasks (e.g. new road networks for the ADT). 
These points can be summarized via three key desiderata (D) and technical challenges (TC) for \IS{}, as follows:
    
\begin{enumerate}[label=\textbf{D\arabic*}]
    \item Reflect competence of \cA{} using solver \solve{} for task \task{} (requires some kind of comparison)\label{itm:d1};
    \item Enable comparison across solver classes (exact vs. approximate vs. deterministic)\label{itm:d2};
    \item Extend to any \task $\in$ \taskclass\label{itm:d3}.
\end{enumerate}
\begin{enumerate}[label=\textbf{TC\arabic*}]
    \item It is unclear how different policies/solvers should be compared. 
    \label{itm:l1}
    \item Large state spaces make exact optimal solutions infeasible to identify or implement, and so approximate policies must be considered in practice. 
    \label{itm:l2}
    \item It is generally impossible to evaluate policies for \emph{all} \task{} $\in$ \taskclass{}. 
    \label{itm:l3}
\end{enumerate}

Section \ref{sec:comparing_solvers} mainly addresses \ref{itm:l1}, which is linked to \ref{itm:d1} and \ref{itm:d2}. Section \ref{sec:no_pi_opt} discusses the remaining two challenges \ref{itm:l2} and \ref{itm:l3}, which are related to \ref{itm:d3}.

\subsection{Comparing Solvers via Policy Outcome Distributions} \label{sec:comparing_solvers}

Setting aside challenges \ref{itm:l2} and \ref{itm:l3} for the time being, how do we first identify a general basis for comparing two solvers (possibly of different types) to determine which one likely bestows \cA{} with greater competency? 
    Since solvers of all types produce some kind of policy \pigeneric{}, it is natural to consider comparing solvers directly on this basis.   
    Unfortunately, when considering how to compare policies, challenges that are similar to those encountered when trying to compare solvers arise: policies for the same \task{} can be equally varied and represent the state-action spaces very differently. How, then, can policies be compared in a general way? Several possibilities are described in Table~\ref{tab:policy_comparison}.

    \begin{table}[htbp]
    \caption{Merits/demerits of different policy comparisons for \ref{itm:d1} and \ref{itm:d2} (\ref{itm:d3} considered in next section)}
    \label{tab:policy_comparison}
    \footnotesize
    \begin{tabulary}{\linewidth}{cLLLccc}
    \toprule
    \# & Method & Merit & Demerit \\ \midrule
    1 & Compare the utilities that the two policies assign to each state & Evaluates whether states are assigned equal utility across solvers, where state utilities should theoretically be independent of the solver. Addresses \ref{itm:d1}. & Doesn't address \ref{itm:d2}---Doesn't apply when different solvers provide different coverage of the state-action space, or represent the state-action space differently. For example, an approximate solver like MCTS may only find a policy for a small `reachable' subset of the state-action space, while value iteration finds a policy for every state-action pair. \\
    \\
    2 & Compare ‘coverage’ of the policies, i.e. proportion of total state-action space covered by each policy & Evaluates how `thorough' the policy is; in concert with the previous suggestion of comparing utilities, could possibly address \ref{itm:d1}. & Doesn't satisfy \ref{itm:d2}---not all policies have the same state-action coverage, typically by design. Also, high coverage does not imply a `good' solution (consider a poorly converged value iteration policy, versus a well-parameterized MCTS policy). \\
    \\
    3 & Compare distributions for cumulative rewards \rwdapprox{}, e.g. from Monte Carlo simulations of following each policy & Meets \ref{itm:d1} in that the expected reward distributions are a metric of the performance of a given policy; also able to satisfy \ref{itm:d2} as expected reward distributions can be simulated from a policy of any class & Determining cumulative reward distribution may be a deterrent (e.g. time to run Monte Carlo simulations may be large, though will typically be less than finding the policy itself). \\ \bottomrule
    \end{tabulary}
    \end{table}

Of the possibilities listed in Table~\ref{tab:policy_comparison}, comparing expected distributions for cumulative rewards is the only option that satisfies \ref{itm:d1} and \ref{itm:d2}. Subsequently, a natural starting point for quantitatively assessing \IS{} for MDPs is to analyze how uncertainty in \rwdapprox{} 
(which can also be used to form \IO{}) provides a measure of the policy $\pi$ (and hence the solver \solve{}) that produced it. Note that although other outcomes \cO{} could also be used more generally in place of \rwdapprox{}, cumulative rewards provide a convenient starting point since they are tractable proxies for value functions and hence often implicitly maximized in expectation by MDP solvers. 

This brings us to consider how \rwdapprox{} behaves as a random variable with probability distribution $P_{\solve{}}(\rwdapprox{})$ produced by some given \solve{} on a particular task \task{}. 
We introduce another meta-utility function $M_{\solve}=M(P_{\solve{}}(\rwdapprox{}),P^{des}(\rwdapprox{}))$ that will be used to formally generate an \IS{} value for using \solve{} on \task{}, based on comparing $P_{\solve{}}(\rwdapprox{})$ to a distribution $P^{des}(\rwdapprox{})$ defined by the competency standard $\Sigma$. The better/worse $P_{\solve{}}(\rwdapprox{})$ looks in relation to $P^{des}(\rwdapprox{})$, the higher/lower $M_{\solve}$ (and hence \IS{}) should become. 
Also, for comparing two solvers $\solve{}_1$ and $\solve{}_2$, it should follow that $M(P_{\solve{}_1}(\rwdapprox{}),P^{des}(\rwdapprox{})) > M(P_{\solve{}_2}(\rwdapprox{}),P^{des}(\rwdapprox{})) \rightarrow$ $\cA$ is more competent with $\solve{}_1$ than $\solve{}_2$. 
Natural choices for $M_{\solve}$ can be derived from distances for probability distributions. However, these typically only return non-negative measures, whereas some form of signed information is needed to ascertain how much of $P_{\solve{}}$'s support and probability mass lies above, within, or below the support of $P^{des}(\rwdapprox{})$. 
One such probability distribution distance measure is considered next; Sec.\ref{ssec:calcIS} later describes how it can be turned into a meta-utility function $M_{\solve}$ with the desired behavior, which can then be mapped to an interpretable competency indicator \IS{}. 
    
\subsubsection{Hellinger Distance $\bm{H}$:} \label{sec:hellinger}
    Let $P(r)$ be the probability distribution for realizations $r$ of random variable \rwdapprox{} under some given policy $\pi_1$, obtained say via \solve{}$_1$, and $Q(r)$ be the distribution under another policy $\pi_2$ obtained via \solve{}$_2$. 
    While there are many methods for comparing two probability distributions \cite{chung1989measures}, one straightforward way of calculating the similarity between two reward distributions is to find the distance or divergence between them. The Hellinger distance $H$, 
    \begin{align}
        H(P,Q) = 1 - \int_{-\infty}^{\infty} \sqrt{P(r)Q(r)}dr, \label{eq:hell_gen}
    \end{align}
    is a metric on the space of probability distributions for \rwdapprox{} that is bounded between 0 and 1 (something that will become useful when addressing \ref{itm:l2} and \ref{itm:l3}). 
    When $H=0$, $P$ and $Q$ are identical; the maximum distance of $H=1$ is achieved when $P$ assigns zero probability density at every point in which $Q$ assigns non-zero probability density (and vice-versa). $H$ obtains different closed-form expressions based on the type of analytical distributions being compared; for two Gaussian distributions $P \sim {\mathcal{N}}(\mu_p,\sigma^2_p)$ and $Q\sim {\mathcal{N}}(\mu_q,\sigma^2_q)$, the square of \hell{} is
    \begin{align}
        H^{2}(P,Q) = 1-\sqrt{\frac{2\sigma_P\sigma_Q}{\sigma_P^2+\sigma_Q^2}}\exp{\left(-\frac{1}{4}\frac{(\mu_P-\mu_Q)^2}{\sigma_P^2+\sigma_Q^2}\right)}. \label{eq:hell_norm}
    \end{align}

For this work it is assumed that all \rwdapprox{} distributions can be sufficiently described by their means $\mu$ and standard deviations $\sigma$, resulting in a Gaussian approximation. In the case of the ADT problem, this assumption can sometimes be a stretch as illustrated in Fig.~\ref{fig:exprdist}, where higher order moments would be needed to fully capture other important distribution features (skewness, multi-modality, heavy-tail behavior, etc.). 
Having said that, the Gaussian approximation often manages to capture enough useful information 
via the Hellinger distance without introducing too much extra complexity. 
If more information about $P$ and $Q$ are required, a different form of the Hellinger distance (or different distance measure altogether) can be used. For example, if $P$ and $Q$ are represented by histograms $\hat{P}$ and $\hat{Q}$ across $N$ discretized reward bins $r_i$, then
\begin{align}
    \hell(\hat{P},\hat{Q}) = 1 - \sum_{k=1}^{N}{\sqrt{\hat{P}(r_i)}\cdot \sqrt{\hat{Q}(r_i)}}. \label{eq:hell_disc}
\end{align}    
The main takeaway is that, by examining the corresponding expected reward distributions of different policies from different solvers, the Hellinger distance provides a good starting point 
to address \ref{itm:l1}, when $P^{des}(\rwdapprox{})$ is provided in the competency standard $\Sigma$. 
However, the Hellinger distance still needs to be converted into the meta-utility $M_{\solve} = M(P_{\solve{}}(\rwdapprox{}),P^{des}(\rwdapprox{}))$ that indicates where the support and mass of $P_{\solve{}}(\rwdapprox{})$ are in relation to those of $P^{des}(\rwdapprox{})$. 
Moreover, to actually calculate $M_{\solve}$,
we must still resolve \ref{itm:l2} (\policyopt{} isn't available to provide $P^{des}(\rwdapprox{})$ for each task \task) and \ref{itm:l3} (can't calculate \policyopt{} for all possible task instances \task). The latter two issues are tackled next to provide all the ingredients needed to modify $H$ and define $M_{\solve}$. 

\subsection{Coping without $\policyopt{}$ and extending to any task} \label{sec:no_pi_opt}
%
Recall that competence self-assessment implies comparison between agent $\cA{}$'s idea of what it expects it can accomplish and what it is expected to accomplish according to a standard. From the standpoint of assessing MDP agent competency, the comparison ideally ought to be informed by the optimal policy \piopt{} for a given task. However, as argued earlier, competency assessment should also allow for weaker forms of assessment, recognizing that if \piopt{} were available, $\cA$ would not need to approximate $\pi$ via methods like MCTS in the first place. 

To this end, we draw inspiration from the fact that, in many instances, many humans are able to reasonably self-assess their competency limits in regard to how they would accomplish tasks without requiring access to the `best possible' way of performing them, e.g. see \cite{Feltz1988}. 
The main insight here is that, even if access to the best possible standard is not available, having a `high enough' standard that serves as an \emph{accessible} reference to a sufficiently wide range of tasks can provide sufficient information to gauge competency. 
%
Following this idea: instead of requiring the optimal MDP policy \policyopt{} (from the `ideal' solver \solveopt{}) to compare to $\cA{}$'s candidate policy $\pi$ (from its available solver \solve{}), a sub-optimal but still fairly good reference solver can be used. In other words, a `trusted solver' \solvetrust{} can be introduced as the reference to which any `candidate solver' \solvecand{} can be compared. This trusted solver \solvetrust{} need not produce a policy identical to \policyopt{} (though that is not prohibited). The key function of \solvetrust{} is that it serve as some standard by which the decision-making agent can judge its own competence. Of course, the closer \solvetrust{} is to \solveopt{}, the closer competence assessment will be relative to the ideal standard. 
In practice, \solvetrust{} could be any `kitchen sink' algorithm that uses more resources (e.g. time, processing power, memory) than are available to a typical candidate solver \solve$=$\solvecand{} and is thus theoretically capable of producing policies \policytrust{} that approximate \piopt{} in some limit for any task instance, even if deploying \policytrust{} on \cA{} is impractical (which we assume to typically be the case). 
The main requirement is that \solvetrust{} be available to produce \policytrust{} for any possible MDP instantiation \task{} $\in$ \taskclass{} that \cA{} may encounter, even though it is impractical (if not impossible) to actually obtain policies for all of \taskclass{}. 

Literature on Empirical Hardness Models (EHMs) provides guidance for overcoming this challenge. Refs. \cite{leyton2014understanding} and \cite{hutter2014algorithm} introduced EHMs to predict the empirical runtime performance (as opposed to the `Big-O' runtime) of an algorithm on a given problem instance. Specifically, they showed how the actual runtime of NP-complete problems can be predicted using learning-based surrogate regression models, which used performance data from previously encountered problem solution instances to predict runtime performance as a function of problem instance features. 

Applying similar logic in the domain of sequential decision-making agents, a surrogate model \surrogate{} can be learned to predict the distribution of the reward $\rwdtrustpredict\approx\rwdtrust$ from the policy \policytrust{} of a reference trusted solver \solvetrust{} for some finite subset of tasks in \taskclass. In this way, it is possible to estimate the performance of \solvetrust{} on other problems in \taskclass. A function to predict the expected reward distribution \rwdtrustpredict{} can be learned given the features $\bm{x}$ of \task{} and \solvetrust{}, since \solvetrust{} and \policytrust allows us to approximate \rwdtrustpredict{} via simulation for any task instance. In the case of the ADT problem, $\bm{x}$ could for instance include the total number of intersections or the transition probabilities in the road-network, as well as MCTS hyper-parameters (among many other possible variables; see Sec. \ref{sec:delivery_numerical_experiments}).  

To summarize, the application of a trusted solver \solvetrust{} and surrogate model \surrogate{} addresses both challenges \ref{itm:l2} 
and \ref{itm:l3}. 
In the next section, we formally develop the computational methods that enable these ideas to be applied within the \famsec{} framework.

\subsubsection{Learning \surrogate}
    \begin{algorithm}[htbp]
        \footnotesize
        \DontPrintSemicolon
        \SetNlSty{texttt}{}{:}
        
        \Begin{
        \SetNlSkip{2em}
        $\solvetrust =$ `trusted' solver (i.e. MDP value iteration \emph{or} MCTS solver with large $d$ and $N$)\\
        $T =$ set of training tasks selected over task feature space $\bm{x}$ \\ 
        $m = $ number of Monte-Carlo runs for simulation\\
            \For(create training data){\taski{} in $T$}{
            calculate \policytrust{} for \taski{} using \solvetrust\\
                \For(simulate \rwdtrustisim){j=1:m}{
                    simulate total cumulative reward from following \policytrust{} on \taski{}. Store reward as $j^{th}$ element of \rwdtrustisim
                }
            }
        instantiate \surrogate{} with initial set of parameters\\
            \Repeat( learn \surrogate){convergence}{
                \For{\taski{} in $T$}{
                \rwdtrustpredict{} = estimate based on predicted distribution from \surrogatearg{\taski{}}\\
                adjust parameters of \surrogate{} based on error between \rwdtrustisim{} and \rwdtrustpredict{}
                }
            }
            \KwRet \surrogate
        }

        \caption{Learning \surrogate}
        \label{alg:train_surrogate}
    \end{algorithm}
    \begin{figure}[htbp]
        \centering
        \begin{minipage}[b]{0.55\linewidth}
            \centering
            \includegraphics[width=0.65\linewidth]{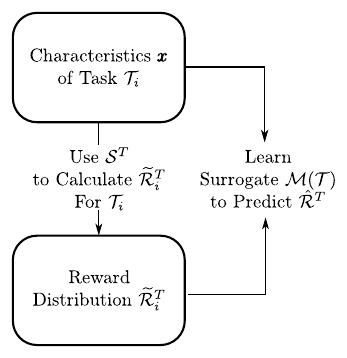}
            \vfill
            \subcaption{Offline Training of \surrogate}
            \label{fig:IS_train}
        \end{minipage}%
        \begin{minipage}[b]{0.50\linewidth}
            \centering
            \includegraphics[width=0.6\linewidth]{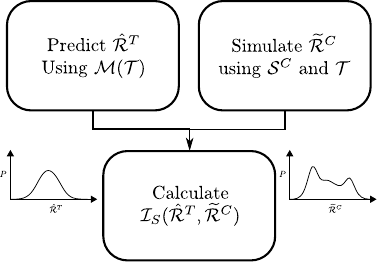}
            \vspace{0.5cm}
            \subcaption{Online Deployment of \surrogate{} to calculate \IS}
            \label{fig:IS_test}
        \end{minipage} 
        \caption{\IS{} calculation phases.}
        \label{fig:IS_test_train}
    \end{figure}

    \begin{algorithm}[htbp]
        \footnotesize
        \DontPrintSemicolon
        \SetNlSty{texttt}{}{:}
        
        \Begin{
        \SetNlSkip{2em}
        \task{} = task on which to evaluate \IS{} (associated features $\bm{x}$ and initial state $s=\{s_{ADT},s_{MG}\}$)\\
        $m =$ number of Monte-Carlo runs\\
        calculate \policycand{} for \task{} using \solvecand\\
            \For(simulate \rwdcandsim){j=1:m}{
                simulate total cumulative reward from following \policycand{} on \task{}, store reward as $j^{th}$ element of \rwdcandsim
            }
            predict distribution of \rwdtrustpredict{} using \surrogate\\
            \KwRet \IS(\rwdtrustpredict,\rwdcandsim) using Eqs. \ref{eq:q}, \ref{eq:f}, \ref{eq:IS}
        }
        \caption{Calculating \IS }
        \label{alg:calc_IS}
    \end{algorithm}

    A surrogate model is an informed proxy for another target function, where the surrogate has properties that make it easier to work with than the target function for purposes of prediction, inference, optimization, etc. The surrogate model \surrogate{} can be any model (black-box or otherwise) capable of predicting the distribution for $\rwdtrustpredict\approx\rwdtrust$ given \task{} with parameters (or features) $\bm{x}$. Popular examples of surrogate models include polynomial functions, neural networks, or Gaussian processes.

Algorithm~\ref{alg:train_surrogate} details a procedure for training a surrogate model to predict the distribution of \rwdtrustpredict; Figure \ref{fig:IS_train} is a simple diagram of the procedure. Several training tasks \task{} that have been solved by a trusted solver \solvetrust{} are used as training data. The features $\bm{x}$ of the various task instances are selected to inform the distribution of \rwdtrustpredict. 
Learning \surrogate{} would typically be done offline when more computation power and time are available to apply \solvetrust{} and the surrogate model learning process. As illustrated in Figure~\ref{fig:IS_test}, \surrogate{} can then be deployed later for use on an autonomous system in order to furnish estimates of \rwdtrustpredict to which \cA{} can probabilistically compare its own estimate \rwdcandsim{} of expected rewards (i.e. running only \solvecand{} and without having to actually implement \solvetrust{}). Pseudocode for this procedure is outlined in Algorithm~\ref{alg:calc_IS}.


\subsubsection{Calculating \IS}
\label{ssec:calcIS}
\begin{figure}[tbp]
    \centering
    \includegraphics[width=0.85\linewidth]{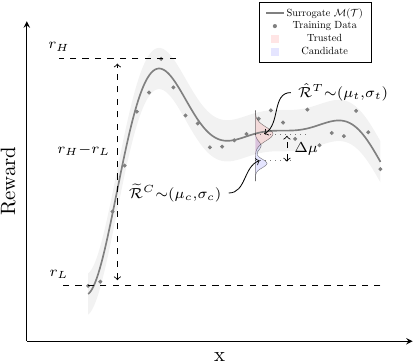}
    \caption{Inputs to $M_{\solve}$ ($x$ represents features of interest for \task $\in$ \taskclass).}
    \label{fig:IS_v2}
\end{figure}
With \ref{itm:l1}-\ref{itm:l3} now addressed, the meta-utility $M_{\solve}$ and \IS{} indicator itself can be formulated. Figure~\ref{fig:IS_v2} illustrates the key quantities involved in calculating $M_{\solve}$ and \IS{}. 
Recall that in combining solutions for \ref{itm:l1}-\ref{itm:l3}, some modification of the Hellinger distance is needed to produce a sensible meta-utilities via $M_{\solve}$ across possible tasks in \taskclass. 
Consistent with the idea of evaluating margins relative to a competency standard $\Sigma$, the general idea is to evaluate $M_{\solve}$ as the \emph{signed} distance between the probability distributions for \rwdtrustpredict{} (modeled via \surrogate{}) and \rwdcandsim{}, \emph{ in relation to the expected range of rewards obtained by} \solvetrust{}. Note that here $\Sigma$ is represented by both the distribution for \rwdtrustpredict{} and the expected range of rewards for \solvetrust{}. 
Assuming the reward distributions are described by their means and standard deviations via Gaussian pdfs $C=P_{\solve{}}(\rwdapprox{})={\mathcal N}(\mu_c,\sigma_c)$ for \solvecand{} and $T=P^{des}(\rwdapprox{})={\mathcal N}(\mu_t,\sigma_t)$ for \solvetrust{}, respectively, the signed distance is given by 
\begin{align}
    M_{\solve{}} = M(C,T) :=  \text{sgn}(\Delta \mu)f^{\kappa}\sqrt{H^{2}(T,C)}, \label{eq:q}
\end{align}
where sgn is the sign function, $\Delta \mu = \mu_t - \mu_c$, and the scaling factor $f$ is given by 
\begin{align}
f = \frac{|\Delta \mu|}{|r_H - r_L|}, \label{eq:f}     
\end{align}
where $r_H$ and $r_L$  respectively correspond to (bounds on) the largest and smallest possible reward realizations observed in the data for training \surrogate{} to predict \rwdtrustpredict{}.  

The extra terms in front of the Hellinger distance in eq. (\ref{eq:q}) ensure that the resulting $M_{\solve}$ conforms to desired meta-utility behavior, and are justified as follows. 
Firstly, eq. (\ref{eq:hell_gen}) is a distance measure (i.e. \hell{}$>=0$), and so information that indicates if one distribution is generally better or worse (i.e. more or less expected reward) is lost. Hence, $\text{sgn}(\Delta \mu)$ recovers this information. 
Secondly, the distributional similarity described by \hell{} (or other such measures) must be calibrated to the overall scale of expected outcomes in the reward space. 
For instance, consider an extreme case of two distributions with means $\mu_t=1$ and $\mu_c=2$ and low variances $\sigma_t^2=\sigma_c^2=1e\-5$. In this case \hell{} between the two reward pdfs would be nearly $1$ (maximum distance/dissimilarity), since the overlap between them is very small. However, if the rewards from many other training tasks are on the range $[-1e3,1e3]$, then the means are nearly equal on the global scale and 
the two distributions in this case are nearly identical. 
The scale factor $f$ allows different contexts like these to be taken into account.
The exponent $\kappa \in (0,1)$ in eq. (\ref{eq:q}) balances the influence of $f$ vs. \hell, 
where ideally \hell{} should be more influential on $M$ as $f$ grows larger, and $f$ should be more influential as it decreases. 
We found empirically $\kappa=0.5$ gives reasonable results (as shown next), though 
other formulations 
could also address this balance. 

Note that \hell{} ranges on $[0,1]$, while $f$ ranges on $[0,\infty)$. It is thus desirable to use a squashing function on $M_{\solve}$ to bound reported \IS{} range and avoid arbitrarily large values that could confuse human informees (ultimately the choice of bounded range is somewhat arbitrary, as long as it can be understood by informees). 
As with \IO{}, a logistic transformation on $M$ is useful for this,
\begin{align}
    \IS &= \frac{2}{1+\exp(-M_{\solve}/5)}\label{eq:IS},
\end{align}
which ranges from $[0,2]$ (in deliberate contrast to \IO{} which ranges from $[-1,1]$). The intent here is for \IS{} to be loosely thought as the `competence gain' of \cA{} when using \solvecand{} instead of \solvetrust{}. That is: when \IS$=1$, then \solvecand{} results in $1\times\solvetrust$ competency, i.e. reward distributions of \solvetrust{} and \solvecand{} are identical and \cA{} is about as competent with either solver ($M_{\solve}=0$); when \IS$=2$, using \solvecand{} results in a much more competent \cA{} than when using \solvetrust ($M_{\solve}>>0$); and when \IS$=0$, using \solvecand{} results in a much less competent \cA{} than when using \solvetrust ($M_{\solve}<<0$). Dividing $M_{\solve}$ by 5 saturates \IS{} near $M_{\solve}=\pm1$. While other tunings or functional forms could be used in (\ref{eq:IS}), the key point is that \IS{} (like \IO{}) is a function of the \emph{margin} between \cA's capabilities and the competency standard $\Sigma$, reflected in this case by the (signed) distance of estimated reward probability distributions. Note that \IS{}$\leq 1$ for the special case where \solvetrust{} is the optimum solver for \task{} and \piopt{} is the trusted policy. This property is useful for assessing the general performance of \solvetrust{} in different contexts, i.e. in that $\IS{}>1$ means \solvetrust{} underperforms relative to some candidate \solve{} in a particular context. 

\subsection{Application Examples}

\subsubsection{Synthetic \IS{} Evaluations}
\begin{figure}[tbp]
    \centering
    \begin{minipage}[c]{0.60\linewidth}
        \centering
        \includegraphics[width=0.90\linewidth]{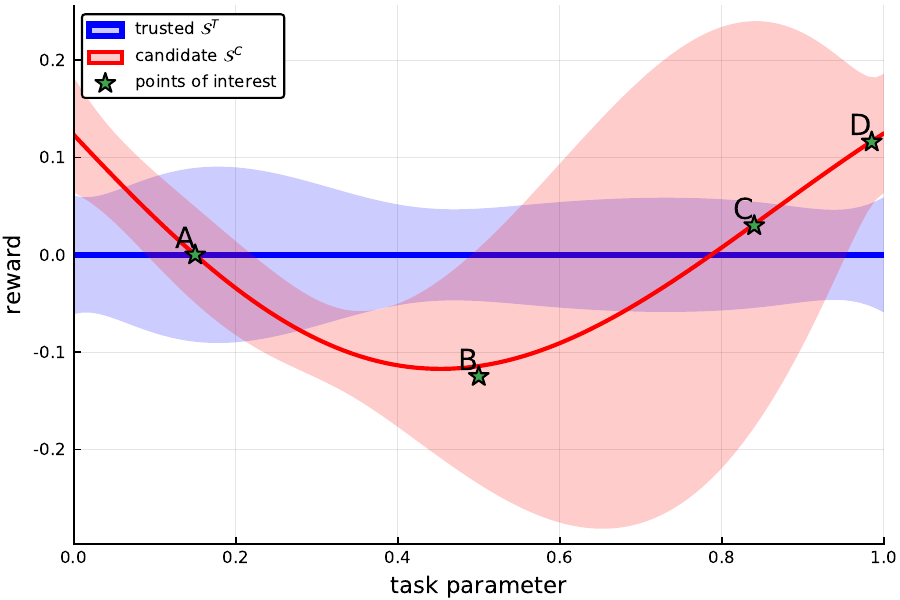}
        \vfill
    \end{minipage}%
    \begin{minipage}[t]{0.40\linewidth}
        \centering
        \includegraphics[width=0.90\linewidth]{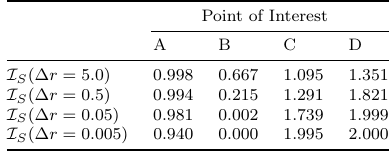}
    \end{minipage}
    \caption{\IS{} for synthetic  $p($\rwdtrustpredict{}$)$ pdf for \solvetrust{} and  $p($\rwdapprox$)$ pdf for \solvecand, showing variation of mean (solid) and $2\sigma$ bounds (shaded regions) as function of a single continuous hypothetical task parameter. }
    \label{fig:IS_thry1}
\end{figure}

We first present a simplified toy example to show how \IS{} typically behaves across a range of reward distributions. 
Figure~\ref{fig:IS_thry1} shows synthetic plots of $p($\rwdtrustpredict$)$ for a hypothetical trusted solver \solvetrust{} vs. $p($\rwdcandsim$)$ for a hypothetical candidate solver \solvecand{} on a simple MDP, where a single continuous task feature is varied. For illustrative purposes, the synthetic distributions shown here are reasonably characterized by their means and standard deviations. 
Different points of interest indicating specific values of the task parameter are highlighted by the star points. The table on the side shows the values of \IS{} calculated for different cases.

For instance, at point $B$, \solvecand{} has a lower expected reward than \solvetrust{} and a larger variance. Intuitively \IS{}, should be less than one. As shown when $\Delta r= r_H-r_L=5$ (a relatively `large' global reward range), a solver quality indicator of $\IS=0.667$ is obtained, implying \solvecand{} is marginally less capable than \solvetrust{}. When $\Delta r=0.05$ then $\IS=0.002$, indicating that \solvecand{} is much less capable than \solvetrust{}.
In contrast, at point $C$, \solvecand{} has higher expected reward than \solvetrust{}, but also larger variance. Intuitively, we would expect \IS{} to be slightly greater than one, and in fact when $\Delta r=5$, $\IS=1.095$. As $\Delta r$ decreases, the assessed difference in capability between \solvecand{} and \solvetrust{}{} increases with $\IS{}=1.995$ at $\Delta r=0.005$. 

\subsubsection{Application to the ADT Problem} \label{sec:delivery_numerical_experiments}

We now show how the process in Fig. \ref{fig:IS_test_train} applies to the ADT problem. 
In particular, we demonstrate how the behavior of \IS{} changes as a function of different delivery problem instance features and of the characteristics of approximate candidate solvers, relative to a baseline reference trusted solver \solvetrust{}. 

The surrogate model \surrogate{} used here combines the output of two deep neural networks: one which predicts the mean of $p($\rwdtrustpredict{}$)$ and another which predict its standard error. The surrogate is trained on 341 randomly generated road network problems using an MCTS-based trusted solver configuration \solvetrust{} that was selected to accommodate a wide variety of potential delivery tasks \task. An MCTS solver was convenient to use for this application for two major reasons. Firstly, the quality of the solver used by \cA{} can be easily modified according to a number of parameters, allowing us to see how well \IS{} reflects changes in second-order competency to their adjustment. Secondly, MCTS solvers are a practically useful class of candidate solvers \solvecand{} in applications of this type, where time-bounded decision-making is required when the number of task states and actions are potentially large. These aspects also allow us to link the behavior of \IS{} to considerations for practical system design. For brevity, we focus here on the results obtained for \IS{} assessments, and provide complete details of the surrogate model training process in Appendix \ref{sec:app_learning_surrogate}. 


\begin{figure}[tbp]
    \centering
    \begin{minipage}[b]{0.50\linewidth}
        \centering
        \includegraphics[width=0.6\linewidth]{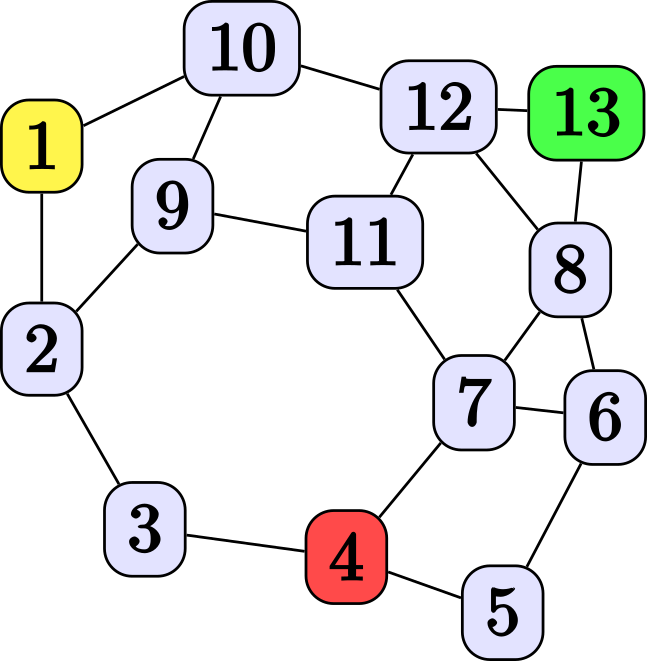}
        \vfill
        \subcaption{Road network N=13}
        \label{fig:roadnet}
    \end{minipage}%
    \hfill
    \begin{minipage}[b]{0.50\linewidth}
        \centering
        \includegraphics[width=0.6\linewidth]{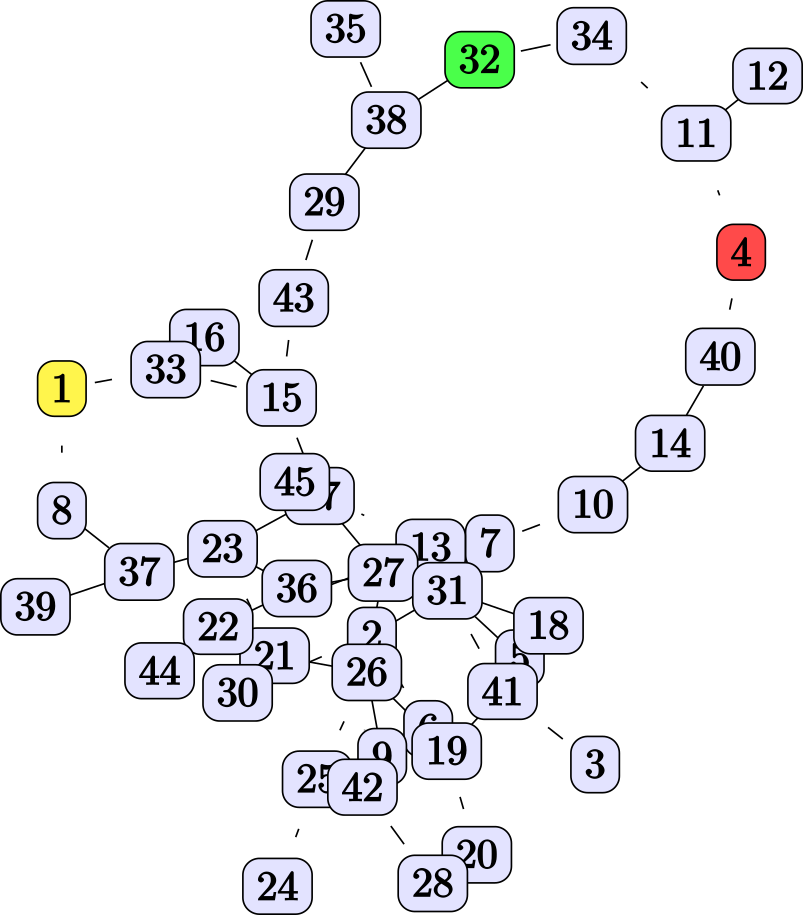}
        \subcaption{Road network N=45}
        \label{fig:med_roadnet}
    \end{minipage} 
    \caption{Example road networks: ADT starts at yellow, MG starts at red, and the Goal is green.}
    \vspace{-0.2cm}
\end{figure}

The behavior of \IS{} is examined for two delivery task instances represented by the road networks shown in Fig.~\ref{fig:roadnet}. In each task, ADT begins at the yellow node (node 1), the MG begins at the red node, and the desired exit is indicated by the green node (Goal). The problem is defined by the parameters listed in Table~\ref{tab:params}. 
Three separate evaluations are shown: (i) \IS{} is calculated for different MCTS candidate solvers \solvecand{} with varying depth parameters (all other parameters held constant); (ii) \IS{} is evaluated for a candidate solver with varying task parameters; and (iii) \IS{} is evaluated for \solvecand{} with varying task \emph{and} solver parameters. Note that in (i), the solvers are compared directly to each other so the surrogate model \surrogate{} is not needed, whereas \surrogate{} is used in (ii) and (iii). 
Note that in all the figures shown in this section, the label `SQ' denotes shorthand for the solver quality indicator \IS.

\tymin=80pt
\begin{table}[tbp]
\centering
\footnotesize
\caption{Table of parameters: Autonomous Doughnut Delivery Problem}
\label{tab:params}
\begin{tabulary}{0.75\linewidth}{LL}
\hline
Parameter    & Description\\
\hline
$p_{trans}$    & MDP transition probability, i.e. the probability ADT moves to selected neighboring road network node at next time step when attempting to move. The ADT goes to a different neighboring node with probability $1-p_{trans}$. \\
$\gamma$            & MDP discount factor. \\
$N$            & Number of nodes (intersections) in the road network.\\
$e_{m}$     & MCTS exploration constant parameter.\\
$d_{m}$     & MCTS search tree depth.\\
$its_{m}$   & MCTS number of Monte Carlo simulations to run to find local policy.\\
$R_{\mbox{goal}}$   & Reward for the ADT successfully reaching Goal node.\\
$R_{\mbox{caught}}$ & Reward (penalty) for MG capturing ADT (occupying same node).\\
$R_{\mbox{loiter}}$  & Reward (penalty) for ADT occupying non-Goal state while not caught.\\
\hline
\end{tabulary}
\end{table}

\begin{table*}
    \footnotesize
    \centering
    \caption{Parameters used for numerical studies}
    \label{tab:exps}
    \begin{tabular}{llcccccccccc} \toprule
        &\multicolumn{10}{c}{Task and Solver Parameters} \\ \cmidrule(r){3-12}
        \#  & $\bm{x}$ variable(s) & Network &$p_{trans}$ & $\gamma$ & $N$ & $e_{m}$ & $d_{m}$ & $its_{m}$ & $R_{\mbox{goal}}$ & $R_{\mbox{caught}}$ & $R_{\mbox{loiter}}$ \\ \midrule
        1 & $d_{m}$ & Fig.~\ref{fig:roadnet} & $0.7$ & $0.90$ & 13 & $[1000.0]$ & $[1:1:10]$ & 100 & 2000 & -2000 & -200\\
        2 & $d_{m}$ & Fig.~\ref{fig:med_roadnet} & $0.7$ & $0.95$ & 45 & $[2000.0]$ & $[1:3:28]$ & 1000 & 2000 & -2000 & -200\\
        3 & $p_{trans}$& Fig.~\ref{fig:roadnet} & $[0.0,1.0]$ & $0.95$ & 13 & $[1000.0]$ & $[8,3,1]$ & 1000 & 2000 & -2000 & -100\\
        4 & $p_{trans},e_{m}$ & Fig.~\ref{fig:roadnet} & $[0.0,1.0]$ & $0.95$ & 13 & $[10.0,1000.0]$ & $[8,3,1]$ & 1000 & 2000 & -2000 & -100\\
    \end{tabular}
    \vspace{-0.3cm}
\end{table*}

\noindent \textit{(i) Different Candidate Solvers:}
This evaluation involved experiments 1 and 2 from Table~\ref{tab:exps}. Candidate MCTS solvers of varying depths were used to simulate the ADT executing the task on each of the two road networks. In each case, the trusted solver \solvetrust{} was also evaluated. For experiment 1, \solvetrust{} was the $d_{m}=9$ solver, whereas \solvetrust{} was the $d_{m}=25$ solver in experiment 2. Recall that \surrogate{} was not used for this evaluation; instead the statistics for $p(\rwdapprox{})$ were simulated directly by multiple Monte Carlo runs for each solver and compared to those of \solvetrust{} via \IS{}. 

\begin{figure}[tbp]
    \centering
    \begin{minipage}[b]{0.95\linewidth}
        \centering
        \includegraphics[width=0.7\linewidth]{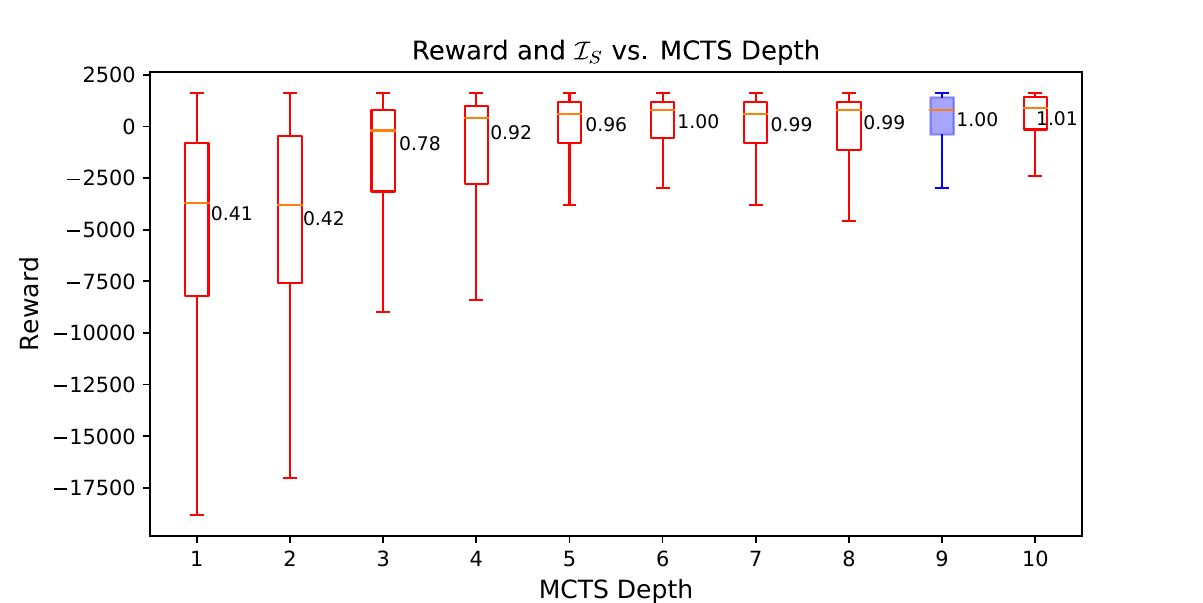}
        \vfill
        \subcaption{Experiment 1}
        \label{fig:mcts_d}
    \end{minipage}%
    \hfill
    \begin{minipage}[b]{0.95\linewidth}
        \centering
        \includegraphics[width=0.7\linewidth]{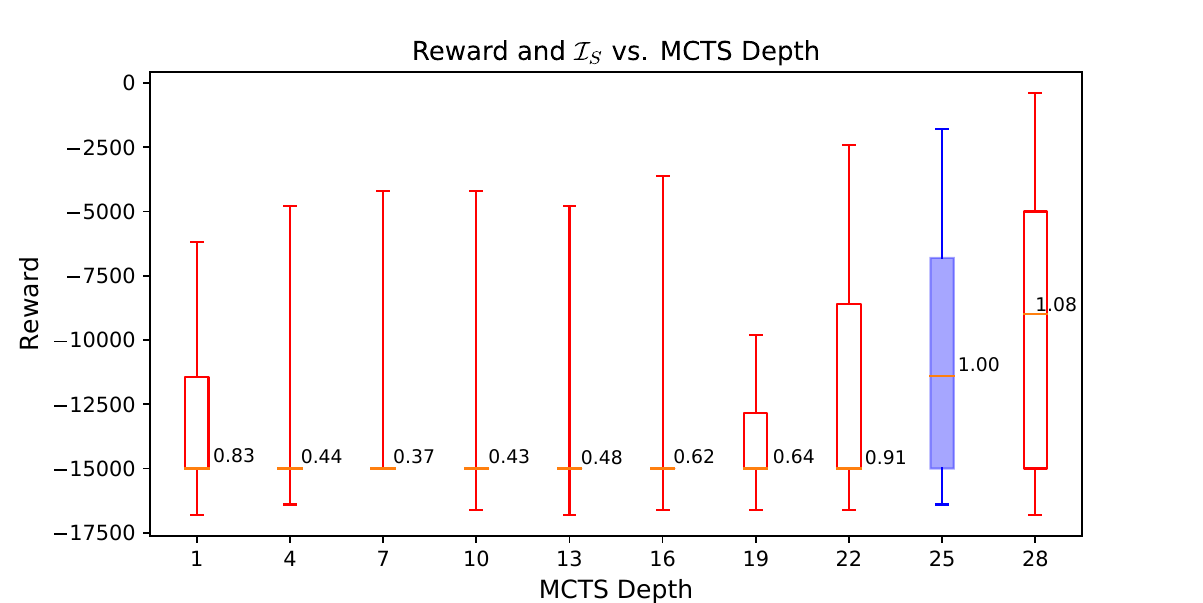}
        \subcaption{Experiment 2}
        \label{fig:mcts_d_med}
    \end{minipage} 
    \caption{Experiment results. \IS{} (SQ) is calculated w.r.t. \solvetrust{} highlighted in blue. }
    \vspace{-0.5cm}
\end{figure}

The results for experiment 1 are shown in Fig.~\ref{fig:mcts_d}. As expected, \IS{}$=1.0$ for \solvecand{}$=$\solvetrust{}. We also see that candidate solvers with $d_m= 6$ through $10$ perform about equivalently to \solvetrust{} (and each other), which indicates they are similarly capable of performing the delivery task. On the other hand, candidate solvers with $d_{m}=1$ through $3$ are much less capable than \solvecand{}. 

Experiment 2 results are shown in Fig.~\ref{fig:mcts_d_med}, where \solvetrust{} uses $d_{m}=25$. These results show that only solvers with $d_m=22$ and $d_m=28$ are similarly capable to \solvetrust{}. Note, the $d_{m}=1$ solver has $\IS=0.83$, possibly due to the fact that this solver makes decisions based on little foresight, whereas solvers with $d_m=4$ through $19$ have enough foresight to force the ADT to hesitate and accumulate negative rewards from not moving quickly to avoid capture (i.e. they tend to loiter longer).

An important insight from both sets of results is that, while the $d_{m}=9$ solver is very capable of completing the task for the smaller network in Fig.~\ref{fig:roadnet}, its performance does not extend to the larger network in Fig.~\ref{fig:med_roadnet}, i.e. the $d_{m}=7 \text{ and } 10$ solvers are in fact very incapable compared to the $d_{m}=25$ trusted solver, as reflected by the \IS{} values and as confirmed by the underlying reward distribution statistics. This demonstrates the useful potential for \IS{} to enable autonomous agents \cA{} to self-assess and report of discrepancies between expected vs. actual second-order competencies when taking on tasks in different contexts. The next two evaluations show how \IS{} can account for such context variations via \surrogate{}. 

\begin{figure}[tbp]
    \centering
    \begin{minipage}[b]{0.5\linewidth}
        \centering
        \includegraphics[width=\linewidth]{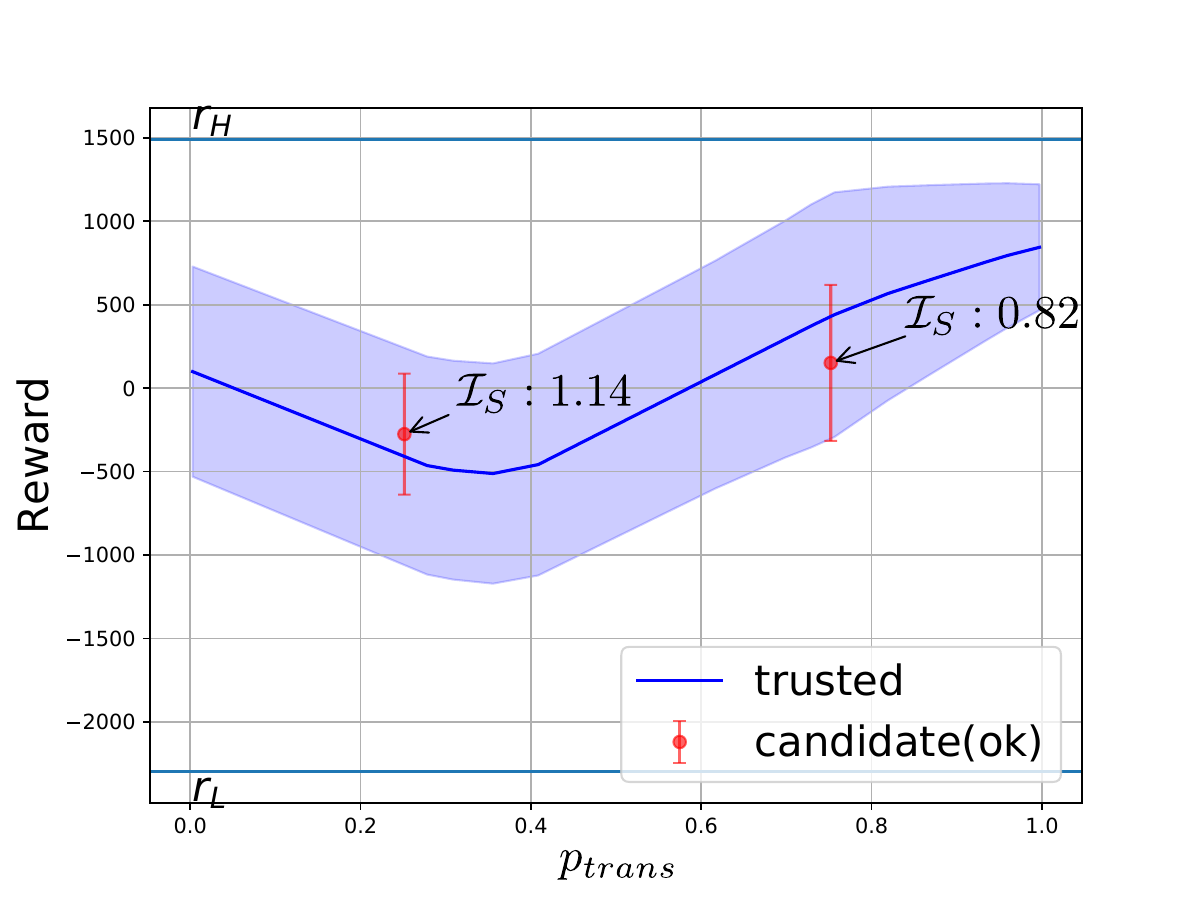}
        \vfill
        \subcaption{\solvecand{} depth 3}
        \label{fig:tprob_ok}
    \end{minipage}%
    \hfill
    \begin{minipage}[b]{0.5\linewidth}
        \centering
        \includegraphics[width=\linewidth]{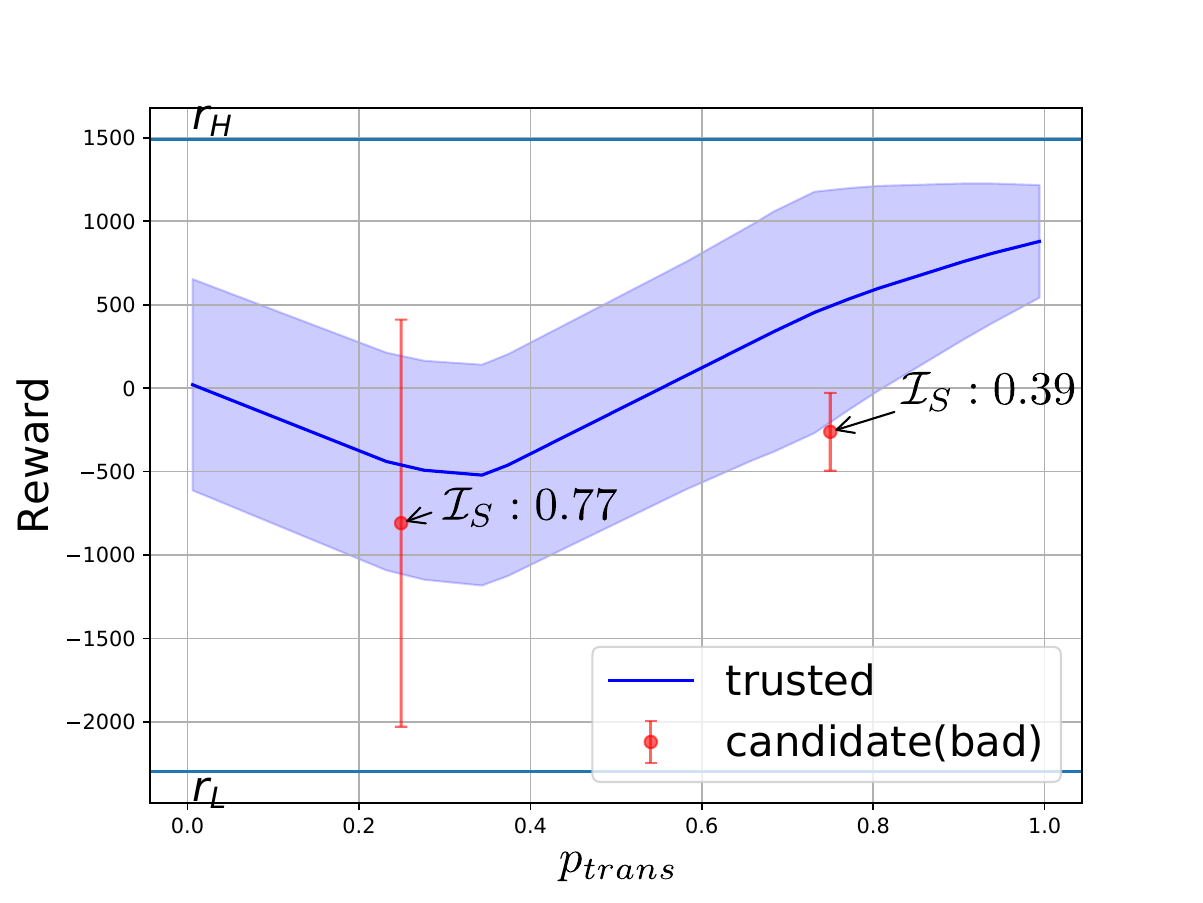}
        \subcaption{\solvecand{} depth 1}
        \label{fig:tprob_bad}
    \end{minipage} 
    \caption{Comparison of \solvecand{} to \solvetrust{} of depth 8. }
    \vspace{-0.5cm}
\end{figure}

\noindent \textit{(ii) Varying A Task Parameter:} Experiment 3 from Table~\ref{tab:exps} was used for this evaluation with the learned surrogate model \surrogate, where \solvetrust{} uses $d_m=8$, while the two candidate solvers use $d_m=3$ and $d_m=1$. Figure~\ref{fig:tprob_ok} shows the results for \solvecand{} with $d_{m}=3$ at two different values of $p_{trans}$. At $p_{trans}=0.25$, this \solvecand{} is slightly more capable than \solvetrust{}, whereas at $p_{trans}=0.75$ the solver \solvecand{} is slightly less capable. Figure~\ref{fig:tprob_bad} also shows the results for \solvecand{} with $d_{m}=1$ at the same values of $p_{trans}$. At $p_{trans}=0.25$ the candidate solver is moderately less capable than \solvetrust{}, whereas it is much less capable at $p_{trans}=0.75$. Note that the \IS{} values here account for differences in both the means and uncertainties in the reward distributions, e.g. such that the extra variance in the candidate reward pdf in the $d_m=1$ case for $p_{trans}=0.25$ incurs an additional penalty on top of the mean deviation. These results also comport with the results of the first two experiments above and show how the surrogate model easily extends them to perform additional `what if?' second-order competency analyses as a function of task context variables.   

\noindent \textit{(iii) Varying Task and Solver Parameters:}
Experiment 4 from Table~\ref{tab:exps} was used for this evaluation along with \surrogate{} where \solvetrust{} again uses $d_m=8$, while the two candidate solvers use $d_m=3$ and $d_m=1$, and both $p_{trans}$ and candidate solver settings for $e_{m}$ were simultaneously varied. 

\begin{figure}[htbp]
    \centering
    \includegraphics[width=0.9\linewidth]{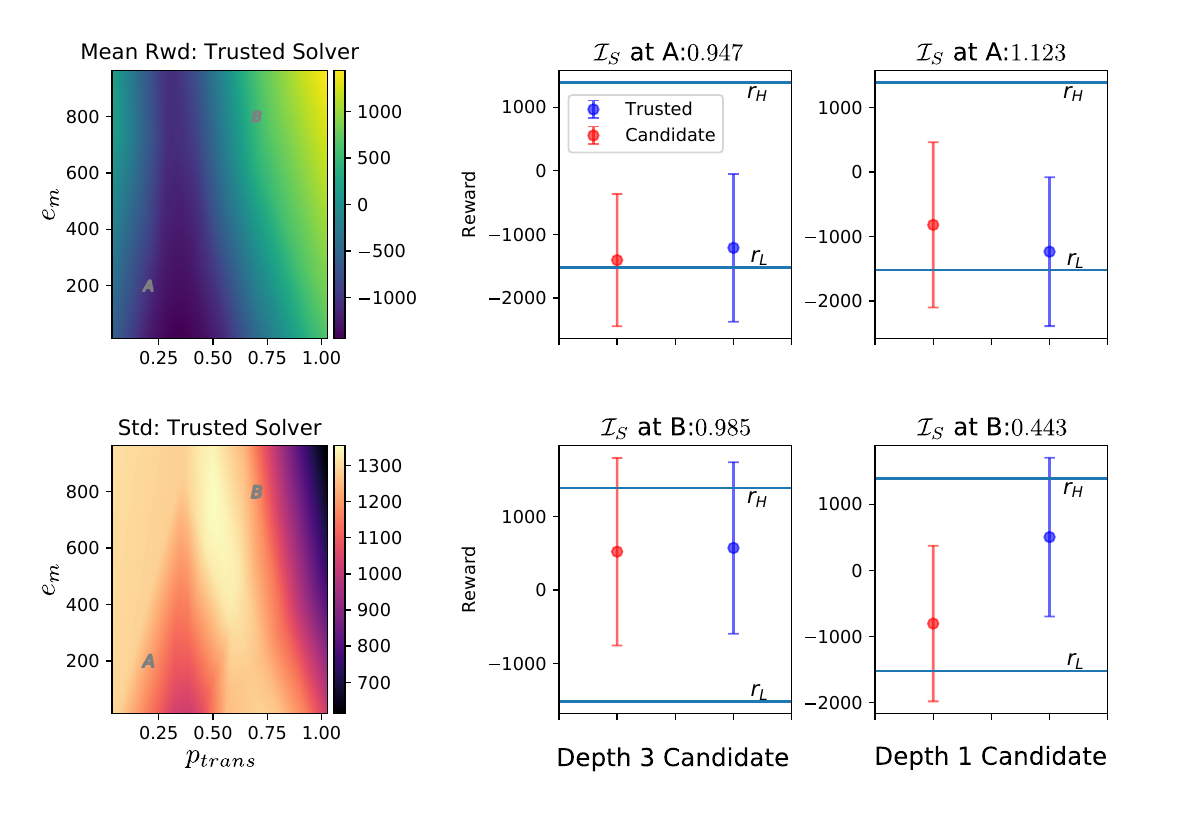}
    \caption{Comparison of \solve{} to \solvetrust{} of depth 8. The top-left shows the mean reward of the `Trusted Solver' \solvetrust, the bottom-left is the standard deviation of \solvetrust{}. Figures to the right show \IS{} at points A, and B for \solvecand{} of depth $d=3$ (middle) and \solvecand{} of depth $d=1$ (right). }
    \label{fig:tprob_emcts}
    \vspace{-0.5cm}
\end{figure}

Figure~\ref{fig:tprob_emcts} shows the results for \solvecand{} with $d_{m}=3$ (middle cloumn) at two different points of interest corresponding to different task and solver settings. At point A, \IS indicates that this \solvecand{} is slightly less capable than \solvetrust{}; \IS{} is similar at location B as well. Figure~\ref{fig:tprob_emcts} (right column) shows the results for the \solvecand{} with $d_{m}=1$ at the same task and solver settings. At location A, \solvecand{} is slightly more capable than \solvetrust{}, whereas at B the same \solvecand{} is moderately worse than \solvetrust{}. 

More generally, the \surrogate{}-predicted reward distribution means and standard errors for the depth 3 candidate solver in Fig.~\ref{fig:tprob_emcts} show how sensitive \solvetrust{}'s policy is to variations in both \task{} and solver settings. Beyond computing \IS{} at run-time, such information can provide useful sanity checks and non-obvious insights about \cA{}'s capabilities at design time. Indeed, the \surrogate{} shows that \solvetrust{} performs best as $p_{trans} \rightarrow 1$ and $e_{m}$ increases (as intuitively expected). On the other hand, the predicted standard error output of \surrogate{} shows that tasks for which $p_{trans}<0.9$ lead to significant volatility in solver performance, such that tuning $e_m$ may do little to improve behavior for $p_{trans}<0.65$. Hence, it may be desirable to consider tuning different variables in \solvetrust{} or else use a different class of solvers if \cA{} must handle tasks in this regime. 

\section{Other \famsec{} Factors and Practical Considerations}\label{sec:otherfactpract}
This section reviews several technical issues relevant to the development and adoption of \famsec{} for competency self-assessment in real-world autonomous systems. Firstly, we revisit the three remaining \famsec{} factors outside of \IO{} and \IS{} to highlight technical challenges and possible strategies for their formulation. Next, the interrelated non-independent nature of all five factors put forward by \famsec{} is considered, along with a look at how the process for pre-task deployment competency self- assessment can be adapted to evaluation contexts both during and after task deployments. We then discuss how the validity and reliability of \famsec{}-based competency reports can be concretely and objectively evaluated. Lastly, we explain how our implementations for MDP agents can be generalized to other kinds of algorithmic autonomous decision-making approaches. 

\subsection{Other Problem-Solving Statistics and Factors}
\label{ssec:otherfactors}
Recall that \IO{} and \IS{} represent two of the five \famsec{} factors, with \IM{} (model and data validity), \IA{} (alignment with user intent), and \IH{} (historical performance and experience) making up the remainder. By modeling autonomous decision-making agents as MDP problem solvers, we have shown how \IO{} and \IS{} can be computed and communicated in human-understandable terms, leveraging problem-solving statistics obtained directly from probabilistic artifacts of the MDP modeling and solution process. What does this all imply for formulating \IM{}, \IA{}, and \IH{}? 

As mentioned in Sec. 3, some \famsec{} factors yield more readily than others to this approach. The further up the chain of algorithmic decision making in Fig. \ref{fig:famsec} one goes, the more challenging it becomes to identify competency standards and problem-solving statistics that can quantitatively be compared. 
Indeed, there are more ways to assess agent behavior in the solver and execution blocks on the basis of variables like states, rewards, etc. than in other parts of the process, which are largely shaped by user intent and designer judgment. That is, the \emph{selection} of rewards, models, states, actions, etc. in an MDP serves as abstract mathematical distillation of user intent and designer knowledge/expertise. 
Hence, measuring `abstraction quality' for an MDP or policy execution becomes a more challenging and involved process. 
This can also be appreciated by noting that \IS{} and \IO{} respectively provide meta-cognitive assessments of a given problem-solving process and its execution, taking the models coming out of an apriori modeling process for granted - whereas \IM, \IA, and \IH{} provide meta-cognitive assessments of different components of the modeling process itself, independent of the subsequent solution and execution processes. 
We next consider how this distinction translates to possible implementations for each of the three latter factors, emphasizing that further work along these lines is needed. 
%

\subsubsection{Model and data validity \IM:}
This factor assesses how well observations and events predicted by \cA{} line up with \cA's experiences in reality. For an MDP-based agent \cA, this corresponds to how well its probabilistic model encoded by the tuple $(S,A,T,R)$ explains \cA's experiences during actual task executions. From a statistical point of view, if \cA's experienced states $s_t$, actions $a_t$, and rewards $r_t$ are recorded as execution trace logs $l(i)= \{s_{t,i},a_{t,i},r_{t,i}\}_{t=0}^{K_i}$ for each task instance $i$, then evaluation of \IM{} can be framed directly in terms of a goodness of fit problem. That is: given a set of trace logs $L(N_K) = \{l(i)\}_{i=1}^{N_K}$ for $N_K$ task execution instances, what is the probability that $L(N_K)$ was generated by \cA's MDP model 
$(S,A,T,R)$? Note that the competency standard $\Sigma$ here corresponds to some criterion for model acceptability, and 
need not depend on whether \cA{} uses an optimal policy $\pi^*$. Moreover, if it can be taken for granted that $S$, $A$, and $R$ are defined consistently with \cA's actual implementation (i.e. such that no $l(i)$ contains uninterpretable `out of bounds' data), then the goodness of fit question reduces to validating the probabilistic state transition model $T$ against $L(N_K)$ (where training data could also be generally included in $L(N_K)$). 

Several established goodness of fit measures are available in cases where $T$ can be explicitly specified as a fully observable Markov chain \cite{titman2008general,titman2010model,weiss2018goodness}. Related measures of model-data agreement also exist which generalize to partially observable and hidden Markov models, as well as other more general probabilistic models; these include performance and uncertainty metrics \cite{marcot2012metrics}, evidence conflict scores \cite{Kim-UAI-1995}, and
the aforementioned surprise index \cite{Zagorecki-AAAIFS-2015}. In cases where $T$ cannot be directly identified, the formulation of goodness of fit tests remains an open challenge. This is most relevant in high-dimensional settings, where $T$ either is implicitly defined through simulation-based forward models (often used for MCTS in discrete action spaces) \cite{kochenderfer2015decision}, or is approximated via deep neural nets or other learning-based regression methods for probabilistic model predictive planning in continuous action spaces \cite{acharya2022competency,acharya2023learning}. 

A drawback of the statistical interpretation of \IM{} is that it requires $L(N_K)$ to be available. Thus, for instance, the ADT in the Doughnut Delivery problem cannot perform a goodness of fit test ahead of its first delivery problem in a new road network if $(S,A,R,T)$ is specified by fiat. 
In such cases, it may be more useful to adopt an alternative view of \IM{} which establishes whether or not $(S,A,R,T)$ satisfies specific properties for safe and efficient planning. This could for instance be implemented through a variety of formal methods (discussed later) to ensure a priori that the physical state and action abstractions embedded in the MDP allow \cA{} to find some set of actions that reach/escape particular states with some minimum probability. Possibilities here include multi-scale fidelity measures \cite{sarkar2019multifidelity}, e.g. for quantifying `sim2real' gaps in simulation-driven model learning \cite{akella2022test, neary2023multifidelity}. 

\subsubsection{User intent alignment \IA:} This factor interrogates the agreement between the algorithmic formulation guiding \cA{}'s execution and user intentions for a given task. 
In an MDP setting, this amounts to determining the confidence that
\cA's $(S,A,R,T)$ specification adequately aligns with a user's mental task model. 
All aspects of casting a given task as an MDP can be considered, including the definition and meaning of states, actions, dynamics, and sources of uncertainty, as well as the specification of the reward function.
In practical terms, this means \IA{} should evaluate two separate but related aspects of the MDP problem formulation process: 
(i) \emph{coverage}: to what extent do the components of a given MDP form a necessary and sufficient description of all relevant aspects of a task?; and 
(ii) \emph{correctness}: to what extent are the state-action preferences encoded by a given MDP aligned with user preferences? 
Note that this generalizes the idea of assessing objective function alignment \cite{brown2021value,sanneman2023transparent,bobu2024aligning,bobu2018learning}, which is an important special case of \IA{} evaluation that implicitly only considers the correctness aspect. That is, if a task is in fact completely described by $S,A,$ and $T$, then assessment of \IA{} reduces to determining how well $R$ and the resulting set of MDP value functions $V(s)$ and $Q(s,a)$ (for some discount rate $\gamma$ and policy solution strategy) reflect user state-action preferences. 
As such, techniques for assessing objective function alignment with respect to a given MDP \rev{ -- including inverse reinforcement learning \cite{ziebart2008maximum,Sadigh-RSS-17} --} could be readily adapted to formulate correctness indicators for \IA, where \cA's utility would be assessed against a user's true utility as a competency standard. \footnote{Recall also that both components of \IA{} are also distinct from the interpretation of \IO{} as a meta-utility, i.e. the utility of \cA{} being tasked to achieve specific outcomes according to \cA's own utility.}

On the other hand, coverage indicators are more challenging to devise and remain an important open area for future development, as there is no straightforward way to assess the necessity and sufficiency of MDP components for a particular task. This is related to the much larger and deeper problem of determining necessary and sufficient representations for AI and learning agents \cite{vanotterlo2009logic, delgrande2023dagstuhl}. Nevertheless, in the context of competency assessment, some desirable properties of such indicators can still be identified as potentially useful starting points. For instance: such indicators should have reasonable properties with respect to composition, abstraction, and refinement of representations, e.g. the addition of new relevant states should improve coverage, while compression/removal of such states should deteriorate it. 
Moreover, since reward, utilities, and state space formulations are all non-unique for any given problem, any coverage indicator must abide an infinite number of MDPs which provide equally good task descriptions. 
That is, any coverage indicator must be invariant to transformations between any two `functionally equivalent' MDP model state spaces. 
%

What might computable coverage indicators for \IA{} even look like? The most basic questions to be examined by a hypothetical approach are whether: (1) an MDP model has the required \emph{capacity} to describe a user's underlying goals and objectives, regardless of the MDP's values for transition probabilities, rewards, etc., and (2) all user goals and objectives are in fact sufficiently represented, i.e. if capacity is sufficient, is it being used to maximum effect? 
While we are unaware of any techniques that can currently address these questions, we speculate that formal model verification and learning-based approaches will offer powerful tools to do so in the future. For example, learning-based language models could be used to build user interfaces to elicit/verify goals and objectives \cite{yu2023language}. Deep learning techniques are also quite capable of identifying useful patterns in large data sets that form features of interest in different learning problems. This indeed has been demonstrated numerous times for both model-free and model-based deep reinforcement learning \cite{geladaPMLR2019}, as well as in deployments of large language models. 
Such powerful feature-learning architectures could therefore conceivably be used to evaluate the representational capacities of candidate MDPs in a number of different ways. 
For instance, if $T$ represents an acceptable state transition model for describing task \task, then $T$ could be encoded via a deep net whose input layers capture essential features forming task states, whence distances to other candidate MDP models $T'$ could then be derived, e.g. using model-based adaptations of techniques described in \cite{winqvist2023optimal, tan2022renaissance}. Alternatively, distance measures between models $T$ and $T'$ belonging to some specific parametric family (or families) could be learned directly.  

\subsubsection{Historical performance and experience \IH:} 
This factor allows \cA{} to assess self-confidence based on prior experiences, and provides a basis for meta-memory and meta-learning capabilities that can improve self-confidence assessments (whether or not \cA{}'s models or decision-making process relies on learning techniques). \IH{} is distinct from \IM{} and \IS{} in that \IH{} leverages \emph{actual} experiences attempting to complete tasks using \cA{}'s task model and policy together, whereas \IM{} and \IS{} gauge these components separately from each other \footnote{Though, as discussed in the next subsection, \IS{} implicitly depends on the model formulation.}. Also, whereas \IM{} and \IS{} are specific to the task at hand, \IH{} allows \cA{} to leverage experiences completing other \emph{related tasks} (knowledge transference), as well as the experiences of \emph{other agents} attempting to complete similar tasks. 
As such, \IH{} 
can pertain to many other useful problem-solving variables and  statistics that are not already reflected in \cA{}'s models, policies, utilities, and outcomes. 

In the Doughnut Delivery problem for instance, suppose high fuel prices force the MG to potentially adopt one of two fuel-saving chase strategies, where they either: (a) quit chasing the ADT after a fixed number of steps, or (b) wait near particularly difficult-to-escape locations to ambush passing vehicles. 
This information is not captured by the MDPs described earlier for the Doughnut Delivery problem, and in practice such kinds of information may not be known in advance to the designers of the ADT's decision making algorithms. Yet, such knowledge may become available through the ADT's other related tasks delivering other items in the same/different settings, or through another ADT's or delivery company's experiences. Depending on when the information is acquired (before, during, or after deployment), the ADT could re-calibrate/re-focus its self-confidence assessments knowing that its designed model and policy may not fully account for other MG behaviors. 

\IH{} can also allow \cA{} to bootstrap self-confidence factor evaluations. For instance,  a meta-surrogate model for \IH{} could be learned to exploit dependencies between other factors (described next) or to make predictions for \IH{} based on margins between expected values for other confidence factors and their online computed values. \IH{} could also be used to modulate/override other factors like \IS{} and \IM{} in certain cases, since these depend more on model-based predictive assessments. 

\subsubsection{The Interrelated Nature of the \famsec{} Factors} \label{sec:IO-limitations}
As stated earlier, one assumption for calculating \IO{} is that the optimal policy \policyopt{} is assumed to be available. However, \policyopt{} is rarely available for problems with large state-action spaces or critical time limitations. Generally, as assumptions are relaxed about \famsec{} factors for more realistic applications, the more apparent their complex interdependencies become.

\begin{figure}[tbp]
    \centering
    \begin{minipage}[b]{0.5\linewidth}
        \centering
        \includegraphics[width=\linewidth]{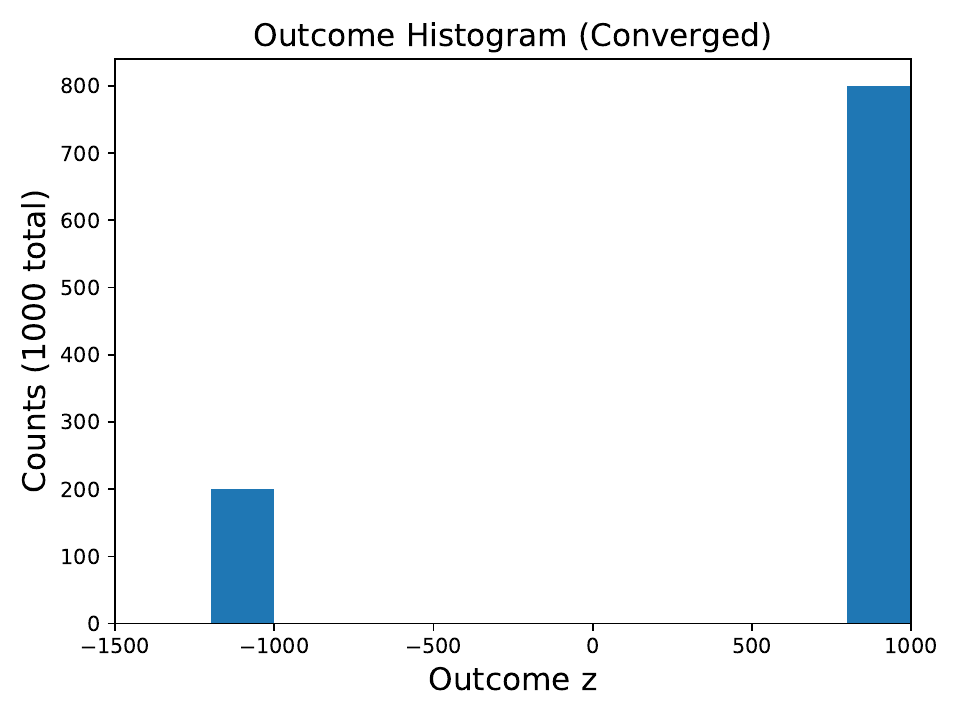}
        \vfill
        \subcaption{Following converged policy (i.e. $\pi\approx\policyopt$)}
        \label{fig:IOconverged}
    \end{minipage}%
    \hfill
    \begin{minipage}[b]{0.5\linewidth}
        \centering
        \includegraphics[width=\linewidth]{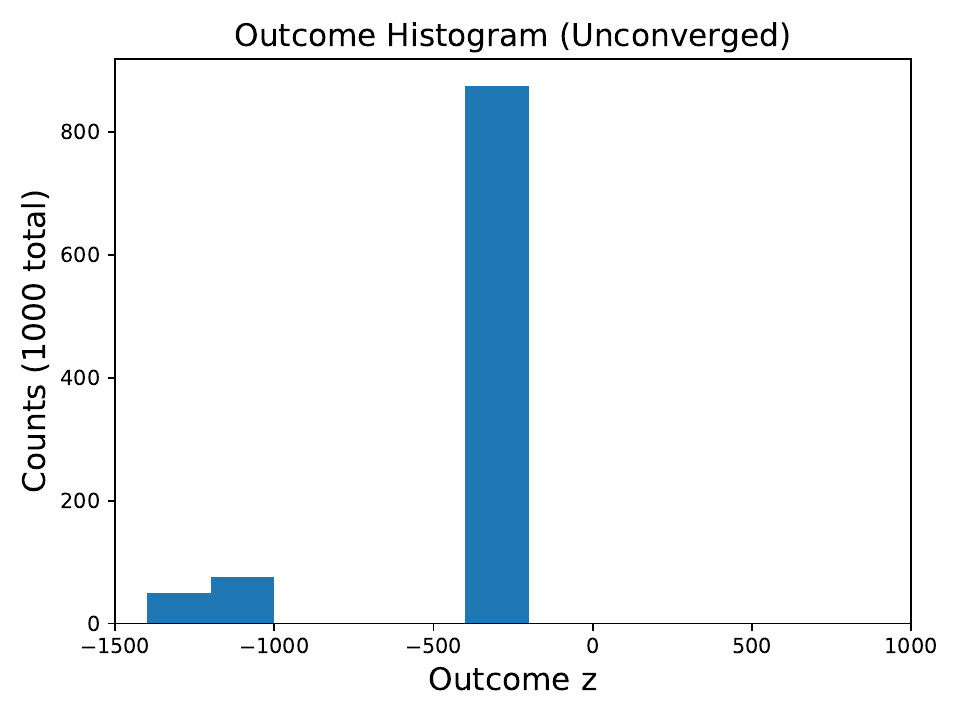}
        \subcaption{Following unconverged policy (i.e. $\pi\napprox\piopt$)}
        \label{fig:IOunconverged}
    \end{minipage} 
    \caption{Reward histograms when following a converged and unconverged policy. Results from Monte-Carlo simulations using Environment (c) in Fig.~\ref{fig:exprdist_env}.}
    \vspace{-0.5cm}
\end{figure}

Figures \ref{fig:IOconverged} and \ref{fig:IOunconverged} illustrate a circumstance in which \IO{} breaks down in the latter case, due to the situation where the policy produced by the solver was not able to converge fully. In both figures the task is navigating `environment (c)' from Figure~\ref{fig:exprdist_env}.
Notice that all of the reward for Fig.~\ref{fig:IOunconverged} is below zero. In this case $\IO=-1$ \emph{is} the correct assessment of the predicted outcome. However, this assessment does not reflect the real cause of the poor performance due to the policy not having converged during run-time. Such cases should be treated distinctly from other causes such as, say, the specific layout of the road-network being extremely difficult for the ADT to navigate. In such cases, \IO{} and \IS{} (at a minimum) should be provided jointly to establish proper context for the competency assessment, 
since generally speaking a shortfall in any one  \famsec{} factors can dominate \cA{}'s competency assessment, even while other factors remain unchanged or appear to indicate high self-confidence. It is also interesting to note the `asymmetric' nature of the relationships between the various factors, such as \IO{} and \IS{}. For instance, referring back to the example in Fig. \ref{fig:IOconverged} and \ref{fig:IOunconverged}, the \IO{} results can be trusted only as long as \IS{} reflects good confidence in the quality of the underlying policy, whereas \IO{} does not influence the trustworthiness of \IS{}. 

\rev{In some cases factors can be straightforwardly ignored or even combined with one another. For example, \IS{} is not relevant if an exact optimal policy can be produced for a sufficiently small/simple MDP, or else \IS{} may be directly combined with \IM{} if certain parameters of a state transition model are unknown and can be inferred online by \cA{}. 
Conlon, et al. \cite{conlon2024event} also describe how \IO{} and a version of \IM{} (based on the surprise index) can be used in combination for mobile robot exploration tasks to trigger online competency self-evaluations in changing environments. 
While relationships and combinations with other factors like \IH{} and \IA{} could also be theoretically examined and exploited, the aforementioned technical challenges for implementing these make it difficult to describe these possibilities in detail. 
All this being said, it is nevertheless worth considering how the `summary' factor $\cI_{F}$ could be instantiated. While this still remains largely notional, several possibilities abound for implementation, the suitability of which all depend heavily on the specific task and needs for such a summary indicator. For instance, a simple possibility is that all 5 \famsec{} indicators are stacked into a single vector-valued report (i.e. report all the relevant indicator values directly -- this was done in \cite{Israelsen2019-xm} for cases just considering \IO{} and \IS{}). Another simple possibility is to define $\cI_{F}$ as the `worst case' indicator. This could correspond to the min among the five indicators or another approach that accounts for the asymmetric nature of these factors.
}


\subsection{Other Practical Considerations}

\subsubsection{\rev{Computational Complexity} and A Priori, In Situ, and Post Hoc Assessment}
Until now, it has been implicitly assumed that \cA{}'s self-confidence assessment occurs in an offline manner prior to attempting its assigned task. In theory, \famsec{} factors could also be evaluated during or after task execution. 
The process of evaluating each factor considered here would in theory look very similar to its corresponding a priori version, with the main differences coming from the source of certain required data inputs and the computation effort required for evaluation. For example, whereas a priori evaluations of \IO{} and \IS{} would rely on model-simulated performance data, in situ and post hoc evaluations of these factors would leverage real execution traces to compare against expected model outcomes. 

\rev{
For a priori competency assessment, our \IO{} indicator has run time complexity $O(N_sT_{max})$, where $N_s$ is the number of generative (Monte Carlo) probabilistic world model simulations run under a given policy and $T_{max}$ is the maximum number of steps required for \cA{} to complete the task. 
Our \IS{} indicator's run time complexity is as follows for the various phases of operation. For generating training samples for the surrogate model, the complexity is $O(N_{s}^{Tr} T_{max}^{Tr})$ , where $N_{s}^{Tr}$ is the number of simulations performed for the trusted solver and $T_{max}^{Tr}$ is the maximum number of steps required for the trusted solver to complete the task in any given condition. For the surrogate model regression, the complexity is $O(c[N_s^{Tr}]^f)$, where $c$ and $f$ are determined by the surrogate model regression technique (e.g. naïve Gaussian process regression gives $c=1$ and $f=3$, whereas ANN regression via stochastic gradient descent gives $c=$ \# parameters $\times$ \# iterations and $f=1$); and $O(N_s^{Can} T_{max}^{Can})$ for online calculation, where $N_{s}^{can}$ is the number of simulations performed for the candidate solver and $T_{max}^{Tr}$ is the maximum number of steps required for the candidate solver to complete the task in any given condition. 
The complexity of a task or environment for a particular MDP problem primarily imposes itself in these scaling relationships via the $T_{max}$ and $N_s$ variables. $T_{max}$ depends on the nature of the task as well as (to some degree) the specific solver being used. 
For instance, with an approximate MCTS policy solver, $T_{max}$ correlates with the size of the action space and degree of reward sparsity, which both determine the amount of exploration required for \cA{} to approximate an optimal policy. The degree of sensitivity to different operating contexts is also reflected in the $N_s$ sample sizes, i.e. more data is needed to predict and evaluate competence if outcomes vary considerably in different situations.
}

The computational cost of offline a priori assessments could also be reduced via online in situ implementations using a number of strategies such as event-triggered reasoning, whereby online \IO{} assessments are only instantiated if a significant deviation from a priori outcomes is detected. This type of check is useful in uncertain dynamic environments where a priori information can become stale.  
For example, in the Doughnut Delivery problem, \IO{} can be recomputed on the fly if a low surprise index value is observed for the ADT and MG joint state history during execution, indicating highly unlikely state transitions that are not well explained by the MDP dynamics model. 
See \cite{ConlonACMS2024} for further details on this concept for event-triggered outcome assessment. McGinley \cite{mcginley2022thesis} also explored an alternative computation reduction strategy for long-term predictive trajectory resampling that recycles and augments previous \IO{} evaluations via an Approximate Bayesian Computation (ABC) Monte Carlo sampling strategy. 

\subsubsection{Applying \famsec{} to other model-based decision-making algorithms} \label{ssec:otherdecalgs}
As mentioned in Sec. \ref{ssec:deliveryMDP}, the ADT problem could also be framed as a partially observable MDP (POMDP), mixed observability MDP (MOMDP), or even as a model-based reinforcement learning problem (e.g. if transition parameters for the ADT and MG were unknown). In any of these cases, the same \famsec{} assessment strategies developed for the MDP could also be adapted and applied with few or no modifications. For instance, the surrogate modeling approach for evaluating \IS{} via a trusted solver would work in almost exactly the same way for a POMDP policy as an MDP one, where an online POMDP policy could be formulated using an extension of MCTS to POMDPs such as the POMCP algorithm \cite{silver2010monte}. Likewise, \IO{} can be evaluated in the same way for any POMDP or model-based RL problem using sampling-based rollouts under a prescribed a solver or policy \cite{aitken2016thesis, conlon2022generalizing, acharya2022competency}. GOA and surprise index formulations of \IO{} and \IM{} have also been applied to robotic
motion planners such as RRT \cite{conlon2024event}, showing that \famsec{} can be easily adapted to non-MDP-based decision-making algorithms, \rev{including those involving continuous as well as discrete state and action variables}. 

\subsubsection{Assessing the Validity of \famsec{} Indicators}
\label{ssec:validity}
The fact that an agent possesses high (or low) self-confidence says nothing on its own about the \emph{validity} of a particular competency self-assessment or the agent's ability to maintain its assessed competency level.
\rev{That is, there is an important difference between the agent's \emph{perceived} competency and its \emph{actual} competency}. 
This distinction has a notable psychological analog: the Dunning-Kruger (D-K) effect describes a well-known cognitive bias in which individuals possessing limited competence in a particular area overestimate their abilities at a task \cite{dunning2011dunning}. The D-K effect is often cited as the source of introspective disconnect observed in people who perform a job poorly but nevertheless project great self-confidence while doing so. 
As such, it motivates the idea that competent performers should strive to maintain accurate assessments of their own limitations. 

How can competency statements be validated for machine agents to avoid self-confidence biases akin to the D-K effect? 
It is worth noting that \famsec{} has two built-in features which can help mitigate such issues. 
Firstly, a hierarchical system of implicit internal checks exists for each indicator, whereby the reliability of \IO{} is  dependent on \IS{} (per the example above), the reliability of \IS{} and \IO{} are both dependent on \IM{}, etc. Thus, potentially `overconfident' lower-level indicators can be called into question if higher-level indicators are less confident. 
Secondly, each indicator's output is grounded by an externally provided competency standard, which allows the goalposts to stay fixed. 
However, these features only provide limited protection against unwarranted machine self-confidence, since the data generated to evaluate the indicators are still largely sourced from the agent's own rendition of the task, and thus may be untrustworthy to start with \footnote{Per Dunning \cite{dunning2011dunning}: the lack of required knowledge and competency is the very source of the D-K effect, i.e. knowledge/competency are required to know whether or not one is sufficiently knowledgeable/competent.}. 
\rev{It is also theoretically possible to design an MDP agent to `optimize' competency or exploit loopholes in reward function logic according to external standards, e.g. this could happen inadvertently if competency standards are only based on rewards and the agent's reward functions happen to favor those standards.} 
Objective measures of predictive capability forecasting and post-hoc decision-making performance are therefore also needed to establish the accuracy of self-confidence assessments. 

For example, DARPA's Competency Aware Machine Learning (CAML) program called on performers to specify metrics describing key aspects of competency assessment validation \rev{for agents that \emph{did not} use competency reports to improve their task performance}. \rev{Metrics} included: coverage (are all influences on competency captured?), correctness (is the impact of each influence known?), fidelity (are competency levels predicted accurately?), and reliability (does agent maintain stated level of competency?) \cite{russell2023alpaca}. 
To address the fidelity requirement in particular, refs. \cite{conlon2022generalizing,acharya2022competency, acharya2023learning} showed how probability-based proper scoring rules such as the Brier score \cite{ferro2007comparing} could be used with actual or simulated task performance data to gauge the validity of \IO{} indicators based on the dUPM/dLPM and Omega ratios. The Brier score provides a mean-squared error metric measuring the average differences between predicted probabilities of desirable task outcomes (each given a number between 0 and 1) and actual task outcomes (expressed as a 1/0 binary outcome). The more accurately an agent predicts task outcomes based on its task model and solver/policy, the lower the Brier score for \IO{} becomes (with 0 as the best possible score). 

Note that although the Brier score is a proper scoring rule (i.e. discourages hedging in probabilistic predictions), it says nothing about whether the agent performs the task well. Rather, it describes how well \IO{} captures uncertainty in actual performance outcomes. 
Hence, an ADT that persistently exhibits either over-confidence or under-confidence in predicting its ability to evade the MG (gets captured more/less often than its sampled outcome distribution predicts) will obtain a higher Brier score than one whose confidence is calibrated correctly to the true uncertainty of the problem (gets captured about as often as predicted by the sampled outcome distribution).   The Brier score could also be adapted to validate other indicators like \IS{} and \IM{} if appropriate ground truth data is available to compare relevant sets of predicted vs. `ground truth' outcomes, although the application of other proper scoring rules and probabilistic measures of accuracy for these and other \famsec{} indicators is an important open area of work.  

\subsubsection{Validating Competency Reporting with Human Users}
Finally, we may consider the effectiveness of machine self-confidence reporting for conveying competency assessments to human users and evaluators. Fully addressing this non-trivial problem requires deeper and more careful consideration of cognitive human factors that are beyond the scope of this paper. However, we note that controlled human participant studies have been conducted to evaluate the efficacy of communicating \famsec-based machine self-confidence statements to aid human supervision and delegation of autonomous decision-making robots. For instance, refs. \cite{israelsen2020machine, Israelsen-thesis} used a between-subjects study design to examine how the presence/absence of \famsec{} reports affects human supervisory performance in a simple event-driven simulated version of the Doughnut Delivery problem. Human supervisors recruited through Amazon Mechanical Turk ($N=255$) were shown randomly generated maps of delivery problems with different goal locations and start points for the ADT and MG\footnote{See Appendix A.}, and then were tasked with deciding whether to authorize or cancel the delivery task based on semantically translated \IO{} and/or \IS{} indicator reports generated for an MCTS-based MDP ADT planning agent (with the absence of both indicators for the same agent in each delivery instance being the control group condition). 
Participants were compensated and thus incentivized to perform well based on the number of delivery runs they approved that resulted in successful ADT task completion without capture by the MG, while they were heavily penalized for approving delivery runs resulting in ADT capture by the MG and lightly penalized for canceling delivery runs. 
It was found that supervisors in the group which received \IO{} and \IS{} reports performed significantly better (completed more delivery runs on average) than supervisors in other groups. 
Participants in the same group also reported slightly higher levels of trust in the autonomous planning agent than in the other groups. The small effect size for increased trust indicated reasonable calibration of supervisor expectations in relation to actual robot abilities based on observed agent performance for the difficult delivery task, as opposed to superficial increases in trust due to the mere presence of algorithmic assurances. 
Human participant studies in \cite{conlon2022iros} (which also compared \famsec{} generated indicator reports to randomly generated reports) and \cite{conlon2024event} (which examined a hardware-based implementation with observable physical consequences for task successes and failures) revealed similar findings when using \famsec{}-based reports to inform supervisors of robotic planning competencies in uncertain navigation and search tasks. 



\section{Related Work}\label{sec:related_work}

\subsection{Competency Self-Assessment in Relation to Other Algorithmic Assurances}
Our approach to algorithmic competency self-assessment has several conceptual links within the growing literature on algorithmic assurances for human-interactive autonomous systems \cite{Israelsen2019-xm}. 
Of particular note here are works on transparent, interpretable, and explainable AI (collectively referred to here as XAI) \cite{ali2023explainable,kaur2022trustworthy,dwivedi2023explainable,minh2022explainable},
algorithmic metareasoning \cite{russell1991principles, herrmann2023metareasoning}, and formal methods for verification and validation \cite{luckcuck2019formal}. 

Conlon, et al. \cite{ConlonACMS2024} describes competency self-assessment as a special problem class within the broader domain of XAI. Specifically, the justification aspect of XAI is very important for making effective algorithmic assurances that engender appropriate levels of human trust in AI agents \cite{malle2021multidimensional}. 
Recalling Figs. \ref{fig:CompAssessGenBlockDiag} and \ref{fig:famsec}: by comparing predicted/observed results to specified competency standards in different tasking contexts,
competency self-assessment (and machine self-confidence in particular) allows an agent to 
justify whether or not it believes it can achieve desired objectives and behaviors. 
However, whereas approaches to XAI generally aim to make the autonomous decision-making process more transparent or understandable, e.g. by explaining each selected action in some logical manner, competency assessment does not necessarily require this. Indeed, \famsec{} presumes agent's underlying algorithmic decision making process can be taken for granted and thereby only seeks to determine whether that process is adequate for competent task execution, without attempting to explain how any particular decision or action in an execution is selected. 
This is in contrast to certain XAI methods such as evaluation of critical states in MDP-based agents \cite{huang2018establishing} or post-hoc causal descriptions of actions derived from executed policies \cite{elizalde2009generating}, which implicitly assume that an agent can accomplish assigned tasks but must also be able to justify each step of how they do so. With MDP agents, it is still often possible to glean agent competency information through other XAI approaches such as these. For example, critical states indicate those situations in which an agent's utility encounters the most sensitivity to sub-optimal actions, and thus can be analyzed to identify bottleneck situations requiring extra attention. 
However, such insights are limited by the extent to which the reward and utility functions are correctly specified 
\emph{and} by which an optimal policy covers all critical states in the first place. 
The possibility (and generally high probability) that neither set of conditions holds in practice is a key driver behind \famsec{}'s meta-analytical approach. 

As such, competency self-assessment also occupies a niche within the landscape of computational methods for metacognition and metareasoning. Like \famsec{} and other competency self-evaluation frameworks described later, metacognitive computing methods allow machine agents to maintain explicit awareness of their limited reasoning and execution capabilities within the inherent confines of bounded rationality. However, the chief aim of metacognitive architectures is usually to provide an agent with self-reflective feedback `in the moment' that permits behavior adaptation for improved task performance. 
This is exemplified by metacognitive approaches to reinforcement learning \cite{grant2018recasting, krueger2017enhancing}, where the learner seeks to learn a higher-level policy for selecting among lower-level policies, based on how lower-level policies perform on a given task in different situations. 
In contrast, competency self-assessment does not necessarily seek to adapt agent behavior. Moreover, metacognitive self-improvement processes do not require the agent to render an accurate human-interpretable assessment of its expected range of capabilities (as assessments which only serve to guide local improvements to performance are sufficient), and again often presume that an agent \emph{should by design} (despite bounded rationality) already be able to perform a task.
Similar remarks also apply to meta-reasoning approaches for other kinds of autonomous capabilities, e.g. introspective perception \cite{grimmett2016introspective, rabiee2023introspective,daftry2016introspective}. 

Formal methods provide yet another powerful set of tools to examine whether a decision-making algorithm allows an agent to competently fulfill its tasks. 
Whereas assessments produced by competency evaluation frameworks like \famsec{} are empirically oriented, automated analysis of task specifications and decision-making algorithms expressed via logical formalisms -- such as linear temporal logic (LTL) -- can provide hard binary `yes/no' guarantees on task outcomes and agent behaviors in deterministic settings. 
Many sophisticated formal approaches also provide mechanisms for synthesizing correct by construction decision-making algorithms that guarantee satisfaction of all task specifications when it is possible to do so. This is enabled, for instance, by some LTL-based robotic task planning frameworks \cite{raman2012explaining, raman2013sorry}, which can also automatically generate examples and explanations of agent behaviors that violate one or more specifications when they are not all achievable. 
Formal methods can thus provide many useful insights to users and designers alike into the competency limits of decision-making agents. Yet, despite much progress in their development and use, formal methods remain challenging for typical users/designers to understand,
as they impose a non-trivial translation burden between human-understandable problem descriptions and precise formal logic specifications \cite{cherukuri2022towards}. 
The computation required for these methods also does not scale well in complex problem settings, particularly those involving stochastic uncertainties such as the Doughnut Delivery problem. In such cases, hard binary property guarantees are often relaxed, e.g. to chance constraints and the like, though conservative approximations and abstractions often must also be introduced for computational tractability. 
Since \famsec{} avoids many of these issues, we view it and formal analysis as complementary `stool legs' that together can support grander future schemes for algorithmic competency self-assessment\footnote{Consistent with the argument made in \cite{Israelsen2019-xm} on the need for both soft and hard algorithmic assurances.
}. 
\rev{For instance, recent progress has been made in the formal methods community to address these limitations for stochastic problem settings by leveraging confidence analysis for probability-based verification of safety properties \cite{chou2020predictive,ruchkin2022confidence, nakamura2023online}. Such techniques could potentially provide a useful foundation to establish more formal guarantees and performance bounds through machine self-confidence indicators that are currently lacking in our formulation of \famsec{}}.

\subsection{Other Algorithmic Competency Assessment Frameworks}
Conlon, et al. \cite{ConlonACMS2024} survey state of the art algorithmic competency self-assessment techniques, which fall into three broad categories: test-based methods, learning-based methods, and knowledge-based methods. Test-based methods leverage probabilistic distance metrics and related concepts from classical statistical testing. Learning-based methods apply supervised, unsupervised, or reinforcement learning strategies to identify agent competency models in different tasking contexts. Knowledge-based methods rely on expert-encoded data structures such as databases and ontologies to reason about agent competency. Although \famsec{} can primarily be thought of as a test-based method, it also applies learning strategies (e.g. for \IS{} solver quality assessment surrogate modeling) and expert knowledge encoding (e.g. to define competency standards), making it something of a hybrid method in practice per the categorization of Conlon, et al. 
We briefly review a representative subset of techniques across these categories
to highlight the most significant similarities and differences from \famsec{}.

Among test-based techniques, the explicit assumption alignment tracking concept of \cite{Cao-TRO-2023, Gautam-ICRA-2022} was already described earlier in Sec. \ref{ssec:strategies} to motivate \famsec{}'s use of probabilistic indicator functions for collectively validating decision-making algorithm design assumptions (rather than validating each individual assumption explicitly). 
Other works have likewise considered assumption-validating probabilistic metrics in a variety of other contexts. For instance, Guyer and Diettrich \cite{Guyer-AAAIFS-2022} propose a conformal prediction technique for MDP-based agents, which provides a calibrated probability estimate that the cumulative reward for a given time horizon falls within a desired target range prescribed by the user. 
This is similar to our \IO{} outcome and \IS{} solver quality assessments, although their method does not rate policy solver performance against a competency standard. 
In line with the previous discussion on the model quality indicator \IM{}, Zagorecki, et al. \cite{Zagorecki-AAAIFS-2015} apply the surprise index as means of assessing goodness of fit for Bayesian network models, while Kaipa, et al \cite{Kaipa-AAAIFS-2015} develop a residuals-based goodness of fit metric for visual object identification for robotic bin-packing tasks. Ramesh, et al. \cite{Ramesh-HRI-2021} developed a framework around `robot vitals', which assesses changes in entropy in a set of physical signals that indicate potential drops in a mobile robot's performance. 
Compared to \famsec{}, these other methods collectively either consider only a single aspect of decision-making performance (e.g. outcome assessment or model quality, but not both), or are tied to variables not directly related to decision-making algorithm competency. 

In the realm of learning-based methods, Svegliato, et al. \cite{Svegliato-IROS-2019} and Basich, et al. \cite{Basich_2020,basich2020learning} developed reinforcement learning strategies for competency awareness in POMDP-based and MDP-based agents, respectively. These works extend the state space of such agents to effectively metareason about nominal vs. off-nominal operating conditions on the basis of operational experiences, so that different levels of autonomy and human intervention can be selected to maintain appropriate online behaviors. Human feedback is also used in the learning process to improve agents' ability to reason about their competencies in different operating conditions. Although these works explicitly focus on meta-reasoning for MDP and related decision-making agents operating under uncertainty, their main objective is to improve agent performance for a single problem. As such, these methods are tailored for agents to optimize behavior and assess competencies on a specific task, rather than evaluate and communicate competency for a range of possible tasks. 

Knowledge-based frameworks provide a variety of ways for autonomous agents to encode expert prior knowledge about tasks and capabilities. 
Frasca, et al \cite{frasca2020can, frasca2022framework} describe a framework which applies pre-conditions and post-conditions to logically reason about possible task completion and failure outcomes, and uses probabilistic likelihoods for these conditions to assess competencies before, during, and after tasks in a manner similar to our \IO{} outcome assessment indicator. Their approach allows agents to answer basic queries about task steps, as well as reason about expected durations and failure points. 
Infantino et al. 2013 \cite{infantino2013humanoid} developed a documentation-based semantic ontology for introspective reasoning for a humanoid robot, enabling it to address queries related to its internal processes and functional capabilities. Aguado et al \cite{Aguado-Sensors-2021} develop an ontological metacognition framework for an underwater vehicle system. Similar to the RL-based MDP and POMDP augmentation methods described above, their approach allows the vehicle to perform self-diagnosis and reconfiguration to ensure resiliency for long-term autonomous operations with minimal human intervention. 
These methods all provide sophisticated forms of introspection at potentially multiple levels of system operation (planning, perception, etc.) and can support different types of user/designer querying, dialog, and feedback. However, these capabilities also require dedicated reasoning architectures and data structures to encode task-specific knowledge bases, which typically do not translate or scale well to decision-making tasks with a potentially wide range of complex uncertainties like the Doughnut Delivery problem. 


Lastly, we point out developments on the concept of machine self-confidence which preceded and inspired \famsec. 
Hutchins, et al \cite{Hutchins-HFES-2015} coined the original term `machine self-confidence' to describe an indicator-based implementation for alerting users to gaps in an agent's decision-making and perception capabilities during task execution. Their approach relies on designer expertise to hand code indicator values defining machine self-confidence for different system implementations, which can then be displayed graphically to users/supervisors. Sweet, et al. \cite{Sweet-SciTech-2016} extended this idea into a hierarchical algorithmic self-assessment process for planning agents that reason about the probability of task success by considering first-order and second-order uncertainties. Along similar lines, Kuter and Miller \cite{ Kuter-AAAIFS-2015} developed a symbolic planning framework for closed-world agents to express self-confidence in task execution via counter-planning, such that agents which devise robust plans with the more contingencies (i.e. healthier planning margins) are more self-confident about ensuring success; this is aligned with the motivation behind indicators \IO{} and \IS{}, in particular. 
These latter works generically formulate machine self-confidence as a combination of probability of task success and the sensitivity of this probability to uncertain data directly which directly informs it. They do not identify or stipulate formal indicators or competency standards for uncertain factors in a decision-making agent's reasoning process, and so describe a more abstract version of \famsec{}'s structured algorithmic machine self-confidence hierarchy 
\footnote{It is also interesting to note the parallels between our hierarchical notion of machine self-confidence and neuro-cognitive computational models of self-confidence, which seek to explain human metacognitive abilities \cite{ackerman2017meta,fleming2017self,maniscalco2012signal,grimaldi2015there,pouget2016confidence,fleming2024metacognition}; in both cases, confidence is distinguished from likelihoods or probabilities via second-order uncertainty reasoning, i.e. reasoning about uncertainty in uncertainties.}.

\section{Summary and Conclusions} \label{sec:concs}
{\em ``A man's GOT to know his limitations."} \\
-- ``Dirty'' Harry Callahan, \emph{Magnum Force} (1973)
\bigskip

\noindent {\em ``All models are approximations. Assumptions, whether implied or clearly stated, are never exactly true. All models are wrong, but some models are useful. So the question you need to ask is not `Is the model true?' (it never is) but `Is the model good enough for this particular application?' ''} \\ 
--George E.P. Box, Alberto Luceño, and María del Carmen Paniagua-Quiñones, \emph{Statistical Control By Monitoring and Adjustment} (2nd ed., 2009)
\bigskip


To summarize, this paper examined how autonomous algorithmic agents can self-assess their decision-making competencies. 
Markov Decision Processes (MDPs) were reviewed as a prototypical class of computational problems for decision-making under uncertainty, through which algorithmic agents are commonly designed and deployed in various robotic and other autonomous systems. An example robotic vehicle delivery task was provided to ground, motivate, link, and illustrate key concepts for algorithmic decision-making under uncertainty and competency self-assessment. 
These concepts generally stem from considerations of bounded rationality and the brittleness of design assumptions in real-world deployment of computational agents. 
A formal framework for algorithmic competency assessment of autonomous agents was presented, emphasizing that competency indicators (\cI) can generally be used to assess both \emph{what} tasks agents (\cA) can accomplish and \emph{how} they accomplish them as a function of outcomes of interest (\cO), agent states, and environment states (\cS) in different contexts and agent settings (\cC), relative to competency standards (\cSig) defined by an evaluator. 

Building on these ideas, the concept of machine self-confidence---a form of meta-reasoning based on self-assessments of an autonomous agent's knowledge about the state of the world and itself, as well as its ability to reason about and execute tasks---was used to define the Factorized Machine Self-confidence (\famsec) framework for competency reporting. \famsec{} is an engineering-focused means for computing and communicating competency self-assessments in terms of five interrelated indicators that reflect algorithmic design confidence factors: outcome assessment (\IO), solver quality (\IS), model quality (\IM), intent alignment (\IA), and historical performance and experience (\IH). 
These competency indicators are informed by problem-solving statistics, which are tied to probabilistic variables and distributions inherent to an agent's algorithmic decision-making process. Margins of exceedance for these artifacts relative to each competency factor's standards are quantitatively rated via meta-utility functions to automatically reflect the agent's (vis-a-vis the designer's) degree of self-confidence in different contexts, which in turn can then be transformed for human-interpretable communication. 

Expected trends and behaviors of each \famsec{} factor were illustrated with several hypothetical delivery task examples, showing that each indicator captures different sensitivities to context-specific variables. Theory and practical implementation details for computing outcome assessment and solver quality factors in model-based MDP agents were each discussed at length and illustrated for different MDP variations of the autonomous delivery task. 
In the case of outcome assessment, meta-utilities were derived via upper partial moment/lower partial moment ratio analysis of simulated probability distributions for non-discounted cumulative rewards, using stipulated minimum acceptable reward outcomes to define task outcome competency standards. Generalizations to distributions of arbitrary task outcomes were also presented. 
For solver quality, meta-utilities were derived from modified Hellinger distances between non-discounted cumulative reward distributions for candidate policy solvers and trusted (competency standard benchmark) policy solvers, where the distributions for the latter can be generalized to different evaluation contexts via learning-based surrogate models. 
Other practical considerations and issues for applying and developing \famsec{} further---including the computation of other factors, validation of self-reported competency statements, and applications beyond MDP agents--- were also discussed.
Finally, a review of related work was presented.

\subsection{Limitations}
We consider here the limitations of our approach. Indeed, many of the features which make our formulation of machine self-confidence and \famsec{}-based competency self-assessment straightforward to implement hinge on several important assumptions. For example, our implementations of \IO{} and \IS{} are heavily reliant on the availability of full task outcome predictions. While it generally possible to obtain these in  model-based MDP problems using empirical sampling (as described earlier) or analytical techniques (e.g. conformal prediction \cite{Guyer-AAAIFS-2022}), such predictions are much more difficult to obtain in model-free decision-making settings, e.g. for direct value function or optimal policy identification in reinforcement learning. 
Moreover, meaningful competency standards for different indicators must be provided externally by human evaluators. We can foresee instances where ensuring that evaluators possess sufficient knowledge and capability to do this in an unbiased objective manner could be a major undertaking. 

Our approach also solely focuses on self-assessment of autonomous algorithmic decision-making capabilities, and in so doing implicitly assumes that other important capabilities like perception can in some sense be taken for granted. 
We emphasize that this assumption is only a first step scaffolding toward deeper insights that (hopefully and eventually) lead to a more general free-standing framework for competency assessment of boundedly rational autonomous agents, which fully embraces the interdependent nature of agent competencies in the real world.
The need for such a capability in the long run is plainly evident, for instance, in the extension from MDP problems to partially observable MDPs (POMDPs), where optimal decision making policies are generally described via functions of entangled stochastic action-observation histories. While the decision-making competencies of a POMDP-based agent could still be analyzed via the currently proposed \famsec{} indicators, such an analysis would not account for limitations in the agent's perception capabilities which drive the POMDP's observation and belief state update processes. 
Despite this, our approach does (for better or worse) naturally comport with the simpler and more popular heuristic system design philosophy which treats sense-making, reasoning, decision-making, and execution as distinctly siloed processes\footnote{This could also be seen as a heuristic extension of the separation principle from control theory \cite{Georgiou-CDC-2012} or of the observe-orient-decide-act (OODA) loop \cite{brehmer2005dynamic}.}. 
Even in this realm, however, more work needs to be done to understand how insights rendered by \famsec{} for decision-making competency self-assessment can be coupled with other techniques for introspectively assessing capabilities like perception. 

Also, unlike other algorithmic introspection techniques, our framework currently does not leverage feedback from machine self-confidence reports to adapt system behavior. However, this reflects a deliberate choice to restrict the scope of work here to only consider the foundational elements for human interpretable and actionable algorithmic competency self-assessment and reporting\footnote{E.g. the DARPA CAML program also restricted scope in this manner.}, rather than a fundamental limitation of \famsec{} or machine self-confidence, per se. For example, an initial foray into applying \IO{} factor analysis for autonomous behavior adaptation was considered by \cite{mellinkoff2020towards}, for adapting and focusing the training behavior of simulated lunar rovers over uncertain terrain via model-based reinforcement learning. 
The results there showed promising potential for the ability of \famsec{}-based trajectory analysis to help identify specific model parameters and operating conditions that improve or degrade expected agent performance during training. However, the practicality of such an approach for real world navigation tasks under uncertainty depends on the availability of high-fidelity environment models, sufficient computing resources, and persistent long-term operability (i.e. so that undesirable events such as slipping down a crater wall or colliding with a boulder do not jeopardize long-term operation). 
\rev{The possibility of coupling self-confidence assessments to mechanisms for autonomous self-improvement in such safety critical situations also underscores the need to better understand whether/how \famsec{} can be used to obtain formal guarantees and performance bounds, e.g. using verification tools, conformal uncertainty quantification, et cetera}.

Lastly, like all statistically-oriented methods for introspection, \famsec{} inherits various blind spots and insensitivities to unknown unknowns from the different kinds of problem-solving statistics and analyses described here. The nature of these is not only task-dependent, but also more fundamentally depends on the sufficiency of the agent's model of its own decision-making process and information available to compare against it. Put in simpler terms: an autonomous algorithmic agent (can at best) only detect whatever it is designed to look for. 


\subsection{Implications}
Technology experts are frequently asked whether they believe the capabilities of artificially intelligent and autonomous machine agents will soon live up to their creators' expectations (e.g. see \cite{goldmanSachs2024}). 
These discussions reflect a shared anxiety hovering over financiers, employers, government officials, workers, and other 
\rev{interested parties}, who (in one way or another) are seeking justification for the huge sums of capital, time, and attention being lavished on such technologies. 
While interest in these systems appears likely to persist, it is also clear that such seemingly simple questions cannot be meaningfully answered in broad strokes, even for limited classes of agents. 
Rather, as we have argued for decision-making agents acting under uncertainty, rigorous engineering-based evaluation methodologies and standards are needed (at a minimum) to objectively assess specific competencies on a case by case basis and clearly communicate these to stakeholders. 
As such, we underscore the key implications of our work for users, designers, and other researchers to consider. 

From a theoretical perspective, even though it is not the same as competency, machine self-confidence provides a useful indicator of competency, insofar as it explicitly considers both \emph{what} things a boundedly rational agent can accomplish as well as \emph{how} it accomplishes them.   
This raises the point that machine self-confidence is generally \emph{not} the same as probabilistic likelihood of success, but rather encodes a more nuanced assessment of how models, data, information, preferences, and actions are holistically processed by an agent to arrive at and act on particular probabilistic assessments. Indeed, a high probability of task success could be the result of a high probability of barely achieving desirable outcomes under an optimal policy, which may or may not translate to high design confidence depending on the evaluation context and standards. This speaks to the need to jointly examine the interaction in sensitivities to both design margins and different sources of uncertainties, including those originating from an agent's imperfect reasoning and information processing mechanisms as well as from the task environment. 

From a practical implementation and engineering perspective, computing and reporting machine self-confidence imposes fairly lightweight requirements that are easily handled by most existing autonomous algorithmic agent design schemes. This is especially true for algorithmic decision-making agents built on probabilistic MDP solvers and the like, as these naturally furnish the artifacts needed to evaluate problem-solving statistics within the \famsec{} framework. As such, machine self-confidence evaluation mechanisms can be readily designed into or retrofitted on top of a broad range of systems without disrupting their designs or capabilities. \rev{While our proposed methods may not offer a suitable `out of the box' competency self-assessment solution in every conceivable situation,} there is also ample opportunity to build on the \rev{competency standards and} indicator functions described for each of the five \famsec{} self-confidence factors, as well as to extend or reformulate the factors for different variations of the generic algorithmic decision-making architecture considered here. 

\begin{acks}
The authors wish to thank Dr. Nicholas Conlon for providing feedback on the manuscript. Support for this research was provided by: the Center for Unmanned Aircraft Systems (C-UAS), a National Science Foundation Industry/University Cooperative Research Center (I/UCRC) under NSF Award No. CNS-1650468 along with significant contributions from
C-UAS industry members; an Early Stage Innovations grant from
NASA’s Space Technology Research Grants Program, under NASA grant 80NSSC19K0222; Defense Advanced Research Projects Agency (DARPA) under Contract No. HR001120C0032; RTX; and a gift from Northrop Grumman. Any opinions, findings and conclusions or recommendations expressed in this material are those of the authors and do not necessarily reflect the views of DARPA, NASA, Northrop Grumman, NSF, or RTX.
\end{acks}

\bibliographystyle{ACM-Reference-Format}
\bibliography{bibliography}

\appendix
\section{Appendix: Monte Carlo Tree Search for MDP Policy Approximation} \label{sec:app_MCTS}
\begin{figure}[hb]
\centering
\includegraphics[width=0.45\columnwidth]{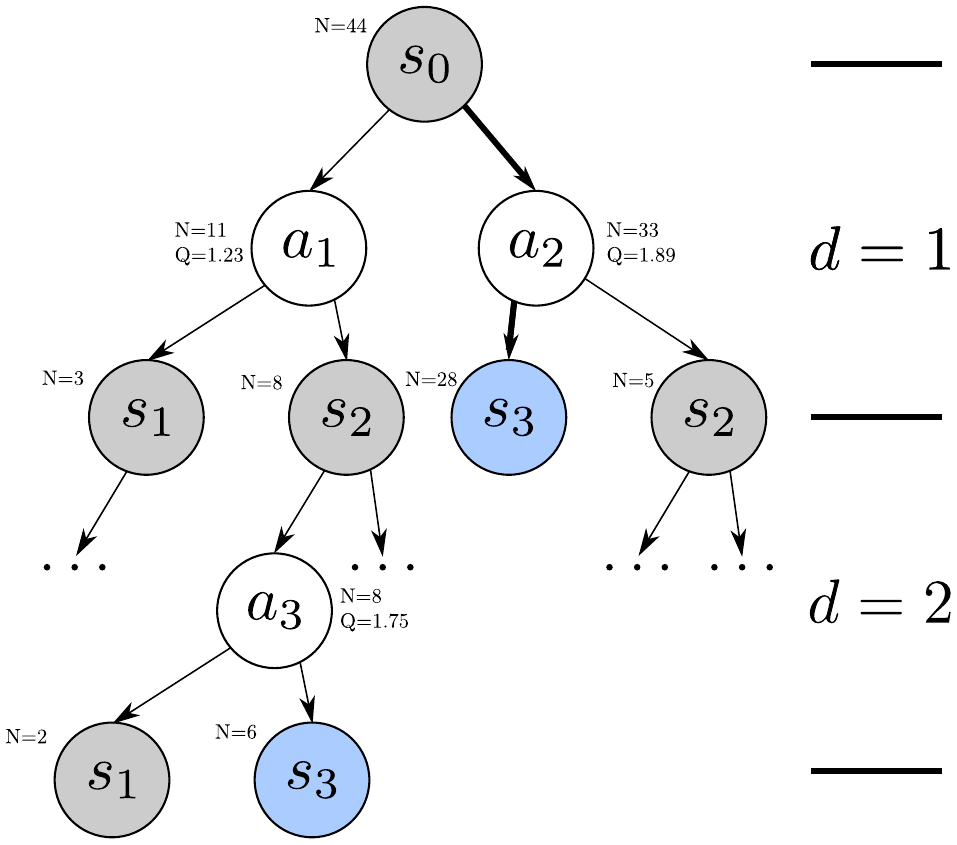}
\caption{Simple MCTS example ($d = 2, N = 44$), where $s_0$ is the initial state, and
$s_3$ is a desired terminal state. Path denoted by thick black arrows (i.e. select $a_2$) is the
final result for $\pi^*(s_0)$ obtained from MCTS.}
\label{fig:mctsexample}
\end{figure}

As shown in Figure \ref{fig:mctsexample}, the core idea behind MCTS is to perform a local search through the state-action space beginning at some initial state $s$. An action $a$ is sampled -- typically via an upper confidence bound explore/exploit metric tied to an initial estimate of $V^*(s)$ (or $Q^*(s,a)$, where $Q$ is the state-action value function) -- and the agent simulates subsequent states, actions, and rewards. These results are then used to update an estimate for $V^{*}(s)$ via backpropagation; subsequent search simulations are then repeated in this manner until a termination criterion is met and an actual action $a$ to execute is selected starting from state $s$. An attractive feature of MCTS is that it can yield a policy $\hat{\pi}$ that is arbitrarily close to $\pi^*$ given a suitably parameterized search and sufficient computation time. 

In particular, MCTS requires three search parameters: $N$, the total number of iterations in which to refine estimates of $V^*(s)$; $d$, the depth of the resulting Monte Carlo state-action-reward tree, which is the same as the number of actions to be considered into the future for a single simulation starting from $s$; and $c$, which governs the trade-off between exploration and exploitation in the metric for evaluating actions that yield the highest utilities starting from $s$. As with value iteration, the action for the approximate policy $\hat{\pi}$ at state $s$ is found via equation (\ref{eq:optmdppol}) once the desired estimate of $V^*(s)$ is available (based on convergence criteria, or samples $N$ taken, or time constraints, etc.). 

\section{Appendix: Training \surrogate{} for the ADT Problem} \label{sec:app_learning_surrogate}

\subsubsection{Road Network Training Data Generation}
Referring to Table \ref{tab:params} for MDP parameter definitions, experiment data was generated by creating random networks within the following bounds: $N = [8,35]$ (based on compute limitations), and $p_{trans}=[0.0,1.0]$. The other parameters are as follows: $\gamma=0.95$, $\emcts=1000.0$, $\itsmcts=1000$, $\rwdexit=2000.0$, $\rwdcaught=-2000.0$, and $\rwdsense=-200.0$\footnote{Soure code is available at \url{https://github.com/COHRINT/FaMSeC}}. The the delivery destination (Goal) and starting positions of the ADT and the MG were randomly selected from the available nodes in the network (i.e. road intersections).

Each road network was created by uniformly randomly selecting a value for $N$ in the given range, then uniformly randomly selecting one of the following graph generation methods: 1) \emph{Watts-Strogatz} \cite{chen2007watts};
2) \emph{expected degree} \cite{chung2002connected}, where the expected degree was a Normal distribution with mean 4 and standard deviation 1; 
3) \emph{Erd\"{o}s-R\`{e}yni} \cite{channarond2015random}
specifying $N$ and $E$ (number of edges) with $E = 2$; and 
4) \emph{static scale free} \cite{li2005towards}
with $E = 2$ and $\alpha_{out}=2$. All graphs were required to be connected. 
A total of 500 road network task instances were generated this way, and \IS{} and \IO{} were calculated for each one using the trusted solver \solvetrust{}. The \solvetrust{} in this case was a Monte-Carlo Tree Search (MCTS) solver with depth 3, and exploration constant 1000, which was determined to be able to best cope with the widest variety of possible delivery tasks under modest computing constraints. 

Figure~\ref{fig:raw_data} plots these measure for 341 of the 500 generated networks. The other 159 networks were excluded due to the edge-to-node ratio being `too high' (i.e. greater than 2.5, making the networks difficult to cognitively process because of the density of edges\footnote{these data were also used for a human participant study described in \cite{Israelsen-thesis} and a forthcoming companion paper}), the distance to the delivery destination being `too low' (i.e. the ADT being right next to the Goal), or the distance to the motorcycle gang being `too low' (i.e. the ADT starting right next to the MG). The networks are plotted in a scatter plot based only on their \IS{} and \IO{} values. Blue dots represent a `successful' delivery (i.e. final reward greater than or equal to 0) and the orange dots represent a `failed' delivery (final reward less than zero). Thus, each colored dot represents the outcome of a \emph{single draw} from the expected reward distribution \rwdapprox{} of a single road network problem ($N$ and $p_{trans}$, ADT starting location, MG starting location, and delivery destination), that was solved by 1000 MCTS iterations via \solvetrust{}. 
The presence of blue dots in the bottom left corner of the plot indicates unlikely, or `lucky', successes (see Figures~\ref{fig:lucky} and \ref{fig:unlucky}) 
It is interesting to point out that this plot was not used while the formulas for \IS{} and \IO{} were being developed; rather, this plot was made only \emph{after} \IS{} and \IO{} were finalized. The fact that the density of successes is higher when both \IS{} and \IO{} are high is an indicator that they are in fact performing their function and conveying the desired information.

\begin{figure}[htpb]
        \centering
        \includegraphics[width=0.6\linewidth]{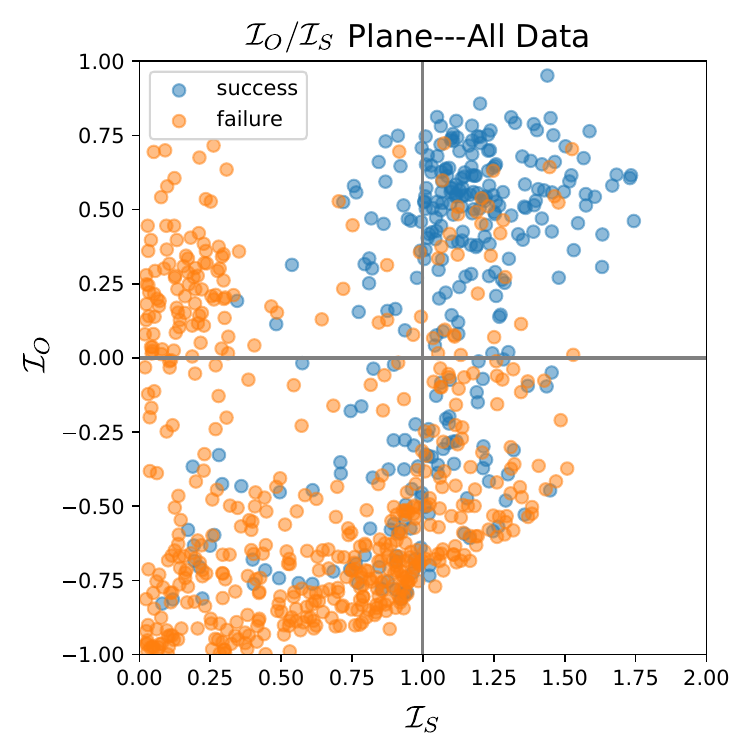}
        \caption{Plot of \IO{} vs. \IS{} for randomly generated road networks, with resulting single task attempt outcomes obtained via \solvetrust{}.}
        \label{fig:raw_data}
\end{figure}
\begin{figure}[tbp]
    \centering
    \includegraphics[width=1.0\linewidth]{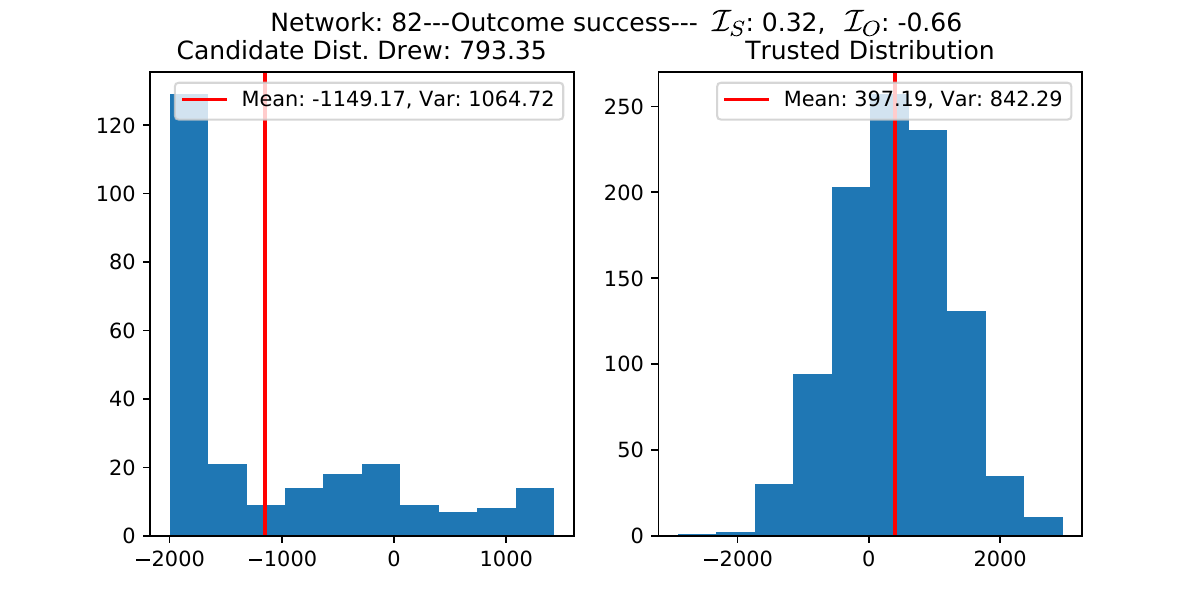}
    \vfill
    \caption{Example of a case in which the policy was `lucky' during a simulation run. (Right) histogram of `Trusted' reward distribution. (Left) Histogram of `Candidate' rewards distribution. In this case the simulation drew an outcome with $\rwdapprox{} = 793.35$ resulting from a successful delivery.}
    \label{fig:lucky}
\end{figure}

\begin{figure}[tbp]
    \centering
    \includegraphics[width=1.0\linewidth]{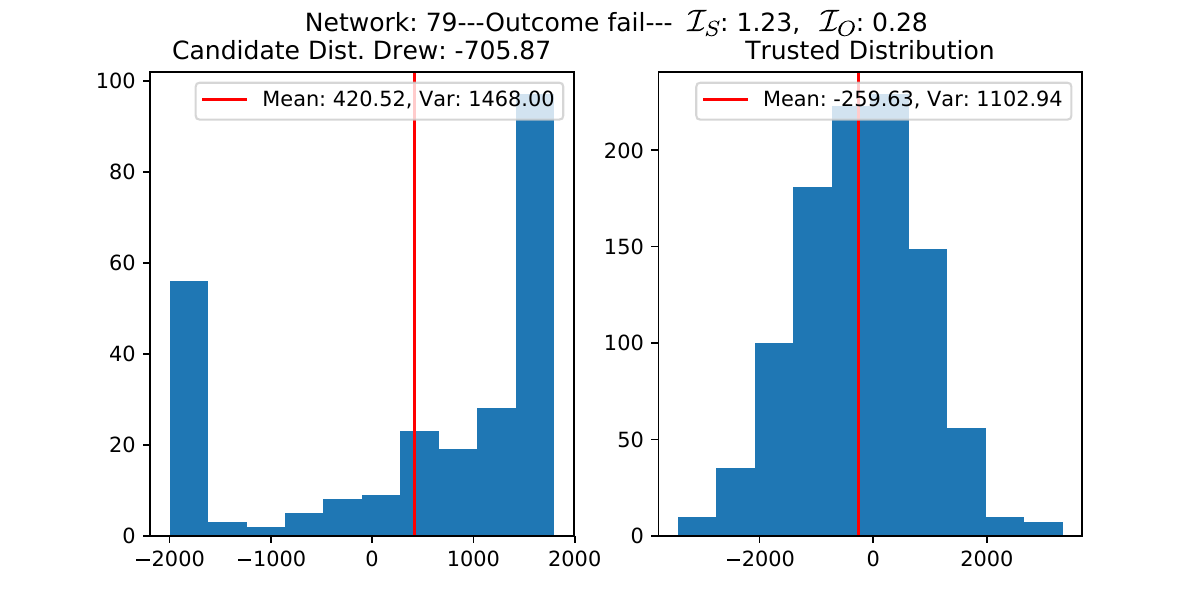}
    \vfill
    \caption{Example of a case in which the policy was `unlucky' during a simulation run. (Right) histogram of `Trusted' reward distribution. (Left) Histogram of `Candidate' rewards distribution. In this case the simulation drew an outcome with $\rwdapprox{}= -705.87$ resulting from an unsuccessful delivery.}
    \label{fig:unlucky}
\end{figure}

\subsubsection{Training \surrogate{}:} 
The following two parameters of interest serve as inputs $\bm{x}$ to the surrogate model \surrogate: 1) The transition probability ($p_{trans}$), and 2) $N$ (the number of intersections in the road network). Figure~\ref{fig:raw_data} plots the total reward \rwdapprox{} obtained by applying \solvetrust{} once to each of randomly generated road network problems. Generally, there is a positive correlation between transition probability and expected reward, and a negative (somewhat weak) correlation between $N$ and expected reward. These plots show that the outcomes are quite uncertain. Note that other task features such as the distance of the initial ADT and MG locations to the Goal, the connectedness of the road network, etc., are also likely to be relevant but were not used here. 

    \begin{figure}[htpb]
        \centering
        \includegraphics[width=0.65\linewidth]{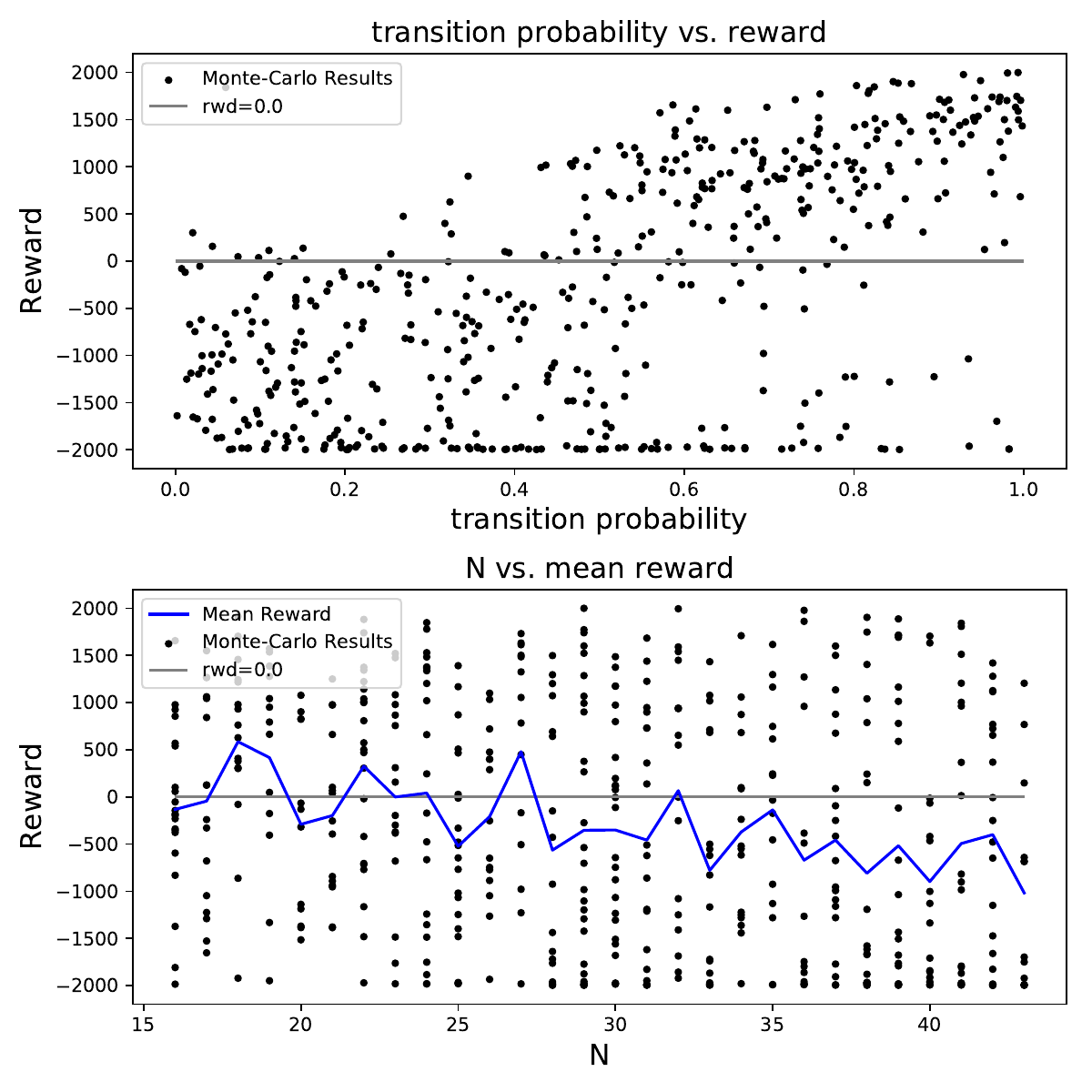}
        \caption{Relationship between transition probability $p_{trans}$ and number of intersection nodes $N$ on the cumulative reward for \solvetrust{}.}
        \label{fig:raw_data}
    \end{figure}

To train a surrogate model with the data from Figure~\ref{fig:raw_data}, we used a simple approach of predicting the mean expected reward \rwdtrustpredict{} for \solvetrust{} with a deep neural network, while a separate neural network was trained to predict the standard error as well. 
To learn \surrogate{} for the expected reward, a neural network was created with two inputs $\bm{x}$: $N$, and $p_{trans}$. This used two fully connected hidden layers with ten neurons each (in order to be certain possible non-linearities were captured), and a single regression output. The activation function for each layer is a rectified linear unit (ReLU), using a mean squared error (MSE) as the loss function. In order to avoid over-fitting a dropout rate of 0.3 was used during training. The same architecture was also used to predict the standard error. It is possible that better models could be used in this situation. While we also briefly investigated other network architectures, we ultimately did not pursue this direction further because we did not want to focus on the unnecessary assumption that the surrogate model be `optimal', and opted instead to demonstrate the flexibility of our approach to non-tailored models.    
    \begin{figure}[t]
        \centering
        \includegraphics[width=0.65\linewidth]{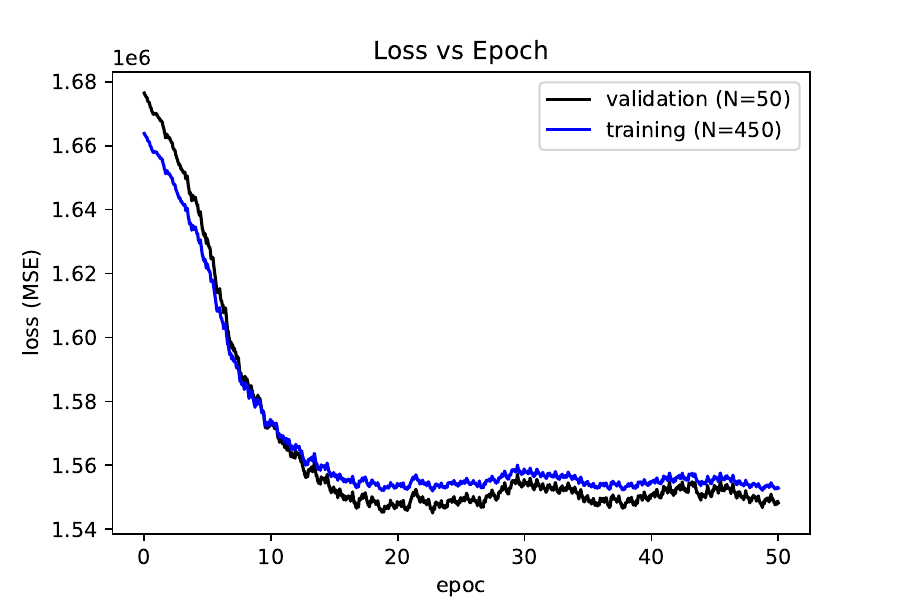}
        \caption{Training and Validation loss (MSE) of the Artificial Neural Network used to predict the mean reward given $N$ and $p_{trans}$.}
        \label{fig:current_fit}
    \end{figure}

In Figure~\ref{fig:current_fit}, the training loss becoming smaller and becoming stable suggests that the neural network is able to learn a model to describe the training data. The MSE of the validation data being so close to the training data suggests that the model did not over-fit to the training data. Together these results suggest that a surrogate model can, in fact, be learned for the ADT problem. Note that the MSE magnitude of this model appears rather high, and does not decrease by orders of magnitude during training. This is mainly an artifact of the uncertain nature of the outcomes (see Fig.~\ref{fig:raw_data}). This raises the question of whether it is really possible that a model with such noisy data can actually be useful in calculating \IS. A key point to remember is that the critical comparison is between $p($\rwdtrustpredict{}$)$ and $p($\rwdcandsim{}$)$. Since these are random variables, it is possible, and in practice quite common, for \rwdcandsim{} and \rwdtrustpredict{} to be significantly different from each other. Thus, the variation of relative error of predicted rewards using \solvetrust{} in different scenarios becomes less critical for this application. 

Note that in this work deep neural networks were mainly chosen to illustrate the ability of this method to work with arbitrary black-box models. One weakness of a standard deep neural net is that they are not very interpretable. There is also an assumption that all of the `test' data is similar to the training data (i.e. \cA{} won't encounter drastically different conditions while deployed than were learned in training). In such a case, other models more cognizant of uncertainty such as Bayesian deep neural nets \cite{jospin2022hands} or Gaussian processes \cite{brochu2010tutorial} might be used.

\end{document}